\pdfoutput=1
\documentclass[pdflatex,sn-mathphys-num]{sn-jnl}%

\usepackage{graphicx}%
\usepackage{multirow}%
\usepackage{amsmath,amssymb,amsfonts}%
\usepackage{amsthm}%
\usepackage[title]{appendix}%
\usepackage{xcolor}%
\usepackage{textcomp}%
\usepackage{manyfoot}%
\usepackage{algorithm}%
\usepackage[normalem]{ulem} 
\usepackage{algorithmicx}%
\usepackage{algpseudocode}%
\usepackage{listings}%
\usepackage{balance}
\usepackage{booktabs}
\usepackage{acro}
\usepackage{mathrsfs}
\DeclareFontFamily{U}{rsfs}{\skewchar\font127 }
\DeclareFontShape{U}{rsfs}{m}{n}{%
  <-6> rsfs5
  <6-8> rsfs7
  <8-> rsfs10
}{}

\newif\ifsubmission
\submissionfalse
\ifsubmission
  \definecolor{teal}{RGB}{0,112,112}
  \definecolor{magenta}{RGB}{155,0,155}
  \definecolor{red}{RGB}{180,0,0}
  \definecolor{blue}{RGB}{0,70,190}
\fi

\makeatother
\DeclareAcronym{llm}{
  short = LLM,
  long  = large language model
}
\DeclareAcronym{vlm}{
  short = VLM,
  long  = vision-language model
}
\DeclareAcronym{cbaa}{
  short = CBAA,
  long  = consensus-based auction algorithm
}
\DeclareAcronym{mpc}{
  short = MPC,
  long  = Model predictive control
}
\DeclareAcronym{cbf}{
  short = CBF,
  long  = Control Barrier Function
}

\theoremstyle{thmstyleone}%

\theoremstyle{thmstyletwo}%
\theoremstyle{thmstylethree}%
\usepackage{setspace}
\usepackage{longtable}
\usepackage{makecell}
\usepackage{listings}
\usepackage[most]{tcolorbox}
\definecolor{pbIndigoInk}{HTML}{3D405B}\definecolor{pbIndigoTint}{HTML}{ECECF1}%
\definecolor{pbTealInk}{HTML}{3C7E61}\definecolor{pbTealTint}{HTML}{E4EFE9}%
\definecolor{pbDeepInk}{HTML}{1D503A}\definecolor{pbDeepTint}{HTML}{E6EEEA}%
\definecolor{pbOchreInk}{HTML}{B3830D}\definecolor{pbOchreTint}{HTML}{F7EBDA}%
\definecolor{pbPeachInk}{HTML}{B34E36}\definecolor{pbPeachTint}{HTML}{FDE8E2}%
\definecolor{pbGreyInk}{HTML}{8F8B85}\definecolor{pbGreyTint}{HTML}{F4F1DE}%
\newtcblisting{promptbox}[2]{breakable, enhanced, listing only, colback=pb#1Tint, colframe=pb#1Ink, boxrule=0.6pt,
  arc=1.5mm, left=4pt, right=4pt, top=4pt, bottom=3pt, before skip=10pt, after skip=8pt,
  attach boxed title to top left={xshift=6pt, yshift=-2.2mm}, boxed title style={colback=white, colframe=pb#1Ink,
  boxrule=0.6pt, arc=1.5mm}, coltitle=black, fonttitle=\small\sffamily, title={#2},
  listing options={basicstyle=\ttfamily\scriptsize, breaklines=true, columns=fullflexible, keepspaces=true, frame=none, xleftmargin=0pt, framexleftmargin=0pt}}

\newcounter{sinote}
\newcounter{sidata}
\newcounter{sivideo}
\newcommand{\sinote}[2]{\refstepcounter{sinote}\label{#1}\section{Supplementary Note~\thesinote\sicapsep{}#2}}
\newcommand{\sidata}[3]{\refstepcounter{sidata}\label{#1}\par\medskip\noindent\textbf{Supplementary Data~\thesidata\sicapsep{}#2}\par\noindent #3\par}
\newcommand{\sivideo}[3]{\refstepcounter{sivideo}\label{#1}\par\medskip\noindent\textbf{Supplementary Video~\thesivideo\sicapsep{}#2}\par\noindent #3\par}
\graphicspath{{figures/}}
\newcommand{\llmquote}[1]{\textit{#1}}
\newcommand{\sicapsep}{\hskip.7em\relax}
\newcommand{\silegendstart}[1][\vspace{3mm}]{\par#1\leftskip0pt\rightskip0pt\parfillskip0pt plus 1fil\figurecaptionfont}
\usepackage{etoolbox}
\makeatletter
\def\email#1{\global\advance\emailcnt by 1\relax
  \ifx\authemail\@empty\g@addto@macro\authemail{\textcolor{blue}{#1}}\else\g@addto@macro\authemail{; \textcolor{blue}{#1}}\fi}
\renewcommand{\@@corrauthor}[2][]{\def\@authfrstarg{#1}\@corauemailtrue\advance\corraucount by 1%
\g@addto@macro\artauthors{%
    \global\@auemailtrue\Authorfont\def\baselinestretch{1}%
    \authorsep{#2}\unskip\ifx\@authfrstarg\empty\else\textsuperscript{\smash{{%
      \@for\@@affmark:=#1\do{\edef\affnum{\@ifundefined{X@\@@affmark}{\@@affmark}{\jmkRef{\@@affmark}}}%
     \unskip\sep\affnum\let\sep=,}}}}\fi\unskip%
    \def\authorsep{\au@and}%
    \global\let\sep\@empty\global\let\@corref\@empty%
}}
\patchcmd{\@maketitle}{Contributing authors:\ }{Emails:\ }{}{\errmessage{arXiv email patch failed}}
\makeatother
\begin{document}

\title[Consensus Within, Competition Between]{Consensus Within, Competition Between: Decentralized LLM Teams for Perimeter Defense}

\author*{\fnm{Khawaja} \sur{Abaid Ullah}}\email{ka4150@rit.edu}

\author{\fnm{Ananya} \sur{Acharya}}\email{aa2334@rit.edu}
\author{\fnm{Mohammad Javad} \sur{Khojasteh}}\email{mjkeme@rit.edu}

\affil{\orgdiv{Department of Electrical and Microelectronic Engineering}, \orgname{Rochester Institute of Technology}, \orgaddress{\street{1 Lomb Memorial Drive}, \city{Rochester}, \postcode{14623}, \state{NY}, \country{United States}}}

\abstract{
The use of large language models (LLMs) to coordinate multi-agent teams raises a fundamental architectural question: whether task allocation should be centralized in a single LLM controller or distributed across autonomous LLM agents, particularly when communication is unreliable. In a deliberately simplified perimeter-defense simulation against LLM-controlled intruders, we compared a central commander, an auction-based consensus team and independent agents under five network conditions. Breach rates were 15\% for the consensus team, 32\% for the commander and 29\% for independent agents. Burst message loss, falsified reports and limited radio range raised the commander's breach rate by 10 to 18 percentage points (pp), whereas the consensus team's changed by at most 5. In the consensus team, giving a low-cost LLM more authority over the bid computation raised its breach rate by 38~pp. With birds and civilian drones in airspace, we added an image for each to the input text state. Only the near-frontier model treated them apart from intruders. Intruders disguised as existing taxidermy bird drones were mistaken for birds by both low-cost and near-frontier LLMs, being dropped from pursuit in 59--61\% of near-frontier decisions (1--16\% before). However, a one-sentence forewarning in the prompt lowered it to 0--25\%. We find that the LLM team with a dedicated external conflict-resolution protocol had the fewest breaches. The code is available at \url{https://github.com/the-kas-lab/consensus-within-competition-between}.
}


\maketitle

\section{Introduction}
Large language models (LLMs) increasingly act as the decision-makers of multi-agent teams, from software agents that solve tasks together~\cite{guo2024large,shen2023hugginggpt,zhan2026agensh,hong2024metagpt} to robot teams that divide work and plan joint actions~\cite{kannan2024smart,mandi2024roco}. In robotics, this division of work is studied as multi-robot task allocation, the question of which robot should execute which task~\cite{gerkey2004formal}. Many LLM teams allocate tasks through a central component, such as an orchestrator that delegates work or a single LLM that assigns subtasks to the whole robot team~\cite{yang2025agentnet,kannan2024smart}. Central coordination brings scalability bottlenecks and a single point of failure, which leaves the team exposed when the coordinator fails or becomes overloaded~\cite{yang2025agentnet}. Decentralized multi-agent LLM systems remove this component. Each agent decides from its own local information and communicates directly with its peers~\cite{tran2025survey}. Without an orchestrator, the agents of AgentNet~\cite{yang2025agentnet} reached higher accuracy than centralized multi-agent baselines on mathematics, logical question-answering and function-calling benchmarks, and robot arms with one LLM agent each proposed sub-task plans and paths through dialogue~\cite{mandi2024roco}. The evidence is not one-sided. In multi-robot planning tasks, Chen et al.~\cite{chen2024scalable} compared four designs for coordinating the LLMs of a robot team. The robots in their study cooperated without opponents, and each LLM agent was assumed to know the full environment. In their decentralized architecture, agents relayed messages through a peer-to-peer network and reached consensus through sequential debate rounds, iterative challenge-response exchanges designed to verify reasoning steps and converge on consistent information. This setup had the lowest success rate, below a central planner, while hybrid designs did best and the central planner scaled worst as robots were added. In a separate study, Kim et al.~\cite{kim2025scaling} evaluated 260 agent-system configurations and found that independent agents, which do not communicate, amplified errors more than centrally coordinated ones, and whether coordination helped depended on the task. Furthermore, network degradation and agent attrition are unavoidable in distributed multi-agent systems; therefore, it is imperative to compare the centralized and decentralized coordination of LLMs under these conditions. We study this coordination problem in an adversarial game under various network conditions, which serves as a simple abstraction of a cyber-physical system, aiming toward applications where an LLM in a hierarchical control setting decides and allocates tasks as an intruder approaches.

In perimeter defense, a team of defenders must intercept intruders before they reach a protected region~\cite{shishika2020review}, and the intruders score for each one that gets through~\cite{shishika2020cooperative}. We call an intruder's entry into the protected region a breach. Perimeter defense belongs to the broader family of pursuit-evasion and reach-avoid differential games~\cite{chung2011search, weintraub2020introduction}, in which a team of intruders attempts to reach a target region while defenders attempt to capture or neutralize them~\cite{chung2011search,weintraub2023range,weintraub2020introduction,zhou2025control}. 
Game-theoretic analyses build team strategies by decomposing the game into local games between small groups and assigning defenders to intruders~\cite{shishika2020cooperative}, and learned policies imitate a maximum-matching assignment from each defender's local perception~\cite{lee2023gnn}.  For instance, small groups of drones in Ukraine reportedly decide among themselves which one strikes and when~\cite{macdonald2025swarms}. In scenarios such as counter-drone defense, defenders must detect intruders with limited sensing~\cite{park2021antidrone}, share observations over a communication network~\cite{maity2024target}, and cope with message loss over non-ideal channels~\cite{otte2020auctions}. Assigning defenders to intruders is therefore an allocation decision made under partial and evolving information, where the contextual reasoning of LLM planners can be beneficial. Therefore, for this study, we build a perimeter defense simulation incorporating environmental characteristics such as partial observability, a communication network with varying conditions, and assign an LLM to each defender and intruder for decision making.

Although foundation models~\cite{bommasani2021opportunities}, including large language models (LLMs) and vision-language-action models, have been applied across many aspects of robotics, from semantic perception to end-to-end control~\cite{firoozi2025foundation,hu2023toward}, incorporating them into the decision making loops of real-time scenarios invites its own challenges, primarily the slow LLM inference~\cite{firoozi2025foundation}. To address these challenges, a common pattern is hierarchical wherein a language model plans at a high level while lower-level skills or controllers execute the plan~\cite{ahn2022can}. Acharya et al.~\cite{acharya2026capture} propose a similar hierarchical architecture specifically for pursuit evasion in which a per-team LLM planner assigns tactical roles to defender agents with execution delegated to low-level model predictive controllers. We follow their core design. In our study, a call to the LLM planner took a median of 2.0\,sec, while model predictive control (MPC)~\cite{morari2015predictive,mayne2000constrained} converted each agent's task into accelerations every 0.1\,sec, and a barrier filter kept agents collision-free except for when a defender pursued to neutralize an intruder. An important design simplification we introduce is synchronization, i.e. the simulation is paused until an API call to the LLM returns with a response. This request-response only consumes 0.1\,sec of simulation time. This is done to ensure that the differences in the comparison between the architectures come from the difference in decision making and not because of difference in API response time.

How a team allocates tasks determines what it needs from its communication network. A central allocator can in principle find the best assignment by gathering all information and planning for the whole team, but it has high communication demands and is a single point of failure~\cite{dias2006market}. When radio range is limited, reports and orders must be relayed, and a partitioned network cuts agents off from both.
Fully distributed allocation avoids this round trip, but good local choices may not add up to a good team assignment~\cite{dias2006market}, and agents that each pick their best task can pick the same one~\cite{gerkey2004formal}. Market-based methods combine the two by letting agents bid for tasks~\cite{dias2006market}. The consensus-based auction algorithm (CBAA) reaches a conflict-free assignment through local message exchange alone and tolerates inconsistent information across the team~\cite{choi2009consensus}. Auction methods with conventional bids have been tested under lossy and fading communication~\cite{otte2020auctions,lott2026decentralized}, and LLM teams have been compared under centralized, decentralized and hybrid communication on cooperative planning tasks~\cite{chen2024scalable}. By framing the problem as an adversarial game, we let teams of LLMs compete directly against one another, and we compare a central LLM commander with decentralized LLM agents that resolve allocation conflicts through a consensus auction, under various different communication networks. In a multi-agent system, one compromised agent can pass adversary-written content into the network, where it spreads from agent to agent~\cite{lee2025prompt,gu2024agent}. We therefore also evaluate centralized and decentralized architectures when one of the  defenders is compromised and falsifies the positions it reports (Section~\ref{sec:methods:comms}).

Most of our experiments give the LLMs only a textual state, which describes where each contact is and how it moves but not what it is: an intruder, or a bird or civilian drone sharing the airspace. Pursuing a bird or civilian drone costs more than just time, as a defender that collides with any contact is lost, resulting in a reduction of defensive capacity.  This is exemplified by counter-UAV missions, in which inexpensive decoy drones must be identified~\cite{mittal2026gerbera}. We extend the input to a multimodal state that adds an image of each contact to the text, so that the defenders' LLMs, used zero-shot, can tell intruders (\emph{hostile} contacts) apart from birds and civilian drones (\emph{non-hostile} contacts) before deciding whether to engage. How much this helps depends on how reliably the model classifies what it sees, so we run each team with both a low-cost and a near-frontier LLM. Image-based recognition, however, has weaknesses of its own. Vision models can be deceived by the appearance of a physical object even when a human observer is not: Athalye et al.~\cite{athalye2018synthesizing} constructed 3D-printed objects that a classifier mislabels consistently across viewpoints. A defender that decides from appearance therefore gives the adversary a new way to attack. We demonstrate one simple case that requires no optimization against the model: intruders disguised as birds, shown as images of existing taxidermy bird drones~\cite{khan2023taxidermy,vosbein2026advancements} in which mechanical parts remain visible. Such failures deserve study as AI enters safety-critical decisions (e.g., AI-based target recognition on drones~\cite{bondar2025ukraine} and the reported misidentification of a ship's cargo by an AI chatbot~\cite{cnn2026aierror}). 

We study how centralized and decentralized LLM teams, with each model, handle hostile contacts, non-hostile contacts and disguised intruders. Inspired by the system-prompt warning that protected agents in~\cite{papadopoulos2026mind}, we also test a one-sentence forewarning against a different attack, the visual disguise discussed below.

\begin{figure}[ht]
\centering
\includegraphics[width=0.9\textwidth]{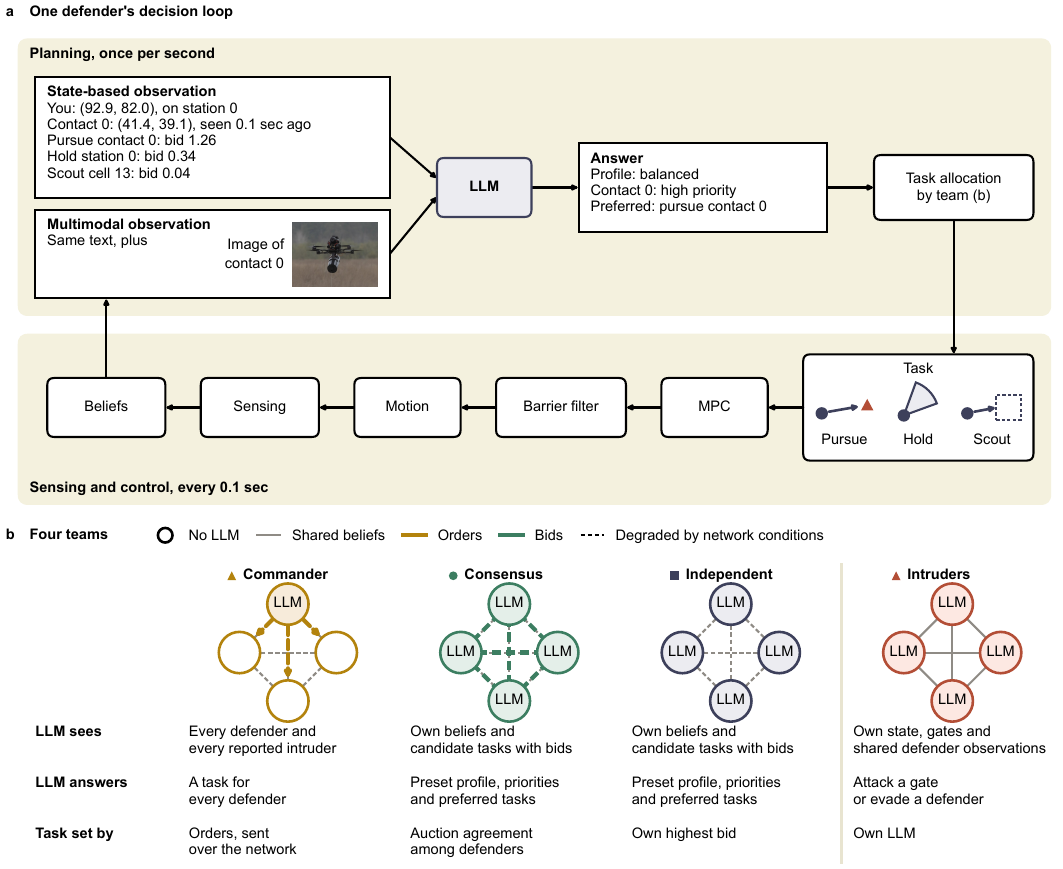}
\caption{\textbf{How the defender and intruder teams decide.}
\textbf{(a)} One defender's decision loop, shown with one recorded call of a consensus-team defender, shortened and in plain words. Once per second the LLM receives an observation built from the defender's beliefs: state-based in the main study, a text description of the game, and multimodal in the airspace challenges, the same text plus an image of every contact. How its answer becomes a task differs by team (See \textbf{b}). Every 0.1~sec, model predictive control (MPC) plans motion for the task, the barrier filter adjusts that motion to keep agents apart, reading the true positions of all agents, and the sensor updates the beliefs, which teammates share. \textbf{(b)} Which team members run an LLM, what they send each other, and the steps from observation to task. The consensus and independent teams' LLMs have low authority (Methods). The network conditions affect only messages between defenders. Drone Image: ArmyInform (CC BY 4.0, cropped).}
\label{fig:architecture}
\end{figure}

In this study we compare three LLM-coordinated defender teams that differ in where LLM decision-making is placed within the defensive architecture (Fig.~\ref{fig:architecture}). The first, the \emph{commander team}, places all authority in a central commander, whose LLM assigns tasks to every agent. The second, the \emph{consensus team}, has decentralized agents that bid for tasks and agree on who takes which through a consensus-based auction. Each agent’s LLM adjusts its own bids. We further study how much control the LLM should have over its bids, which we refer to as its \emph{authority}. At \emph{low authority} the LLM can only choose a weight profile, a priority for each intruder and up to three preferred tasks. At \emph{high authority} it sets the weights of the bid formula and a multiplier from 0 to 2 for each intruder’s pursuit value. An LLM at high authority can therefore \emph{drop} an intruder, removing its pursuit by setting its pursuit multiplier to zero, as a commander can by assigning it no defender, whereas at low authority it can only \emph{deprioritize} one, lowering the value of its pursuit while the pursuit stays possible. The third, the \emph{independent team}, uses the same LLMs and bids, but each agent chooses its task independently based on its own highest bid, without an auction.

Each team defends a region for 40 seconds against LLM-controlled intruders from 30 random starting configurations (seeds), with two and four intruders, under five network conditions (an ideal network, independent message loss, burst message loss, a compromised defender that falsifies reports, and limited radio range with relaying). Five lessons emerged. First, conflict resolution mattered. Across the four network conditions without falsified reports, intruders breached the region in 15\% of episodes against the consensus team, 32\% against the commander team, and 29\% against the independent team, 3 percentage points (pp) below the commander team, with most of the difference arising in episodes with four intruders. Second, the commander team was the most sensitive to network degradation and the consensus team the least. The commander team's breach rate rose by 10~pp under burst message loss, 12~pp under falsified reports and 18~pp under limited radio range (95\% confidence interval (CI): +1 to +36), whereas the consensus team's changed by at most 5~pp under any condition. Independent message loss changed no team's breach rate by more than 2~pp. Third, level of authority assigned to LLM for decision making should depend on the capability of LLM being used as with a low-cost model, high authority raised the number of breaches by 38~pp, whereas a near-frontier model at high authority was breached as often as the low-cost model at low authority. As commander, the near-frontier model was breached 12~pp less often than the low-cost model on the ideal network and 30~pp less often under falsified reports. Fourth, non-hostile contacts, which tested the models' visual-language capabilities, showed the same divide. Across five seeds, the near-frontier model never lost a defender to a non-hostile contact in either the centralized or the decentralized role, whereas each low-cost model team, lost 3 to 5 defenders in total across the five episodes. Fifth, without a forewarning about disguised intruders in the prompt, both low-cost and near-frontier LLMs were fooled by the intruders disguised as birds.

These lessons come from a testbed that is deliberately simplified, so that coordination can be studied under control, and the effects of team architecture, model capability, and communication degradation can be isolated from simulation factors such as perception difficulty. Agents move as point masses with bounded acceleration in a two-dimensional arena of abstract units, with no wind or other disturbance. They sense opponents without noise and without missed detections inside a fixed field of view. The images are reference pictures rather than views from an onboard camera. We make no claim that this setup represents a deployed counter-drone system. Its value is as a benchmark for LLM-driven coordination, a simplified abstraction of cyber-physical teams.
In summary, we make the following contributions.
\begin{itemize}
\item We establish a benchmark for multi-agent LLM coordination in a perimeter-defense game, in which LLM-driven defenders face LLM-controlled intruders over a network that can be degraded. The LLMs are used as released, without fine-tuning, task-specific training or example answers, and every episode can be re-simulated %
from its recorded model answers.
\item We compare central and decentralized LLM coordination on matched episodes under message loss, limited radio range and falsified reports, which gives the first two lessons.
\item We test how much an LLM should be allowed to decide, by running the consensus team at high authority with a low-cost and a near-frontier model, which gives the third lesson.
\item We extend multi-agent LLM coordination to multimodal, image-based contact classification, with birds, civilian drones and intruders disguised as taxidermy bird drones.
\end{itemize} 

\section{Results}
\label{sec:results}
We first describe the testbed, the three kinds of defender teams, the five network conditions and the outcome we measure. Then we present the results of comparisons of the three different kinds of defender teams, all run with the low-cost model on the same 30 seeds. These results describe how often each team was breached, how well it divided its tasks among its defenders, and how it responded to message loss, falsified reports and limited radio range. We then present the findings of the comparison of authority level assigned to LLMs in the consensus team, along with the effect of model capability, by using a near-frontier model in place of the low-cost LLM. Finally, we present the results of introducing civilian drones, birds and disguised intruders in the simulation and how that affected different teams.

\subsection{Teams, conditions and outcome}
We built a testbed in which four defenders protect a region against two or four intruders, and every agent on both teams  uses the same low-cost LLM, Gemini 3.1 Flash-Lite, once per decision epoch of 1\,sec (Fig.~\ref{fig:arena}; Section~\ref{sec:methods}). A defender’s task is to pursue an intruder, hold one of four stations around the region or scout part of the arena, and an intruder’s task is to attack the region or evade a defender. \ac{mpc}~\cite{mayne2000constrained} turns each agent's task into motion every 0.1\,sec, and a barrier filter adjusts that motion to keep agents apart except for when a defender is pursuing to neutralize an intruder, in which case the barrier filter of the pursuing defender is switched off for only the target being pursued. Each agent’s sensor covers a 60\textdegree{} field of view (FOV) and reaches 35 units, a distance we call the \emph{sensor range}. An agent sees opponents only inside this area, so it acts on its beliefs about them. A belief holds an opponent’s last observed position and velocity, from the agent’s own sensor or a teammate’s message, projected forward based on the velocity and time of observation.
The three defender team configurations considered in this study differ how their beliefs are turned into tasks (Fig.~\ref{fig:architecture}, Supplementary Videos 1--3). In the commander team, the defender with the lowest index (by default defender 0) is assigned the commander role, and its LLM assigns a task to every defender, including itself.  In the consensus team, each defender scores every task it could take, and this score is its bid for the task. The defender's LLM sees its beliefs and these candidate tasks and, at low authority, adjusts the bids only by choosing among options written in advance. The bids then enter the consensus-based auction algorithm (CBAA)~\cite{choi2009consensus}. The independent team runs similar to the consensus team but with the auction messages switched off, so each defender commits to its own highest bid. 

Each team ran under five network conditions: an ideal network, independent message loss, burst message loss with the same average rate, falsified reports from one compromised defender, and a radio range of 15 units that forces relaying (defenders on their stations are 11 to 18 units apart). With two and four intruders across 30 seeds, this gave 900 episodes (Section~\ref{sec:methods}). 

A breach occurs if an intruder enters the defended region within the 40 sec episode. Breaches were rare with two intruders (35 of 450 episodes) and common with four (196 of 450, Supplementary Fig. 1 and Supplementary Table 1). Many episodes without a breach ended at the time limit with an intruder still alive, 126 of 254 with four intruders and 78 of 415 with two, and 113 of the 161 intruders alive at the end of four-intruder episodes were still closing on the region. Breach rates therefore only count breaches within 40 sec. 

Additionally, intruder-side decision making is less complex as they always share an ideal communication channel (without any degradation) and either decide to evade or attack at each planning epoch. Therefore, they do not have task conflicts and do not require an explicit consensus mechanism like the defender team (Fig.~\ref{fig:architecture}b; Section~\ref{sec:methods:intruders}). Every intruder uses the low-cost model in every experiment, so all defender teams face the same opponent. Therefore, each episode is a competition between two teams of LLM agents.

\clearpage
\subsection{Consensus-based allocation led to fewer breaches}
\begin{figure}[ht]
\centering
\includegraphics[width=\linewidth]{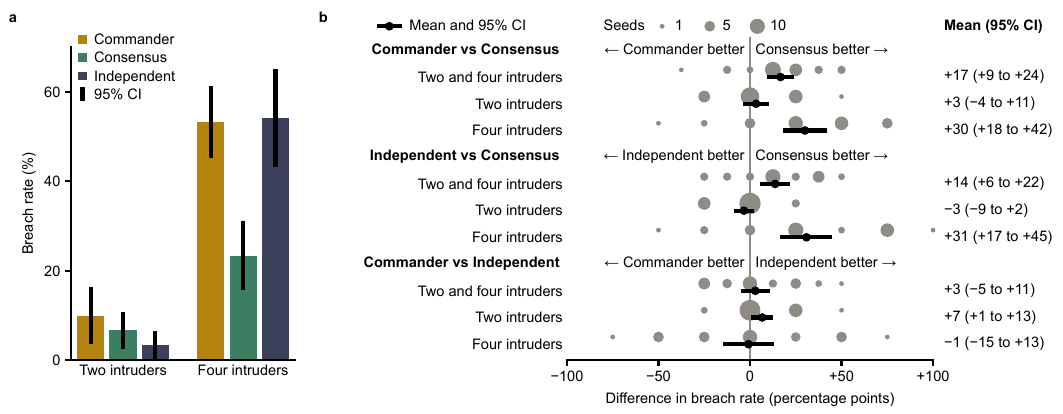}
\caption{\textbf{Consensus allocation lowered breaches.}
\textbf{(a)} Breach rate of each team with two and with four intruders, over the four network conditions without falsified reports (120 episodes per bar). \textbf{(b)} Difference in breach rate between the two teams of each block, the first-named team's rate minus the second's, for both intruder counts together and for each; ``better'' means fewer breaches. Grey circles are the per-seed differences, with area proportional to the number of seeds sharing a value. Black dots are means over the 30 seeds; black bars in \textbf{a} and \textbf{b} are 95\% CIs from the t distribution over seeds. CI, confidence interval.}
\label{fig:contrasts}
\end{figure}

With each agent on both sides running the low-cost model, we study the breach rates of the three kinds of defender teams considered. Out of the three teams, the consensus team was breached least often (Fig.~\ref{fig:contrasts}). Over the four conditions without falsified reports, intruders breached the region in 32\% of commander team episodes, 29\% of independent team episodes and 15\% of consensus team episodes (240 episodes per team). The commander team's breach rate---share of episodes with a breach---was 17 pp higher than the consensus team's (95\% confidence interval: +9 to +24), and the consensus team's was 14 points lower than the independent team's (95\% CI: \textminus22 to \textminus6). The commander team's breach rate was 3 pp higher than the independent team's (95\% CI: \textminus5 to +11). The differences were concentrated in episodes with four intruders (Supplementary Table~2). With four intruders, the commander team's breach rate was 30 pp higher than the consensus team's (95\% CI: +18 to +42), and the independent team's was 31 pp higher than the consensus team's (95\% CI: +17 to +45). Resolving allocation conflicts through the auction thus gave the consensus team the fewest breaches of the three teams, mostly through the episodes with four intruders, in which breaches were common.

\subsection{Duplicated commitments and missed scouting}
\begin{figure}[ht]
\centering
\includegraphics[width=\linewidth]{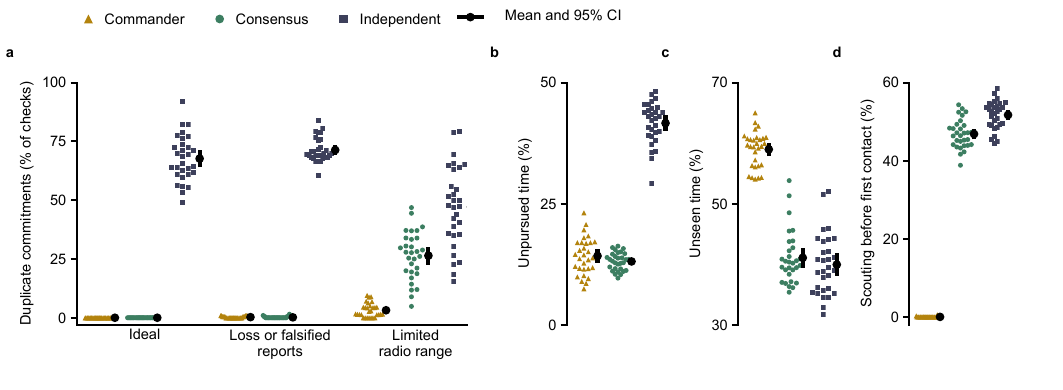}
\caption{\textbf{The auction removed duplicate commitments, and the commander barely scouted.}
\textbf{(a)} Share of the once-per-second checks at which two or more active defenders were committed to the same task, by network condition; independent loss, burst loss and falsified reports are pooled. For the commander team, a defender's commitment is the newest order it has accepted. \textbf{(b)} Unpursued time: share of the time a known intruder had no defender pursuing it. \textbf{(c)} Unseen time: share of an intruder's active time in which no defender saw it. \textbf{(d)} Share of defender time spent scouting before the team's first contact. \textbf{b} and \textbf{c} cover the four conditions without falsified reports, \textbf{d} all five conditions; every panel pools two and four intruders. Each team-colored marker is one seed; black dots and bars are means and 95\% CIs from the t distribution over the 30 seeds. CI, confidence interval.}
\label{fig:allocation}
\end{figure}

\emph{Duplicate commitments} occur when two or more defenders commit to the same task, such as pursuit of the same intruder, which wastes defenders that could cover other tasks. In our study, for the decentralized teams, the auction mechanism in the consensus team removed duplicate commitments (Fig.~\ref{fig:allocation}a and Supplementary Table~3). We checked for duplicate commitments once per second, at the end of each decision epoch, using the tasks the defenders were carrying out at that moment. On the ideal network, two or more consensus-team defenders were never committed to the same task (0 of 2,035 checks), whereas two or more independent-team defenders were in 1,288 of 1,896 checks (68\%). Supplementary Figs. 2 and 3 show one episode of each of these two teams from the same starting positions. Across the ideal network, both loss conditions and the falsified-report condition, the consensus team had duplicate commitments in only 10 of 7,988 checks, with no seed above 1.1\%, whereas the independent team's per-seed share ranged from 61\% to 81\%. Under limited radio range, duplicate commitments returned: they occurred in 27\% of consensus-team checks (508 of 1,906) and 48\% of independent-team checks, and the consensus team's share was lower in 25 of the 30 seeds.

Although the commander team had almost no duplicate commitments, it however showed a passive behavior, which we refer to as the \emph{missed scouting}, in which defenders hold their stations rather than proactively searching the arena for intruders that no defender has seen (Fig.~\ref{fig:allocation}d). The commander team's LLM assigned scouting in only 16 of 39,077 answered tasks. Before the team's first contact, its defenders scouted in 58 of 194,566 defender-steps (one defender during one 0.1\,sec step) and otherwise held their stations, whereas defenders in the two decentralized teams scouted in 47\% to 54\% of their defender-steps before first contact. Consistent with this rare scouting, intruders went unseen by every defender for 59\% of their active time against the commander team, compared with 41\% against the consensus team (Fig.~\ref{fig:allocation}c).

Furthermore, we found that duplicated pursuit co-occurred with intruders that no defender pursued (Fig.~\ref{fig:allocation}b). Over the four conditions without falsified reports, a known intruder had no defender pursuing it for 42\% of its time in the independent team and 13\% in the consensus team. The consensus team had 28.6 pp (95\% CI: \textminus30.1 to \textminus27.0) lower unpursued time than the independent team. In the commander team it was 14\%, 1.1 pp above the consensus team (95\% CI: \textminus0.2 to +2.5). 

\subsection{Degraded networks hurt the commander team most and the consensus team least}
\begin{figure}[ht]
\centering
\includegraphics[width=\linewidth]{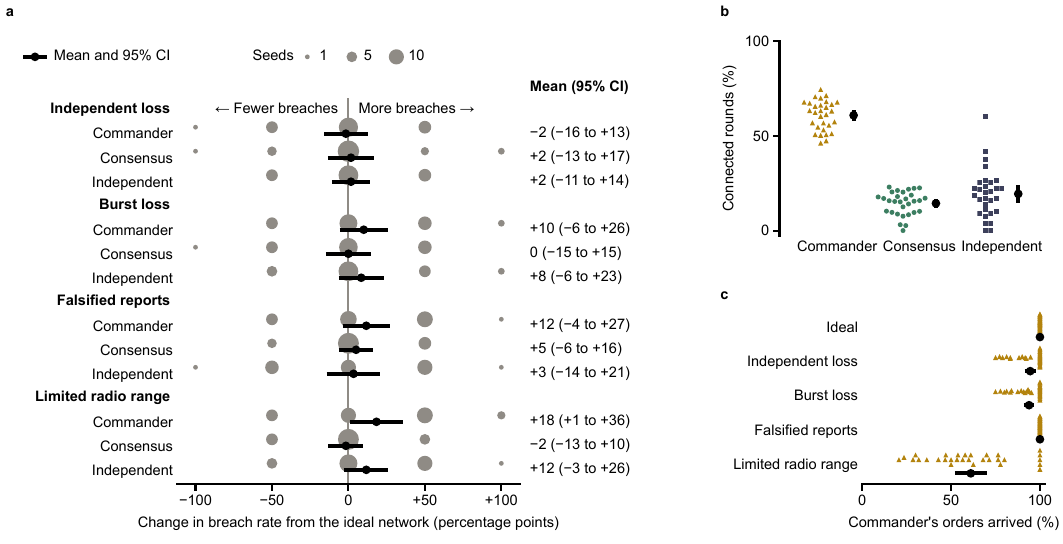}
\caption{\textbf{Degraded networks raised the commander team's breach rate the most.}
\textbf{(a)} Change in breach rate from the ideal network under each other network condition, by team, pooling two and four intruders (two episode pairs per seed). Grey circles are the per-seed changes, with area proportional to the number of seeds sharing a value. \textbf{(b)} Under limited radio range, the share of communication rounds in which the active defenders formed one connected group; each marker is one seed, its two episodes averaged. \textbf{(c)} Of the commander's orders that would have changed a defender's task, the share that arrived, by network condition; each triangle is one seed. Black dots and bars are means and 95\% CIs from the t distribution over the 30 seeds. CI, confidence interval.}
\label{fig:range}
\end{figure}

Besides the ideal network, we tested four degraded networks (Section~\ref{sec:methods:comms}): independent message loss; burst loss, with the same average loss rate but losses arriving in runs; falsified reports, in which one compromised defender misreports an intruder's position; and limited radio range. Burst loss hurt more than independent loss at the same average rate (Fig.~\ref{fig:range}a; Supplementary Table~4). Independent loss changed each team's breach rate by at most 2 pp. Burst loss raised the commander team's breach rate by 10~pp (95\% CI: \textminus6 to +26) and the independent team's by 8~pp (95\% CI: \textminus6 to +23), and left the consensus team's unchanged (0 points; 95\% CI: \textminus15 to +15).

Falsified reports also hurt the commander team most. The compromised defender, drawn at random from the defenders other than defender~0 (as defender~0 serves as the commander in the commander team), falsified an intruder's position in its reports in 55 to 60 of each team's 60 episodes in this condition. In the commander team, the commander accepted at least one falsified report in 55 of them (Supplementary Table 5). Pooled over teams and intruder counts, falsified reports raised the breach rate by 7~pp relative to the ideal network (95\% CI: \textminus2 to +15): by 12~pp for the commander team (95\% CI: \textminus4 to +27), 5~pp for the consensus team (95\% CI: \textminus6 to +16) and 3~pp for the independent team (95\% CI: \textminus14 to +21).

Limited radio range fragmented the decentralized teams' networks the most, yet raised the commander team's breach rate the most (Fig.~\ref{fig:range}). With a radio range of 15 units, the commander team's network was connected on 61\% of communication rounds, against 14\% for the consensus team and 19\% for the independent team; with two intruders, the consensus team's network was connected on only 2\% (Fig.~\ref{fig:range}b and Supplementary Table 6). The commander team's breach rate rose by 18~pp (95\% CI: +1 to +36) and the independent team's by 12~pp (95\% CI: \textminus3 to +26), whereas the consensus team's fell by 2~pp (95\% CI: \textminus13 to +10). The commander team's rise was 20~pp larger than the consensus team's (95\% CI: +2 to +38) and 7 points larger than the independent team's (95\% CI: \textminus16 to +30).

\subsection{Low-cost model with high authority resulted in more breaches in consensus team}
\begin{figure}[!ht]
\centering
\includegraphics[width=\linewidth]{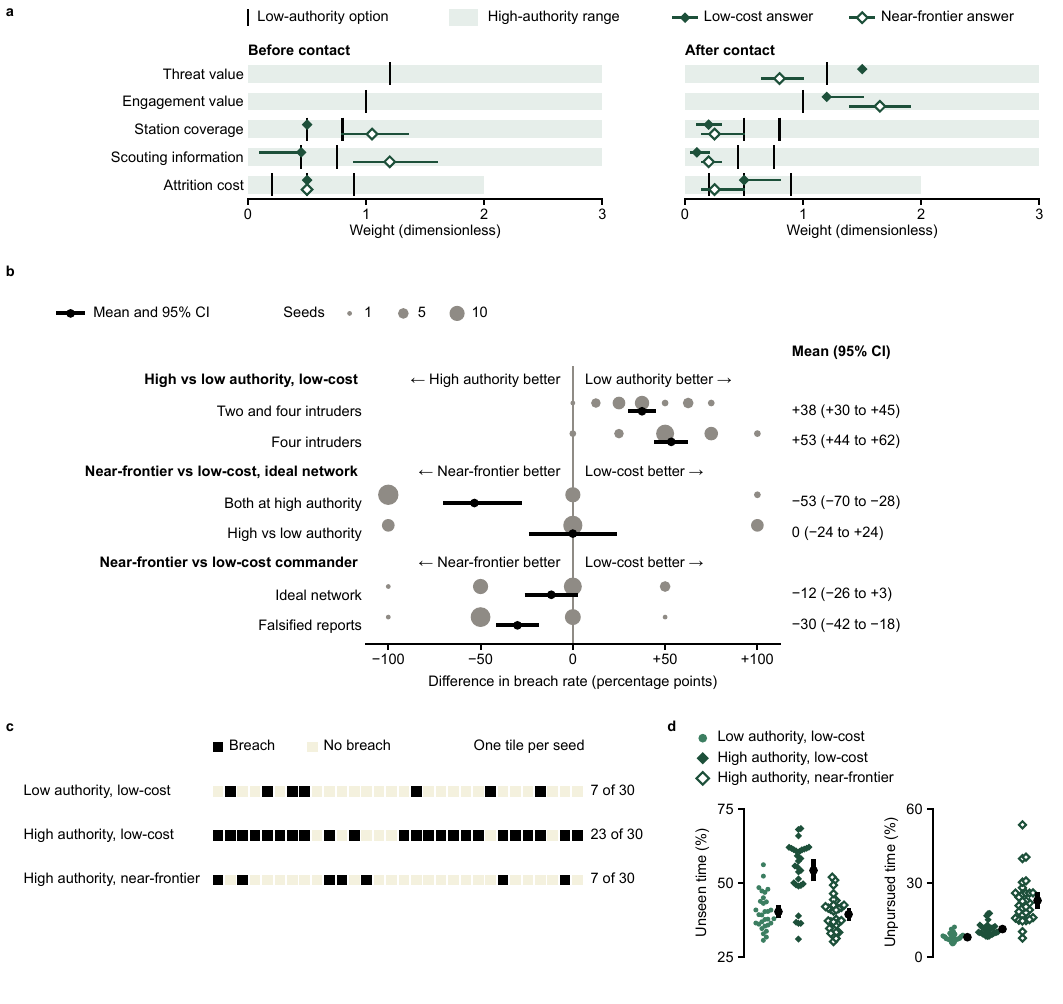}
\caption{\textbf{High authority hurt the low-cost model.}
\textbf{(a)} Bid weights of the consensus team (Supplementary Table 12). Markers are median answers at high authority on the ideal network with four intruders, bars their 25th to 75th percentiles (Supplementary Table 7); pursuit weights act on no task before contact. \textbf{(b)} Breach-rate differences, first-named minus second; ``better'' means fewer breaches. Low-cost rows pool the four conditions without falsified reports; ideal-network rows have four intruders and one episode per seed; commander rows pool two and four intruders. Grey circles are per-seed differences. \textbf{(c)} Breach in each seed on the ideal network with four intruders. \textbf{(d)} For the episodes of \textbf{c}, intruder time unseen by every defender and known-intruder time with no defender pursuing; one marker per seed. Black dots and bars are means and 95\% CIs (t distribution over 30 seeds). CI, confidence interval; low-cost LLM, Gemini 3.1 Flash-Lite; near-frontier LLM, GPT-6.1 Sol.}
\label{fig:authority}
\end{figure}

Low authority keeps the LLM from dropping a known intruder or rewriting the formula's tradeoffs. We therefore also study how the consensus team performs when its LLMs have more authority, which helps determine how much authority LLMs should be given in a given scenario. For this purpose, we ran the consensus team with high authority (Fig.~\ref{fig:authority}a). Each defender's LLM then sets 12 weights of the bid formula and a multiplier from 0 to 2 on each intruder's pursuit value, its pursuit multiplier, every epoch, rather than choosing among options written in advance (Section~\ref{sec:methods:defender:bids}). Supplementary Table~7 lists the values chosen. 

With the low-cost model, the consensus team was breached in 53\% of episodes at high authority against 15\% at low authority, over the four conditions without falsified reports and both intruder counts. That is 38~pp higher (95\% CI: +30 to +45). The gap was larger with four intruders, where the two breach rates were 77\% and 23\%, 53~pp apart (95\% CI: +44 to +62). High authority co-occurred with weaker detection. Once it knew of an intruder, the low-cost model's median answer raised the threat value and set the weights on station coverage and scouting below any value low authority allows (Fig.~\ref{fig:authority}a). Intruders went unseen by every defender for 53\% of their active time at high authority against 41\% at low authority, 12~pp more (95\% CI: +10 to +14).

The near-frontier model at high authority was breached as often as the low-cost model at low authority (Supplementary Table 8). On the ideal network with four intruders, the consensus team at high authority run with GPT-6.1 Sol was breached in 7 of 30 episodes, against 23 of 30 with the low-cost model, a breach rate 53~pp lower (95\% CI: \textminus70 to \textminus28). The near-frontier model used its authority differently. Before contact its median answer weighted station coverage and scouting above any low-authority value, and after contact it lowered the threat value and raised the engagement value (Fig.~\ref{fig:authority}a). Intruders went unseen for 39\% of their time, close to the 40\% at low authority and below the low-cost model's 54\% at high authority. Known intruders had no pursuer for 24\% of their time, against 8\% at low authority, with the same number of breaches.

With the near-frontier model as the commander's LLM, the team was breached less often both on the ideal network and under falsified reports. Over both intruder counts, the commander team using GPT-6.1 Sol was breached in 8 of 60 ideal-network episodes and 4 of 60 episodes with falsified reports, compared with 15 and 22 for the commander team using the low-cost model. Its breach rate was 12~pp lower on the ideal network (95\% CI: \textminus26 to +3), and 30~pp lower under falsified reports (95\% CI: \textminus42 to \textminus18).

\subsection{Airspace challenges: civilian drones, birds, and intruders disguised as birds}
\label{sec:results:airspace}
\begin{figure}[ht]
\centering
\includegraphics[width=\linewidth]{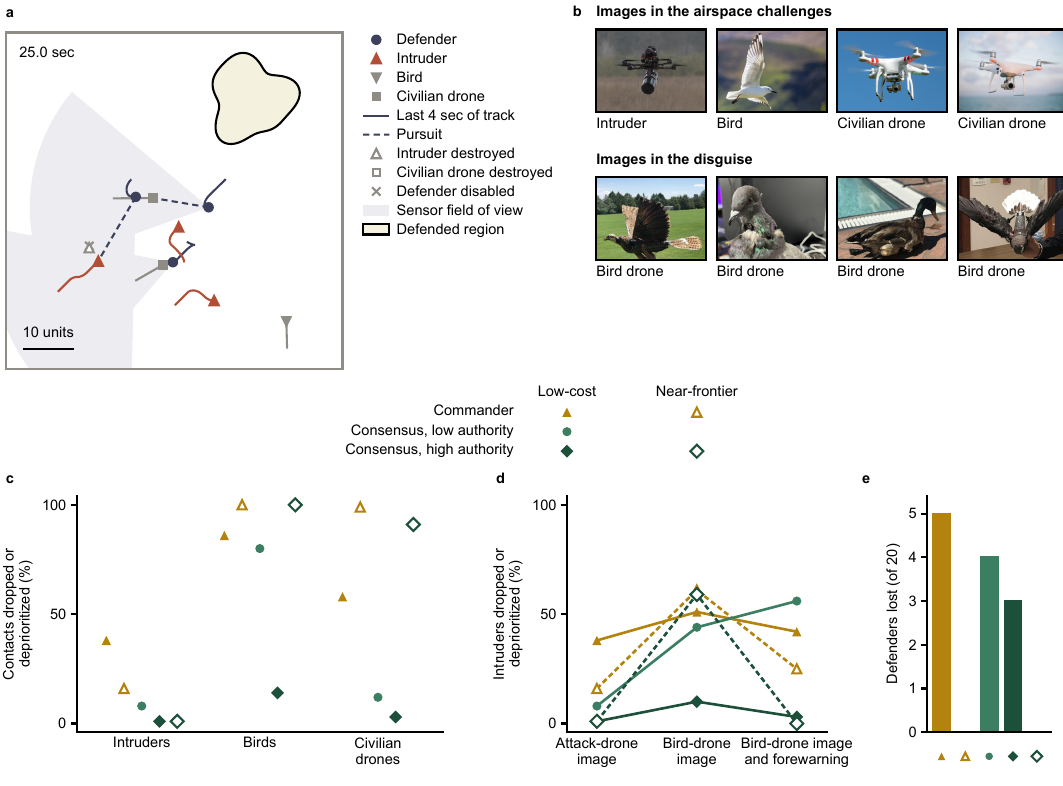}
\caption{\textbf{Airspace challenges divided the two models.}
\textbf{(a)} One frame of a consensus-team episode, low-cost LLM, at 25~sec, with one bird and two civilian drones among four intruders. The rightmost defender and the civilian drone it was pursuing were lost 0.7~sec later. \textbf{(b)} The images the LLMs received in this episode, and the four bird drones shown instead for the intruders in the disguise. \textbf{(c)} Share of the LLMs' answers that dropped a contact (commanders and high authority) or deprioritized it (low authority), by kind of contact, without the disguise; for commanders, share of decisions about a contact, one per contact per epoch, with no defender assigned. \textbf{(d)} Intruders dropped or deprioritized with each kind of image, and with the forewarning; dashed lines join the near-frontier teams. \textbf{(e)} Defenders lost to birds and civilian drones over the five episodes without the disguise. Five seeds, ideal network, four intruders. Drone and bird images: ArmyInform (CC BY 4.0, cropped); R. Long/Unsplash; succo/Pixabay and J. Sorenson/Pexels (CC0); bird drones, M. Hassanalian, New Mexico Tech.}
\label{fig:airspace}
\end{figure}

We extend our study to test multimodal and zero-shot capabilities of modern LLMs by introducing non-hostile objects such as birds and civilian drones, typical of an airspace, into the simulation in addition to the intruders. We added one bird and two civilian drones to each episode on the ideal network with four intruders (Fig.~\ref{fig:airspace}a; Section~\ref{sec:methods:airspace}). These objects follow inbound trajectories like those of intruders but cannot breach the defended region and do not affect the score; however, any defender that collides with one is disabled. In the main study every LLM works from a state-based observation (Section~\ref{sec:methods:sensing}), a text description of positions, velocities and the values computed from them, with no images. Here the defenders' LLMs utilize their multimodal capabilities, and reason over images and text together. They receive the same text plus one image of every contact they are asked about, intruders included, so each decision now also rests on image-based contact classification, telling from the image what a contact is. The images are reference pictures rather than views from a defender's sensor. Their prompts also gain a factual airspace paragraph stating that not every contact is an intruder and giving no tactical advice (Section~\ref{sec:methods:prompts}). We evaluated five team configurations over five seeds each: the commander team with each of the two models, the consensus team at low authority with the low-cost model, and the consensus team at high authority with each of the two models (Fig.~\ref{fig:airspace}; Supplementary Tables~9 and~11). As in the main study, an LLM \emph{drops} a contact when its answer removes the contact's pursuit: a commander assigns it no defender, or an LLM at high authority gives it a pursuit multiplier of zero, which leaves its pursuit with no value. At low authority an LLM can only \emph{deprioritize} a contact, by giving it a low priority, which halves the value of its pursuit but leaves the pursuit among the defender's options.

The near-frontier model lost no defender to a non-hostile contact in either setting, whereas each low-cost team lost 3 to 5 of its 20 defenders over the five episodes, never more than one per episode (Fig.~\ref{fig:airspace}e). Of these 12 losses, 11 were to civilian drones. As commander, the near-frontier model assigned a defender to a bird or civilian drone in 2 of the 361 epochs in which it saw one, and at high authority it dropped 100\% of birds and 91\% of civilian drones in its answers (Fig.~\ref{fig:airspace}c and Supplementary Table~10). The low-cost model deprioritized or dropped birds more often than civilian drones in each of its three interfaces.

With these images the low-cost model rarely deprioritized or dropped civilian drones in the consensus team, at either level of authority. Only 12\% of its answers at low authority deprioritized them and 3\% of its answers at high authority dropped them, so most of its losses began with a wrong judgment. The interface then sets what a correct judgment can achieve.

At low authority a low priority sets the contact's priority $\pi_j$ to 0.5, a value we chose in design rather than derived, which halves its engagement value of pursuit but does not remove its pursuit task (Section~\ref{sec:methods:defender:bids:low}). At high authority a pursuit multiplier of zero leaves the contact's pursuit bid with only its costs, so the contact loses the auction, and a commander can leave a contact unassigned. The barrier filter (Section~\ref{sec:methods:dynamics}) steers each defender away from every contact except its own pursuit target, so the 12 losses of defenders to birds and civilian drones are consistent with allocation errors rather than accidental collisions.

We then disguised the intruders (Fig.~\ref{fig:airspace}b). On the same seeds and teams, each real intruder was shown to the LLMs as an image of a flapping-wing drone built from a preserved bird, making it a disguised intruder \cite{khan2023taxidermy,vosbein2026advancements}, and nothing else changed. 
In a text-based multi-agent LLM system, Papadopoulos et al.~\cite{papadopoulos2026mind} found that ideas which induce each agent that adopts them to pass them on can spread between agents, and that a brief warning in an agent's system prompt made agents nearly immune to them. Our deception instead arrives as an image, in a multimodal abstraction of a cyber-physical system, so in a second condition we tested whether a warning helps here too by adding one sentence of warning to the airspace paragraph (the prompts call intruders red): ``Some red intruders may be disguised as birds.''

With bird-drone images, every team deprioritized or dropped more of the intruders than it had with attack-drone images (Fig.~\ref{fig:airspace}d). The near-frontier model changed the most, because its teams had dropped few intruders before. At high authority it gave intruders a pursuit multiplier of zero in 59\% of its answers, against 1\% before, and as commander it left intruders unassigned in 61\% of its decisions about them, against 16\%, counting one decision per intruder per epoch. Its two teams destroyed 4 and 2 of 20 intruders, against 20 and 14 before, and lost 2 and 2 defenders to civilian drones. Intruders often flew faster than any bird or civilian drone can, and the LLMs received each contact's velocity. In 4 of the five teams such fast intruders were deprioritized or dropped less often than slower ones, yet 10\% to 56\% of them still were.

The warning restored the near-frontier model but not the low-cost one. With it, the near-frontier consensus team at high authority and the near-frontier commander team dropped 0\% and 25\% of intruders, respectively, and each destroyed 14 of 20 intruders, whereas with the low-cost model the consensus team at low authority still deprioritized 56\% of intruders and the commander team still dropped 42\%.

\section{Discussion}
\label{sec:discussion}
We studied how a team of LLM agents allocates tasks in an adversarial perimeter-defense game designed as a simplified abstraction of a cyber-physical system in which a team of LLM-driven defenders competes with a team of LLM-driven intruders, and used it as a benchmark to compare a central LLM commander with decentralized LLM agents that either resolve their task allocation conflicts through a consensus auction or act independently. Rather than relying on pure LLM-to-LLM chat (which often suffers from coordination breakdowns), we demonstrate the value of pairing LLMs with structured an external conflict-resolution mechanisms. We varied the communication network, the authority of the LLM over the bids in decentralized teams, the model's capability, and tested multimodal capabilities of modern LLMs by introducing birds and drones in addition to the intruders and provided LLMs with an image for each object along with the text state.  The consensus team was breached least often: over the network conditions without falsified reports, intruders breached the region in 15\% of episodes against the consensus team, 32\% against the commander team and 29\% against the independent team. Burst message loss, falsified reports and limited radio range raised the commander team's breach rate by 10 to 18~pp, but the consensus team's by at most 5~pp. High authority raised the low-cost model's breach rate by 38~pp, whereas the near-frontier model at high authority was breached as often as the low-cost model at low authority. Intruders disguised as birds fooled both models, and a one-sentence forewarning restored only the near-frontier model. From the results, we learn three main lessons.

First, an LLM can make the tactical decisions when it sits above the fast low level controllers, though the communication with the LLM itself over an API is slow, and how much the LLM should decide should follow its capability. In our study, a call to LLM took a median of 2.0~sec with the low-cost model and 5.7~sec with the near-frontier model (Supplementary Table 13), whereas MPC and the barrier filter acted every 0.1~sec. We intentionally simplified the simulation further by pausing the game for every API call, but a deployed team would have to deal with the long delays of API times. Inference latency is a recognized obstacle to deploying foundation models on robots~\cite{firoozi2025foundation}, and our observations are consistent with it. Smaller, faster models are one direction: small language models (SLMs) have been argued to be powerful enough, and more economical, for many narrow, repetitive agentic tasks~\cite{belcak2025small}. However care must be taken as our authority results highlight the importance of model capability: given more authority over how the auction scores tasks, the low-cost model weakened detection and was breached far more often, whereas the near-frontier model performed as well as the low-cost model at low authority. In LLM-driven robots, filtering an LLM's proposed actions is one way to keep the robot within constraints~\cite{yang2024plug}, and constraining what the LLM may set was noted to make the system more stable at some cost in flexibility~\cite{yu2023language}. Our results suggest that a low-cost model may be better kept at low authority. How to balance expensive and low-cost models within one team, for example by reserving the more capable model for high-authority roles, is an interesting direction for future work. This question matters beyond our setting, as any organization deploying LLM agents at scale faces the same trade-off between cost and capability.

Second, the communication network matters, and an explicit protocol that resolves conflicts outside the LLMs made the team less sensitive to a degraded network and to false data injection. The auction removed the duplicate commitments that the independent team made in 61\% to 81\% of checks per seed, and the consensus team was breached least often. The commander team instead depended on the network for every order. Under limited radio range only 61\% of its task-changing orders arrived (Fig.~\ref{fig:range}c and Supplementary Fig. 4), and its breach rate rose the most, although its network stayed connected more often than those of the decentralized teams. A likely explanation is that a central allocator needs reports to reach it and orders to reach every agent~\cite{dias2006market}, whereas a cut-off defender of a decentralized team still acts on its own bids. In multi-robot planning, a decentralized LLM framework that coordinated through dialogue had the lowest success rate of four designs~\cite{chen2024scalable}, across agentic benchmarks independent agents amplified errors more than centrally coordinated ones~\cite{kim2025scaling}, and inter-agent misalignment is one of three categories of failure in multi-agent LLM systems~\cite{cemri2025why}. Our independent team is consistent with these observations, and our consensus team shows that a decentralized LLM team can do well when a protocol outside the LLMs resolves its conflicts. Under false data injection, the commander, which assigns every defender's task, accepted at least one falsified report in 55 of its 60 episodes. In text-based collaboration a hierarchical structure proved the most resilient to faulty agents~\cite{huang2024resilience}, but our attack differed: it shifted one intruder's reported position and left every LLM's instructions untouched. More realistic communication channels, tighter limits such as bandwidth, and attacks that reach the LLMs themselves, which can spread between LLM agents~\cite{papadopoulos2026mind}, are natural next tests.

Third, multimodal LLMs are easily fooled by intruders disguised as birds, and this matters because AI already informs safety-critical decisions.
Drones in Ukraine already use AI for target recognition~\cite{bondar2025ukraine}, while bird-like drones such as Dove~\cite{chen2018dove} and decoy drones such as Gerbera~\cite{mittal2026gerbera} are built to be mistaken for something else, which makes reliable recognition essential. Image classifiers have long been known to be fooled in the physical world, as when a 3D-printed turtle was consistently classified as a rifle~\cite{athalye2018synthesizing}. Our disguise needed no such optimization. The images of the taxidermy bird drones show that the birds are artificial: two of the four drones are held in a person's hand, one stands on an office desk and none is in flight. Yet both models often treated the disguised intruders as birds, and the near-frontier model, which had told birds and civilian drones apart from intruders and lost no defender to them, dropped more of the disguised intruders than any low-cost team dropped or deprioritized. In a deployed system, each such decision could become a breach. In a text-based multi-agent LLM system, a brief warning in an agent's system prompt made agents nearly immune to self-propagating ideas~\cite{papadopoulos2026mind}. Our observation is consistent with this for the near-frontier model, which chased the disguised intruders again after a one-sentence forewarning, but not for the low-cost model, whose consensus team at low authority and commander team still often left the disguised intruders unchased. The interface also mattered: at low authority a contact can only be given a low priority, which leaves its pursuit among a defender's options, so LLM planners that share airspace with non-hostile contacts need an explicit option to ignore one. Multimodal models that resist such deception, and defenses that do not depend on a forewarning that names it, remain open problems.

Our testbed is intentionally a simple abstraction of a cyber-physical system. Agents move as point masses in two dimensions and sense their opponents without noise. They exchange messages in rounds, see reference images rather than camera views, and wait for every LLM answer. Each defender also knows positions of its teammates exactly, even when the network does not connect them (the network carries the defenders' beliefs about intruders, the consensus team's auction messages, and the commander's orders, so the intruders that a cut-off defender sees do not reach its teammates). We kept these parts simple so the differences between teams come from how their LLMs coordinate rather than from errors in control or perception, and so that each factor can be varied while the rest stay fixed. Higher-fidelity dynamics, sensing, communication and imagery, real call latency and, finally, hardware are left to future work. The framework is therefore best understood as a benchmark for LLM-driven coordination of cyber-physical teams rather than as a deployable counter-drone controller. Moving forward, we expect LLM agents to take on more of the decisions of physical teams as faster and more capable models narrow the gap between how quickly they decide and how quickly the world changes. Much remains to be done, but given the pace of progress in agentic AI teams, we look at this future with a positive outlook.

\section{Methods}
\label{sec:methods}
This section describes the simulation environment and the experimental procedure. We first present the agent dynamics, the motion controllers, and the collision-avoidance constraints, followed by the belief model that each agent maintains about the opposing team. We then describe the communication network, including the degraded network conditions and the falsified-report attack. Next, we detail the three defender architectures: the consensus auction, with its task generation, bid formulation, and LLM interfaces; the independent agents; and the commander. This is followed by the intruder team and the prompts and language models used by all agents. Finally, we describe the experimental design, the outcome measures, and the statistical analysis.

\subsection{Game and arena}
\label{sec:methods:game}
\begin{figure}[ht]
\centering
\includegraphics[width=\linewidth]{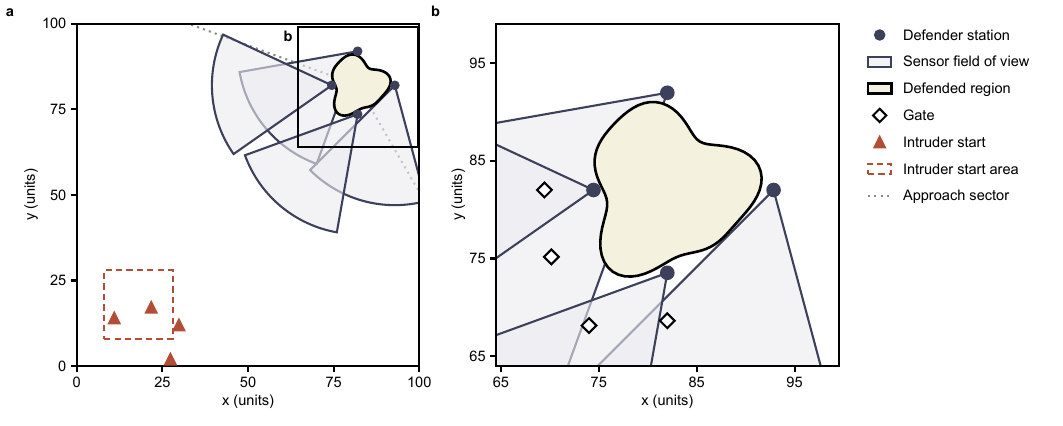}
\caption{\textbf{The arena, to scale.}
\textbf{(a)} The whole arena at the start of one episode with four intruders. Each intruder starts in the start area and is then rotated about the region's center by up to 10\textdegree{}; birds and civilian drones, when present, enter within the approach sector. Each defender has 60\textdegree{} FOV and a sensor range of 35 units. \textbf{(b)} The defended region enlarged, with the four defender stations and the four gates through which intruders attack. Distances are in the abstract units of the arena (One unit can be read as 10~m).}
\label{fig:arena}
\end{figure}

The arena is a 100 by 100 square in abstract units. Every length, speed and acceleration in the game is in these units and time is in seconds, so the game plays out the same at any length scale; read with one unit as 10~m, the arena is 1~km across, the sensor range is 350~m, and defenders and intruders fly at up to 20~m/sec along each axis. The defended region is a concave star-shaped polygon centered at (82,\,82) with a radius between 6 and 10 units, and four gates lie on its south-west side. Four defenders start near four stations placed on a polygon around the region. Two or four intruders start in a 20 by 20 box centered at (18,\,18), and each intruder's start is then rotated about the region's center by its own angle of up to 10\textdegree{} in either direction, so the intruders start about 90 units from the region (Fig.~\ref{fig:arena}). An episode lasts at most 400 time steps of $\Delta t = 0.1$\,sec, that is, 40~sec. A defender that comes within 0.4 units of an intruder intercepts it: the intruder is destroyed and the defender is disabled for the rest of the episode. When birds and civilian drones are present, a defender that comes within 0.4 units of either collides with it, with the same result. An intruder inside the region or on its boundary is a breach, and the breach ends the episode. An intruder that leaves the arena is removed. An episode also ends when every intruder has been destroyed or has left the arena. Furthermore, we divide the arena into a 4 by 4 grid of 16 coverage cells of 25 by 25 units, and a defender scouts a cell by flying to it with its sensor pointed towards the cell, and the sensor sweeps the ground ahead of it on its way and watches the cell once it arrives (Section~\ref{sec:methods:defender:bids}).

\subsection{Dynamics and hierarchical control}
\label{sec:methods:dynamics}
Every agent has the state $x = [p^\top, v^\top, \theta]^\top$, with planar position $p$, velocity $v$ and sensor heading $\theta$, and the input $u = [a^\top, \omega]^\top$, with translational acceleration $a$ and yaw rate $\omega$.
The state evolves by semi-implicit Euler integration,
\begin{equation}
p_{k+1} = p_k + \Delta t\, v_k + \Delta t^2 a_k, \qquad v_{k+1} = v_k + \Delta t\, a_k, \qquad \theta_{k+1} = \theta_k + \Delta t\, \omega_k ,
\end{equation}
with per-axis limits $|a_x|, |a_y| \le 2$, $|v_x|, |v_y| \le 2$ and $|\omega| \le 2\pi$\,rad\,sec$^{-1}$.
Each task (holding a station, flying to and watching a coverage cell (scouting), pursuing an intruder, flying to a gate, or evading a defender) is executed by a model predictive controller (MPC)~\cite{acharya2026capture} with a horizon of 60 steps (6\,sec). The MPC problems are formulated in CasADi~\cite{andersson2019casadi} and solved as quadratic programs with OSQP~\cite{stellato2020osqp}, except the evasion problem, which is solved as a nonlinear program with IPOPT~\cite{wachter2006ipopt}.

A pursuer's MPC keeps it close, over the whole horizon, to its target's trajectory predicted at constant velocity from the pursuer's latest belief about the target. The sensor heading follows fixed rules: a pursuer points at its target, a defender that sees an intruder tracks it, a defender on a station looks along that station's configured heading, and a scout looks toward its cell.

To avoid collisions, each agent minimally modifies the acceleration planned by MPC through a quadratic program with one constraint for each other agent that it avoids, based on the following barrier. For two agents at relative position $r$ and a minimum separation $D = 3$ units, we define $h = \lVert r \rVert^2 - D^2$ and impose a linear sufficient condition for
$h_{k+1} \ge (1-\gamma)\, h_k$,
with $\gamma = 0.05$. This barrier is a simplification suited to our double-integrator model; for more complex dynamics, control barrier functions (CBFs) can serve as the safety filter, which, under suitable assumptions, are necessary and sufficient for rendering the safe set forward invariant~\cite{ames2019control,dhiman2021control,cheng2020safe}.

In the barrier filter, two defenders each take on half of the barrier condition between them, disabled defenders are treated as immobile obstacles, intruders other than a defender's pursuit target are predicted at constant velocity, and a defender's barrier filter ignores the intruder it pursues. Intruders' barrier filters avoid only the other intruders, so that defenders can reach them. Intruders still evade approaching defenders, but through their planning layer rather than the barrier. If the quadratic program is infeasible, a relaxed problem with slack variables is solved, as in the barrier filter of Acharya et al.~\cite{acharya2026capture}, and the agent brakes only if that also fails or the velocity limits leave no admissible acceleration. Under the input limits, the barrier condition is not always satisfiable (Supplementary Fig. 14). One-step separation is therefore guaranteed only when the barrier filter's  quadratic program is feasible, and against intruders only under the constant-velocity prediction. As the lowest level of the hierarchical control architecture, the barrier filter uses the true positions and velocities of all other agents, whether or not they are inside the sensor FOV, as a stand-in for onboard proximity sensing.

\subsection{Sensing and beliefs}
\label{sec:methods:sensing}
Each agent $i$ keeps its own belief about the opposing team. At step $k$, this belief is a set of contact records $\mathcal{B}_i(k) = \{ b_{i,c} \}_{c \in \mathcal{C}_i(k)}$, where $\mathcal{C}_i(k)$ is the set of opposing agents that $i$ has seen or been told about. Each record
$b_{i,c} = \left( \hat{x}_c,\; t_c,\; o_c \right)$
holds the state $\hat{x}_c = [\hat{p}_c^\top, \hat{v}_c^\top]^\top$ of contact $c$ as observed, the step $t_c$ at which a sensor observed it, and the agent $o_c$ that made the observation. Sensing is noise-free: an agent observes the true state of every opposing agent within its sensor FOV (half-angle $30^\circ$) and sensor range (35 units) and nothing outside it. Agents use a record through its constant-velocity prediction
$\hat{p}_c(k) = \hat{p}_c + \hat{v}_c\,(k - t_c)\,\Delta t$,
so the age $k - t_c$ measures how out of date it is. When a record arrives from a teammate, the receiver keeps whichever record for that contact has the later observation step $t_c$, and keeps its current record when two observation steps are equal. Relaying does not change $t_c$, so a relayed record is as old as the original observation. Records are never discarded for being old, and not seeing a contact does not remove it. A contact is removed only when it is resolved: when an intruder is intercepted or leaves the arena, or a defender is disabled, every agent on the opposing team deletes that agent's record immediately and permanently. Defenders also keep a coverage belief $\kappa_i(k) \in \{-1, 0, 1, \dots\}^{16}$, which holds, for each of the 16 coverage cells (Section~\ref{sec:methods:game}), the last step at which a defender's sensor saw the cell's center ($-1$ if never); it is shared like the contact records and merged by taking the elementwise maximum. Teammates' states are not part of the belief: as a simplification, each defender knows its teammates' positions and sensor headings exactly under every network condition, and uses them to count active teammates, to compute which ground its teammates already watch (Section~\ref{sec:methods:defender:bids}) and, for the commander, to build its task catalog. We list this simplification among the limitations in the Discussion (Section~\ref{sec:discussion}). Intruders pool their observations over an ideal channel, whereas defenders share theirs over the simulated network, an asymmetry that is part of the design.

\subsection{Communication and network conditions}
\label{sec:methods:comms}
Defenders exchange beliefs, auction states, and commander assignments in synchronous communication rounds at intervals of 0.1\,sec. Links are two-way: in a round, the two defenders at the ends of a link receive each other's messages, and a loss erases the exchange in both directions. A message sent in one round is delivered in the next, so information traverses exactly one link per round, and each defender rebroadcasts its merged beliefs in every round. Upon receiving a record, a defender retains whichever record has the later observation time, keeping its existing record in the event of a tie. We consider five network conditions. The ideal network is complete and lossless but retains a latency of one round per link. Under independent loss, the exchange on each link is erased in each round with probability 0.3. Burst loss is modeled as a Gilbert--Elliott channel~\cite{gilbert1960burst,elliott1963burst} with transition probabilities of 0.15 from the good to the bad state and 0.30 from the bad to the good state, and loss probabilities of 0 and 0.9 in the good and bad states, respectively; the stationary mean loss is therefore also 0.3. Because every link is initialized in the good state, the realized loss varies across episodes. Loss draws are a deterministic function of the seed, the link, and the round, so matched episodes experience identical losses on identical links. Under limited radio range, only links shorter than 15 units are retained, and messages reach distant defenders solely through rebroadcast. We measure the realized connectivity as the fraction of communication rounds in which the active defenders form a single connected group over the links within radio range. Under message loss conditions this fraction is always 1, because a loss erases messages on a link that exists rather than removing the link itself.

In the falsified-report condition, a single defender is compromised. It is never defender 0, so that the commander of the commander team is never compromised, and for consistency the same rule holds in every team. It is drawn from a dedicated random stream so that matched episodes share the same compromised defender. From its first direct observation of an intruder until the end of the episode, the compromised defender displaces that intruder's position by 10 units toward the south-west in its outgoing first-hand reports; if it observes several intruders, it targets the one with the lowest identifier. This is a simple false data injection attack, in which an adversary alters the sensor data that a cyber-physical system relies on~\cite{miao2016coding}; unlike the stealthy attacks studied there, ours is not designed to evade detection. The intruder's identity, velocity, and observation time are left unaltered, as are the compromised defender's own stored beliefs. Recipients apply the standard merge rule, so the order of arrival determines whether a falsified or an honest copy of the same observation is retained.

\subsection{Defender teams}
\label{sec:methods:defender}

\subsubsection{Tasks and bids}
\label{sec:methods:defender:bids}
The consensus and independent teams allocate tasks through bids, which are scores that each defender assigns to its candidate tasks, and their LLMs influence the allocation by modifying these bids. The two teams differ in how bids become tasks: the consensus team resolves conflicts through an auction, whereas each independent-team defender takes the task with its own highest bid (Section~\ref{sec:methods:defender:bids:auction-ind}). The bid described below is a heuristic designed for this game, one reasonable choice among many rather than a fundamental or optimal rule. We constructed the bid from terms that each capture a single consideration a defender should weigh, and we justify each term alongside its definition. Default weights were set from the arena's geometry and from constructed situations on development seeds, never from the outcomes of study episodes. In these situations, the default weights led the auction to assign a pursuer to every known intruder, to keep the stations with the greatest coverage occupied and to send defenders at redundant stations to scout.

Each defender builds its own list of candidate tasks from its beliefs: pursuing a known intruder $j$, holding one of the four stations $s$, or scouting one of the 16 cells $g$ of a $4 \times 4$ coverage grid.
It assigns every task $q$ on this list a bid,
\begin{equation}\label{eq:bid}
b_q = V_q - C_q + \beta\,\mathbf{1}[q \in Q] ,
\end{equation}
where $V_q$ is the task's value, and $C_q$ its cost, both defined below, $\beta = 0.2$ is a constant called bonus, $Q$ is the set of up to three tasks that the defender's LLM has marked as preferred, and $\mathbf{1}[\cdot]$ is the indicator function.

\noindent Subtracting the cost from the value puts unlike tasks, such as a chase and a station, on one scale, so that a defender can rank them and the auction can compare defenders. The bonus lets the LLM decide between tasks whose bids lie within 0.2 of each other but cannot overturn larger differences. The value and the cost add up terms, each a quantity that the defender computes from its beliefs times a weight $w$. Each weight's subscript is a capital letter that indicates its term: $T$ threat, $E$ engagement, $H$ holding a station, $C$ coverage, $I$ information, $D$ distance, $R$ route (scouting travel), $A$ age, $X$ attrition, $L$ leaving, $W$ switching and $P$ pursuit continuity. The weights are the numbers that the LLM can change, as the two interfaces below describe. Supplementary Table~12 lists every symbol used in this section with its value or default and, for each weight, the values that the LLM can set.

\noindent The value depends on the kind of task,
\begin{equation}\label{eq:value}
V_q =
\begin{cases}
\rho_j \left( w_T + w_E\, \pi_j U_j O_j \right) & \text{pursue intruder } j,\\[2pt]
w_H + w_C\, m_s & \text{hold station } s,\\[2pt]
w_I\, I_g & \text{scout cell } g.
\end{cases}
\end{equation}
\noindent The cost also depends on the kind of task,
\begin{equation}\label{eq:cost}
C_q =
\begin{cases}
w_D\, d_q/L + w_A\, \alpha_j + w_X/(1+n) + K_q & \text{pursue intruder } j,\\[2pt]
w_D\, d_q/L + K_q & \text{hold station } s,\\[2pt]
w_R\, d_q/L + K_q & \text{scout cell } g.
\end{cases}
\end{equation}
where $K_q$ is a change cost,
\begin{equation}\label{eq:change}
K_q = w_L\, \ell_q\, m_0 + w_W\, \sigma_q + w_P\, c_q .
\end{equation}
\noindent We explain the terms of the value first and then those of the cost. A pursuit is worth a threat value $w_T$, the same for every intruder known to the defender, plus an engagement value $w_E$ times three factors. The threat value treats every known intruder as a threat, even one that is far away, moving away or hard to catch, because its LLM can turn it back to attack. It raises all of a defender's pursuits by the same amount, so it does not change which intruder the defender prefers or which defender is best placed for an intruder. It only lifts pursuits above stations and scouting. The engagement term then ranks intruders against each other. The first of the three factors is a priority $\pi_j$, which the LLM sets at low authority and which is otherwise 1.
The second is the intruder's urgency,
\begin{equation}\label{eq:urgency}
U_j = \max\!\left( \frac{1}{1 + \tau_j / T},\; \frac{1}{1 + \delta_j / L} \right),
\end{equation}
where $\tau_j$ is the forecast time for the intruder to reach the region and $\delta_j$ is its distance from the region, and the scales are $T = 60$\,sec and $L = 25$ units, a quarter of the arena.
The forecast uses the position and velocity in the defender's belief and is infinite if the intruder is not approaching, and the distance is measured from the intruder's predicted position. An intruder is therefore urgent when it is forecast to reach the region soon or is already close to it. We take the larger of the two terms because the forecast alone can be misleading. An intruder that dodges a defender near the region stops heading for it, so its forecast becomes infinite although it is still close. Each term falls smoothly from 1 and halves at its scale, so no threshold makes a bid jump.
The third factor is the opportunity $O_j$ to catch the intruder, because a defender should prefer intruders that it can reach before they breach. It is 1 if the defender can intercept intruder $j$ before its forecast breach and the end of the episode, and falls to 0.25 for an interception 10\,sec or more too late, or none.
It stays above zero because the forecast assumes that the intruder keeps its velocity, which its LLM may change. $O_j$ is averaged over 49 intruder velocities within 0.3 units\, per second per axis of the one in the defender's belief. The believed velocity is only an estimate, and without the average a small change in it could make an interception possible or impossible and make the bid jump. This opportunity is optimistic, because it assumes that the defender can fly anywhere its limits allow and it ignores the barrier filter and the intruder's evasion.
A pursuit multiplier $\rho_j$ scales the whole pursuit value, threat value included. The LLM sets it $\rho_j$ only at high authority, where it can make any intruder's pursuit more attractive, less attractive or worth nothing, and it is otherwise 1.

Defenders expect intruders from an approach sector, the bearings between 160\textdegree{} and 300\textdegree{} as seen from the region's center, a known approach corridor that is the same for every team. A station is worth a small fixed station value $w_H$ plus $w_C$ times its added coverage $m_s$. The added coverage is the area of the approach sector that the sensor of a defender at station $s$ would watch and no teammate currently watches, divided by the area of one sensor FOV. Counting only ground that no teammate watches pays a station for what it adds, so two defenders watching the same ground are not both rewarded for it. The small fixed value keeps holding a station worth something even where it adds little coverage. A cell is worth $w_I$ times its information $I_g$, the unwatched area of the approach sector that the defender's sensor would first sweep on a straight route to cell $g$, in the units of $m_s$. Scouting is worth what it may find, so each part of that area is weighted by how long it has gone unwatched according to the coverage belief $\kappa_i$, capped at 20\,sec. The area is also discounted with distance along the route, because ground swept sooner finds an intruder sooner, and without the discount the longest route, which sweeps the most, would always win.

Turning to the cost (Eq.~\ref{eq:cost}), every task costs travel in proportion to its distance $d_q$ from the defender, measured to an intruder's predicted position, a station or a cell center, with weight $w_D$ for pursuits and stations and $w_R$ for scouting. Travel makes the nearest suitable defender bid highest, so the auction tends to give each task to the defender best placed for it. Scouting travel weighs less, because on a scouting route the travel is itself the work. A pursuit costs more when the defender's belief about the intruder is old, by $w_A$ times the age term $\alpha_j$: the belief's age in seconds, times the intruders' per-axis speed limit of 2 units/sec, divided by $L$. The age term thus measures how far the intruder may have moved since it was last seen, as a fraction of $L$. A pursuit also risks the defender itself, because intercepting the intruder disables it, so it carries an attrition cost $w_X/(1+n)$, where $n$ is the number of active teammates. This cost grows as fewer teammates remain, because losing one of four defenders leaves the region better guarded than losing the last.

By default the value weights are $w_T = 1.2$, $w_E = 1$, $w_H = 0.05$, $w_C = 0.5$ and $w_I = 0.45$. The default cost weights are $w_D = 0.15$, $w_R = 0.05$, $w_A = 0.1$, $w_X = 0.5$, $w_L = 0.25$, $w_W = 0.05$ and $w_P = 0.15$. The threat value $w_T$ is the smallest multiple of 0.05 at which the auction assigned a pursuer to every known intruder in constructed situations with one or two intruders, including intruders far away and moving away. With these defaults the threat value alone, $w_T = 1.2$, is worth more than any station (whose value is at most 0.55). A station can outbid the pursuit of a known intruder only when the pursuit costs much more, for example when the intruder is far away. Several defenders that know of the same intruder can therefore all bid highest for it. The auction resolves this by giving each task to one defender only, so the others take their next-best tasks.

The change cost $K_q$ (Eq.~\ref{eq:change}) penalizes three kinds of change: leaving a useful station ($w_L$), switching tasks ($w_W$) and abandoning a pursuit ($w_P$). Because bids are recomputed every second (10 simulation steps) from updated beliefs, small fluctuations in those beliefs would otherwise cause defenders to oscillate between tasks without completing any of them. The leaving indicator $\ell_q$ equals 1 when task $q$ is not the station the defender currently holds (or its own station if it has no task). The leaving cost is scaled by the added coverage $m_0$ of that station, which is 0 while the defender is pursuing or scouting. A defender at a station that contributes little coverage therefore leaves more readily and, other things being equal, outbids a defender at a more valuable station for the same task. The switching indicator $\sigma_q$ equals 1 when the defender has a current task and task $q$ differs from it. The drop indicator $c_q$ equals 1 when task $q$ would abandon the pursuit of an intruder the defender still knows about, so that a pursuit already under way is worth completing. Its weight $w_P$ exceeds the largest change in a pursuit bid caused by velocity-belief errors of 0.1\,units/sec in constructed test situations, so such errors alone cannot terminate a pursuit. The drop indicator is reduced toward zero when the pursued intruder is moving away and task $q$ instead pursues another intruder that is forecast to breach within the episode and that the defender can intercept in time. This allows a defender to abandon a retreating intruder in favor of one that is about to breach.

\subsubsection{Auction and independent selection}
\label{sec:methods:defender:bids:auction-ind}
In the consensus team the bids enter a consensus-based auction \cite{choi2009consensus} run over the network. The auction restarts every epoch (every 10 communication rounds) and breaks ties in favor of the lower agent identifier. A defender adopts the auction's result once its own table has been unchanged for three rounds and keeps its previous commitment until then, and commitments persist across epochs. The independent team runs the same code with the auction messages switched off, and its prompt describes independent selection instead of the auction. Each defender then commits to its own highest bid, so two defenders may pursue the same intruder. The independent team therefore isolates the effect of the auction; it is not meant as the best possible policy for independent agents.

\subsubsection{Commander}
\label{sec:methods:defender:bids:commander}
Unlike the consensus and independent teams, the commander team needs a central node: the LLM of defender 0 assigns one task to every active defender each epoch. If defender 0 is disabled, the lowest-index active defender takes over and plans from its own beliefs. The commander chooses from a catalog of candidate tasks that the same rules build from its own beliefs. It sees each intruder's forecast status (approaching, moving away, crossing or stationary) and time to breach, each defender's optimistic interception time and its last issued assignment. A duplicate or invalid entry sends the affected defender to its own station. The commander's own order applies at once.
Every other order is broadcast, relayed by the other defenders and applied by its addressee if it is newer than the last order that defender accepted, and an addressee with no belief about the ordered intruder holds its station until it has one.

\subsubsection{Low-authority LLM interface}
\label{sec:methods:defender:bids:low}
Authority applies only to the decentralized teams, because the commander's LLM assigns every task directly. The independent team always runs at low authority, and the consensus team runs at low authority unless stated otherwise. High authority (Section~\ref{sec:methods:defender:bids:high}) only applies to the consensus team. Every number in a bid then comes from a formula fixed before any run, each answer is a choice among named options that can be read in the logs, and no answer can stop a known intruder from being pursued.
Once per epoch, each defender's LLM sees only that defender's beliefs and its task list, with every bid term computed. It can change the bids in three ways. First, it picks one of four weight profiles. The balanced profile uses the default weights, and the aggressive profile lowers $w_X$ to 0.2 and raises $T$ to 90\,sec. The conservative profile raises $w_C$ to 0.8, $w_L$ to 0.4 and $w_X$ to 0.9, and the scouting profile raises $w_I$ to 0.75.
Second, it gives each known intruder a priority $\pi_j$ of low, medium or high (0.5, 1 or 2). Third, it names up to three preferred tasks. It also gives one sentence of reasoning. The multiplier $\rho_j$ stays at 1, and no number written by the LLM enters a bid.
No profile changes $w_T$, and the priority multiplies only the urgency and opportunity term, so a known intruder's pursuit is always worth at least 1.2. A low priority makes a pursuit less attractive but cannot remove it. Invalid parts of an answer are dropped and logged, and a failed call leaves the defender's bids at their defaults.

\subsubsection{High-authority LLM interface}
\label{sec:methods:defender:bids:high}
At high authority, the consensus team's LLM can change much more.
Every epoch it sets each of the 12 weights within an allowed range (Supplementary Tables~7 and~12), and the five value weights may sum to at most 3.2, the sum of their default values.
Instead of a priority, it gives each known intruder a multiplier $\rho_j$ from 0 to 2, and the multipliers may sum to at most the number of known intruders.
It can also name up to three preferred tasks, while $\pi_j$ stays at 1 and $T$ at 60\,sec. An invalid or out-of-range value falls back to its default, and weights or multipliers over their budget are scaled down in proportion, as the prompt states. Each fallback and each scaling is logged. Because $\rho_j$ multiplies the whole pursuit value, threat value included, a multiplier of 0 leaves the pursuit with only its costs. High authority does not contain low authority, because the conservative and scouting profiles exceed the value budget, $T$ stays at 60\,sec and no multiplier reproduces a priority, which scales only part of the pursuit value. The bids then enter the same auction.

\subsection{Intruders}
\label{sec:methods:intruders}
The intruders are one team of independently instantiated LLM agents, the same for every defender team. Each intruder's LLM chooses to attack, by flying to one of the four gates and then to the region's center, or to evade a named defender that its beliefs place nearby.
All intruders' calls in an epoch use observations frozen before the first call, and an invalid answer keeps a still-valid decision or falls back to attacking. The intruders chose to attack in 73\% to 89\% of their answers across teams and intruder counts. Every intruder's LLM is the low-cost model, Gemini 3.1 Flash-Lite, in all experiments, including those where the defenders use the near-frontier model, so that every defender team faces the same opponent.

\subsection{Prompts and language models}
\label{sec:methods:prompts}
Each team receives a factual brief that states its own objective, the episode length, what happens on an interception or an intruder leaves the arena, and the end conditions. The defender brief is the same for every team. The prompts call the defenders blue and the intruders red. No prompt gives tactical advice, although the task catalog, the bid formula and the controllers encode domain choices. Every LLM observation in the main study is state-based: a text rendering of the agent's beliefs and of the options and values computed from them, with no images. Unless stated otherwise, every LLM agent is Gemini 3.1 Flash-Lite \cite{gemini31flashlite}, the low-cost model, called through OpenRouter at the Google AI Studio Flex tier with medium reasoning effort, temperature 1.0, an allowance of 8,000 output tokens and a strict JSON schema whose first field is the reasoning. The near-frontier model is GPT-6.1 Sol \cite{gpt61sol}, called at the standard tier with medium reasoning effort, no temperature parameter and an allowance of 25,000 output tokens. Calls are synchronous: the simulation waits for every answer of an epoch, so call latency does not affect the game (calls, latency and cost in Supplementary Table~13). Each response is stored under a hash of its full request, and each run records an execution identity that covers the code, the interfaces, the model profile and the settings. All interfaces and prompts are versioned and recorded with every run, and Supplementary Note~1 reproduces them.
Supplementary Note~2 explains both choices. The low-cost model came out of a selection study run on development seeds under a rule fixed before any answer was scored (Supplementary Table~15). For the near-frontier model we needed one clearly more capable than the low-cost model yet affordable in the settings where capability was in question, and GPT-6.1 Sol's developer describes it as close in performance to its most capable model at a lower cost. In a commander check run before the main study, all 180 answers of GPT-6.1 Sol passed, and the low-cost model passed at the medium reasoning effort used in the study but not at minimal effort (Supplementary Table~16).

\subsection{Airspace challenges}
\label{sec:methods:airspace}
In the main study, every LLM observation is text only, and every contact the defenders detect is an intruder. To test multimodal LLM decisions within the same simplified abstraction, we drew inspiration from a typical airspace, where birds and civilian drones share the sky with threats and telling drones apart from birds in video is hard enough to be the subject of a dedicated benchmark challenge~\cite{coluccia2021drone}. In our game a wrong call is expensive, because a defender that collides with any contact is disabled, so pursuing a bird or civilian drone costs the team a defender just as an interception does. Motion alone does not settle the question either, because our birds and civilian drones fly inbound like intruders, at up to the intruders' speed limit. Vision can help in this regard, and the LLMs we study support that capability. Both LLMs used in this study are multimodal: besides the state-based text they read in the main study, they accept images, so they can act as vision-language models that weigh a picture together with the state. We therefore built two challenges on top of the game. In the first, birds and civilian drones fly among the intruders, and each defender's LLM sees an image of every contact before it decides. In the second, we turn that ability against the defenders, and the intruders arrive disguised as birds. The disguise is not far-fetched, and it builds on existing work. Camouflage has been applied to fixed-wing, quadcopter and ornithopter drones~\cite{upshaw2023application}, and wildlife-monitoring drones have been built from taxidermy birds that fly by flapping their wings~\cite{khan2023taxidermy,vosbein2026advancements} and from a taxidermy duck that swims~\cite{vosbein2025swimming}. Both challenges use the ideal network with four intruders, on seeds 4001 to 4005, so that whatever changes comes from the airspace and not from the network.

\subsubsection{Defender observation space}
\label{sec:methods:arispace:see}
Each defender's LLM, including the commander's, receives a multimodal observation. It is the same state-based text as in the main study, followed by one image of every contact the LLM is asked about, each labeled ``image of contact\_id $N$''.
A defender's LLM is asked about every contact it holds a belief about, and the commander about the contacts in its catalog. Intruders are shown as images too, so a defender has to recognize threats as well as rule out harmless contacts. The intruders' own LLMs keep their state-based observations, as every LLM does in the main study. We chose the images so that recognition is honest rather than easy. They come from a set of nine, three per kind of contact, assigned the same way in every team. The intruders are two first-person-view attack drones and a commercial quadcopter carrying a grenade, so a consumer airframe alone does not mark a contact as civilian. The civilian drones are three consumer camera quadcopters, and the birds are a gull, a raven and a black kite. Each image is a public-domain or openly licensed picture of one object in flight, cropped to 4:3 at 640 by 480 pixels, with any readable text removed, so that no brand name or marking gives the answer away. Sources, licenses and the script that builds the images from their originals are in the code repository. Every defender and commander prompt also gains one factual paragraph, the airspace paragraph: not every contact is an intruder, birds and civilian drones share the airspace and cannot breach, a defender that collides with any contact destroys it and disables the defender, and a contact's identifier does not tell what it is. The paragraph gives no tactical advice, and no field was added in which the model must say what it thinks a contact is, so a model's judgment shows only in what it chooses to pursue. Supplementary Figs. 7--9, 12 and 13 show every image as the LLMs received it, with quotes from the reasoning of the commander's LLM, and Supplementary Figs. 5, 6, 10 and 11 show screenshots of example episodes; Supplementary Videos 1--3 show example episodes of the main study.

\subsubsection{Birds and civilian drones among the intruders}
\label{sec:methods:airspace:birds}
Each episode adds one bird and two civilian drones to the matching ideal-network episode of the main study. The intruders and defenders start exactly as in that episode, so every difference traces back to the newcomers. To be hard to rule out by motion, the birds and civilian drones behave like intruders on approach. They spawn 55 to 80 units from the region's center inside the approach sector and fly inbound through waypoints, the first 20 to 35 units from the center and later ones 20 to 70 units out, with no leg passing within 16 units of the center. They cruise at 0.6 to 1.0 times the intruders' speed limit, sense nothing, cannot breach and earn nothing, and a defender that collides with one is disabled and the contact destroyed. Five LLM teams ran on the five seeds (25 episodes).

\subsubsection{Intruders disguised as birds}
\label{sec:methods:airspace:disguised}
Distinguishing birds from drones might not be that hard of a challenge for a modern LLM, but making the distinction between a real bird and a disguised one might very well be. To test this, the five LLM teams ran again on the same seeds with the same starts, birds, civilian drones and images of them, but each real intruder was now shown as one of four images of taxidermy bird drones (Fig.~\ref{fig:airspace}b). The images show taxidermy bird drones from Dr. Hassanalian's group at New Mexico Tech, introduced above. They were chosen so that the drones are visibly artificial: two of the four drones are held in a person's hand, one stands on an office desk and none is in flight. The four images were assigned in an order that rotated from seed to seed, so every episode showed each of the four drones once. Only the picture changed, and the disguised intruders flew exactly as before. A cheap countermeasure has worked against manipulation between LLM agents before. Papadopoulos et al.~\cite{papadopoulos2026mind} showed that a one-paragraph warning in an agent's system prompt, written in a single attempt without tuning, made agents almost completely immune to adversarial ideas that spread through a multi-agent LLM system by inducing each agent that adopts them to pass them on. That manipulation traveled through text, whereas our deception arrives as a picture, so we probe whether the same defense carries over. Each team ran without and with one sentence added to the airspace paragraph after ``A contact's id does not tell what it is.'': ``Some red intruders may be disguised as birds.'' Nothing else in any prompt differed, and the prompt version recorded with every call names the condition. The two conditions ran with the same model snapshots and providers (50 episodes).

\subsection{Experimental design}
\label{sec:methods:exp_design}
The main study crossed three defender teams, two intruder counts and five network conditions on seeds 4001 to 4030, for 900 episodes.
Episodes that share a seed and an intruder count share their start positions, rotations, link-loss draws and compromised defender, because starts are drawn first and the other three use their own random streams, the method of common random numbers~\cite{glasserman1992crn}. Matched episodes diverge once the LLMs, sampled at temperature 1.0, answer differently. The consensus team at high authority ran every network condition and intruder count with the low-cost model, like a main-study team (300 episodes). With the near-frontier model it ran the ideal network with four intruders (30 episodes). The near-frontier commander ran the ideal network and falsified reports with both intruder counts (120 episodes), with everything but the commander's model taken from the commander team. Every prespecified episode ended in a valid record under its run's execution identity (Supplementary Table 14). In the main study, the 900 episodes made 160,788 calls with 0 transport failures, 43 provider retries all recovered, and in 7 of 10,657 commander epochs one defender fell back to its station.

\subsubsection{Metrics}
\label{sec:methods:exp_design:metrics}
The primary metric is the occurrence of a breach within the 40\,sec episode.  Duplicate commitments are evaluated at every epoch boundary after the first, and a duplicate is recorded at each boundary at which two or more active defenders are committed to the same task. To quantify pursuit coverage, at each time step we considered every active intruder contained in the belief set of at least one active defender and recorded whether any active defender was executing a pursuit of that intruder. The unpursued time is the proportion of these intruder-steps with no such pursuit, pooled within each seed. Intruders absent from every defender's belief set are excluded from this fraction and reported separately as the unseen time, a share of active-intruder time. Scouting before the first contact is defined as the fraction of defender-steps preceding the team's first intruder observation during which a defender was executing a scouting task. To quantify the influence of collision avoidance, a defender's control step is classified as modified when the barrier constraints altered its acceleration by more than 0.01 after the input limits had been applied. At the time limit, each intruder that remained active was classified on the basis of its final 10\,sec as being in a stand-off if it was evading on at least half of the steps and had a defender within 6 units on at least half of the steps; otherwise as closing if its distance to the defended region decreased by more than 1 unit; and otherwise as other.

\subsubsection{Statistics and reproducibility}
\label{sec:methods:exp_design:intervals}
We compare two conditions, for example the commander team and the consensus team, on the same random seeds.
A seed fixes the starting positions of all agents, so two episodes with the same seed and intruder count start from exactly the same situation. We call such two episodes a matched pair. For each pair we record whether each episode had a breach (1) or not (0). The pair's difference is +1 if only the first episode had a breach, \textminus1 if only the second did, and 0 if both or neither did. Episodes on the same seed are not independent of one another, so we do not count each episode as a separate data point. Instead we group the pairs by seed and average their differences, which gives one number per seed. For example, if on one seed the first team was breached in 3 of 8 episodes and the second team in 1 of 8, that seed's number is $(3-1)/8 = 0.25$. We report the mean of the 30 seed numbers, in percentage points (pp), as the difference in breach rate between the two conditions. We use the mean rather than the median: with a yes-or-no outcome and one or two pairs per seed, which most comparisons have, most seed numbers are zero, so the median is zero even when many more seeds differ in one direction than in the other. Its 95\% confidence interval (CI) is the range of true differences that the 30 seeds leave plausible. We compute it in the standard way as the mean plus or minus a multiplier times the standard error, where the multiplier comes from the $t$ distribution with 29 degrees of freedom because the spread is itself estimated from only 30 seeds. This is an approximation, since the seed numbers are bounded between \textminus1 and +1. Computing the interval over seeds rather than over episodes keeps it from being too narrow. A few comparisons have only one pair per seed, for example the near-frontier model at high authority on the ideal network with four intruders. There is then nothing to average, so each seed's number is simply that pair's difference, \textminus1, 0 or +1. For this case we use Newcombe's hybrid score interval, method 10~\cite{newcombe1998paired}, instead of the $t$ interval.

Before any outcome was run, we fixed four primary comparisons. Three compare teams: commander against consensus, commander against independent, and consensus against independent, each taken over the ideal network, both loss conditions and limited radio range with both intruder counts (4 conditions $\times$ 2 intruder counts = 8 pairs per seed). The fourth is the effect of falsified reports: each team with falsified reports against the same team on the ideal network, over all three teams and both intruder counts (3 teams $\times$ 2 intruder counts = 6 pairs per seed). In every comparison the difference is the first-named condition minus the second-named. Comparisons by intruder count, by team and by condition are secondary breakdowns of these, also fixed in advance, and are reported as estimates with CIs. Some comparisons were added during the study, after outcomes had been seen, and are therefore exploratory: the difference in unpursued time between the consensus and independent teams, three comparisons of the effect of falsified reports (four rows), and high against low authority in the consensus team together with its falsified-report difference (two rows). We report all results, including the primary comparisons, as estimates with CIs, and Supplementary Data 1 lists every paired comparison of breach rate with its per-seed values. Each result is read from its mean, the CI shows how precisely the 30 seeds estimate it. The runs with birds, civilian drones and disguised intruders (five seeds each) are descriptive, and their counts are reported without CIs.

\subsection{Use of AI assistance}
An AI assistant (Claude, Anthropic) was used with code assistance and polishing the writing. The authors take full responsibility for the content.

\subsection{Code availability}
\label{sec:methods:code}
The simulation, the experiment runners and the analysis code are available on GitHub at \url{https://github.com/the-kas-lab/consensus-within-competition-between}, and the version used for this paper is archived in the Zenodo repository, \url{https://doi.org/10.5281/zenodo.24209157}~\cite{abaidullah2026code}.

\section{Data availability}
\label{sec:methods:data}
The data generated and analysed in this study are available in the Zenodo repository, \url{https://doi.org/10.5281/zenodo.24208099}~\cite{abaidullah2026data}. They comprise the episode records of every run, which hold each episode's configuration and outcome; the cached model responses, with which every episode can be re-simulated step by step without calling a model; and the outputs of the analysis.

\bmhead{Acknowledgements}
This work was supported in part by Gleason Endowment at RIT. The funder played no role in study design, data collection, analysis and interpretation of data, or the writing of this manuscript.

\bmhead{Author contributions}
M.J.K. conceived the idea of formalizing consensus for agentic AI teams and its application to perimeter defense, as well as the multimodal airspace challenges, including intruders disguised as bird-like drones. K.A.U. conceived the LLM-authority concept, designed the decentralized testbed, including its communication and network model and the consensus-based conflict resolution, wrote the simulation and analysis code, ran all experiments, analyzed the results, and wrote the manuscript. A.A. contributed to the model predictive control and control design, helped refine and validate the study’s ideas, and contributed to writing the manuscript. M.J.K. supervised the project, acquired funding, and reviewed and edited the manuscript. All authors read and approved the final manuscript.

\bmhead{Competing interests}
All authors declare no financial or non-financial competing interests.

\bibliography{references}


\begin{thebibliography}{65}
\ifx \bisbn   \undefined \def \bisbn  #1{ISBN #1}\fi
\ifx \binits  \undefined \def \binits#1{#1}\fi
\ifx \bauthor  \undefined \def \bauthor#1{#1}\fi
\ifx \batitle  \undefined \def \batitle#1{#1}\fi
\ifx \bjtitle  \undefined \def \bjtitle#1{#1}\fi
\ifx \bvolume  \undefined \def \bvolume#1{\textbf{#1}}\fi
\ifx \byear  \undefined \def \byear#1{#1}\fi
\ifx \bissue  \undefined \def \bissue#1{#1}\fi
\ifx \bfpage  \undefined \def \bfpage#1{#1}\fi
\ifx \blpage  \undefined \def \blpage #1{#1}\fi
\ifx \burl  \undefined \def \burl#1{\textsf{#1}}\fi
\ifx \doiurl  \undefined \def \doiurl#1{\url{https://doi.org/#1}}\fi
\ifx \betal  \undefined \def \betal{\textit{et al.}}\fi
\ifx \binstitute  \undefined \def \binstitute#1{#1}\fi
\ifx \binstitutionaled  \undefined \def \binstitutionaled#1{#1}\fi
\ifx \bctitle  \undefined \def \bctitle#1{#1}\fi
\ifx \beditor  \undefined \def \beditor#1{#1}\fi
\ifx \bpublisher  \undefined \def \bpublisher#1{#1}\fi
\ifx \bbtitle  \undefined \def \bbtitle#1{#1}\fi
\ifx \bedition  \undefined \def \bedition#1{#1}\fi
\ifx \bseriesno  \undefined \def \bseriesno#1{#1}\fi
\ifx \blocation  \undefined \def \blocation#1{#1}\fi
\ifx \bsertitle  \undefined \def \bsertitle#1{#1}\fi
\ifx \bsnm \undefined \def \bsnm#1{#1}\fi
\ifx \bsuffix \undefined \def \bsuffix#1{#1}\fi
\ifx \bparticle \undefined \def \bparticle#1{#1}\fi
\ifx \barticle \undefined \def \barticle#1{#1}\fi
\bibcommenthead
\ifx \bconfdate \undefined \def \bconfdate #1{#1}\fi
\ifx \botherref \undefined \def \botherref #1{#1}\fi
\ifx \url \undefined \def \url#1{\textsf{#1}}\fi
\ifx \bchapter \undefined \def \bchapter#1{#1}\fi
\ifx \bbook \undefined \def \bbook#1{#1}\fi
\ifx \bcomment \undefined \def \bcomment#1{#1}\fi
\ifx \oauthor \undefined \def \oauthor#1{#1}\fi
\ifx \citeauthoryear \undefined \def \citeauthoryear#1{#1}\fi
\ifx \endbibitem  \undefined \def \endbibitem {}\fi
\ifx \bconflocation  \undefined \def \bconflocation#1{#1}\fi
\ifx \arxivurl  \undefined \def \arxivurl#1{\textsf{#1}}\fi
\csname PreBibitemsHook\endcsname

\bibitem[\protect\citeauthoryear{Guo et~al.}{2024}]{guo2024large}
\begin{bchapter}
\bauthor{\bsnm{Guo}, \binits{T.}},
\bauthor{\bsnm{Chen}, \binits{X.}},
\bauthor{\bsnm{Wang}, \binits{Y.}},
\bauthor{\bsnm{Chang}, \binits{R.}},
\bauthor{\bsnm{Pei}, \binits{S.}},
\bauthor{\bsnm{Chawla}, \binits{N.V.}},
\bauthor{\bsnm{Wiest}, \binits{O.}},
\bauthor{\bsnm{Zhang}, \binits{X.}}:
\bctitle{Large language model based multi-agents: a survey of progress and
  challenges}.
In: \bbtitle{Proc. 33rd International Joint Conference on Artificial
  Intelligence (IJCAI)},
pp. \bfpage{8048}--\blpage{8057}
(\byear{2024}).
\doiurl{10.24963/ijcai.2024/890}
\end{bchapter}
\endbibitem

\bibitem[\protect\citeauthoryear{Shen et~al.}{2023}]{shen2023hugginggpt}
\begin{barticle}
\bauthor{\bsnm{Shen}, \binits{Y.}},
\bauthor{\bsnm{Song}, \binits{K.}},
\bauthor{\bsnm{Tan}, \binits{X.}},
\bauthor{\bsnm{Li}, \binits{D.}},
\bauthor{\bsnm{Lu}, \binits{W.}},
\bauthor{\bsnm{Zhuang}, \binits{Y.}}:
\batitle{Hugginggpt: Solving ai tasks with chatgpt and its friends in hugging
  face}.
\bjtitle{Advances in Neural Information Processing Systems}
\bvolume{36},
\bfpage{38154}--\blpage{38180}
(\byear{2023})
\end{barticle}
\endbibitem

\bibitem[\protect\citeauthoryear{Zhan et~al.}{2026}]{zhan2026agensh}
\begin{botherref}
\oauthor{\bsnm{Zhan}, \binits{Z.}},
\oauthor{\bsnm{Song}, \binits{T.}},
\oauthor{\bsnm{Dong}, \binits{L.}},
\oauthor{\bsnm{Huang}, \binits{S.}},
\oauthor{\bsnm{Lian}, \binits{J.}},
\oauthor{\bsnm{Xia}, \binits{Y.}},
\oauthor{\bsnm{Wei}, \binits{F.}}:
Agensh: Scaling organizational intelligence to 1,024 agents.
arXiv preprint arXiv:2609.26781
(2026)
\end{botherref}
\endbibitem

\bibitem[\protect\citeauthoryear{Hong et~al.}{2024}]{hong2024metagpt}
\begin{bchapter}
\bauthor{\bsnm{Hong}, \binits{S.}},
\bauthor{\bsnm{Zhuge}, \binits{M.}},
\bauthor{\bsnm{Chen}, \binits{J.}},
\bauthor{\bsnm{Zheng}, \binits{X.}},
\bauthor{\bsnm{Cheng}, \binits{Y.}},
\bauthor{\bsnm{Wang}, \binits{J.}},
\bauthor{\bsnm{Zhang}, \binits{C.}},
\bauthor{\bsnm{Yau}, \binits{S.}},
\bauthor{\bsnm{Lin}, \binits{Z.}},
\bauthor{\bsnm{Zhou}, \binits{L.}}, \betal:
\bctitle{Metagpt: Meta programming for a multi-agent collaborative framework}.
In: \bbtitle{International Conference on Learning Representations},
vol. \bseriesno{2024},
pp. \bfpage{23247}--\blpage{23275}
(\byear{2024})
\end{bchapter}
\endbibitem

\bibitem[\protect\citeauthoryear{Kannan et~al.}{2024}]{kannan2024smart}
\begin{bchapter}
\bauthor{\bsnm{Kannan}, \binits{S.S.}},
\bauthor{\bsnm{Venkatesh}, \binits{V.L.N.}},
\bauthor{\bsnm{Min}, \binits{B.-C.}}:
\bctitle{{SMART-LLM}: smart multi-agent robot task planning using large
  language models}.
In: \bbtitle{Proc. IEEE/RSJ International Conference on Intelligent Robots and
  Systems (IROS)},
pp. \bfpage{12140}--\blpage{12147}
(\byear{2024})
\end{bchapter}
\endbibitem

\bibitem[\protect\citeauthoryear{Mandi et~al.}{2024}]{mandi2024roco}
\begin{bchapter}
\bauthor{\bsnm{Mandi}, \binits{Z.}},
\bauthor{\bsnm{Jain}, \binits{S.}},
\bauthor{\bsnm{Song}, \binits{S.}}:
\bctitle{{RoCo}: dialectic multi-robot collaboration with large language
  models}.
In: \bbtitle{Proc. IEEE International Conference on Robotics and Automation
  (ICRA)},
pp. \bfpage{286}--\blpage{299}
(\byear{2024}).
\doiurl{10.1109/ICRA57147.2024.10610855}
\end{bchapter}
\endbibitem

\bibitem[\protect\citeauthoryear{Gerkey and
  Matari{\'c}}{2004}]{gerkey2004formal}
\begin{barticle}
\bauthor{\bsnm{Gerkey}, \binits{B.P.}},
\bauthor{\bsnm{Matari{\'c}}, \binits{M.J.}}:
\batitle{A formal analysis and taxonomy of task allocation in multi-robot
  systems}.
\bjtitle{International Journal of Robotics Research}
\bvolume{23}(\bissue{9}),
\bfpage{939}--\blpage{954}
(\byear{2004})
\doiurl{10.1177/0278364904045564}
\end{barticle}
\endbibitem

\bibitem[\protect\citeauthoryear{Yang et~al.}{2025}]{yang2025agentnet}
\begin{bchapter}
\bauthor{\bsnm{Yang}, \binits{Y.}},
\bauthor{\bsnm{Chai}, \binits{H.}},
\bauthor{\bsnm{Shao}, \binits{S.}},
\bauthor{\bsnm{Song}, \binits{Y.}},
\bauthor{\bsnm{Qi}, \binits{S.}},
\bauthor{\bsnm{Rui}, \binits{R.}},
\bauthor{\bsnm{Zhang}, \binits{W.}}:
\bctitle{{AgentNet}: decentralized evolutionary coordination for {LLM}-based
  multi-agent systems}.
In: \bbtitle{Advances in Neural Information Processing Systems (NeurIPS)}
(\byear{2025})
\end{bchapter}
\endbibitem

\bibitem[\protect\citeauthoryear{Tran et~al.}{2025}]{tran2025survey}
\begin{botherref}
\oauthor{\bsnm{Tran}, \binits{K.-T.}},
\oauthor{\bsnm{Dao}, \binits{D.}},
\oauthor{\bsnm{Nguyen}, \binits{M.-D.}},
\oauthor{\bsnm{Pham}, \binits{Q.-V.}},
\oauthor{\bsnm{O'Sullivan}, \binits{B.}},
\oauthor{\bsnm{Nguyen}, \binits{H.D.}}:
Multi-agent collaboration mechanisms: a survey of {LLMs}.
Preprint at https://arxiv.org/abs/2501.06322
(2025)
\end{botherref}
\endbibitem

\bibitem[\protect\citeauthoryear{Chen et~al.}{2024}]{chen2024scalable}
\begin{bchapter}
\bauthor{\bsnm{Chen}, \binits{Y.}},
\bauthor{\bsnm{Arkin}, \binits{J.}},
\bauthor{\bsnm{Zhang}, \binits{Y.}},
\bauthor{\bsnm{Roy}, \binits{N.}},
\bauthor{\bsnm{Fan}, \binits{C.}}:
\bctitle{Scalable multi-robot collaboration with large language models:
  centralized or decentralized systems?}
In: \bbtitle{Proc. IEEE International Conference on Robotics and Automation
  (ICRA)}
(\byear{2024})
\end{bchapter}
\endbibitem

\bibitem[\protect\citeauthoryear{Kim et~al.}{2025}]{kim2025scaling}
\begin{botherref}
\oauthor{\bsnm{Kim}, \binits{Y.}},
\oauthor{\bsnm{Gu}, \binits{K.}},
\oauthor{\bsnm{Park}, \binits{C.}},
\oauthor{\bsnm{Park}, \binits{C.}},
\oauthor{\bsnm{Schmidgall}, \binits{S.}},
\oauthor{\bsnm{Heydari}, \binits{A.A.}},
\oauthor{\bsnm{Yan}, \binits{Y.}},
\oauthor{\bsnm{Zhang}, \binits{Z.}},
\oauthor{\bsnm{Zhuang}, \binits{Y.}},
\oauthor{\bsnm{Liu}, \binits{Y.}},
\oauthor{\bsnm{Malhotra}, \binits{M.}},
\oauthor{\bsnm{Liang}, \binits{P.P.}},
\oauthor{\bsnm{Park}, \binits{H.W.}},
\oauthor{\bsnm{Yang}, \binits{Y.}},
\oauthor{\bsnm{Xu}, \binits{X.}},
\oauthor{\bsnm{Du}, \binits{Y.}},
\oauthor{\bsnm{Patel}, \binits{S.}},
\oauthor{\bsnm{Althoff}, \binits{T.}},
\oauthor{\bsnm{McDuff}, \binits{D.}},
\oauthor{\bsnm{Liu}, \binits{X.}}:
Towards a science of scaling agent systems.
Preprint at https://arxiv.org/abs/2512.08296
(2025)
\end{botherref}
\endbibitem

\bibitem[\protect\citeauthoryear{Shishika and Kumar}{2020}]{shishika2020review}
\begin{bchapter}
\bauthor{\bsnm{Shishika}, \binits{D.}},
\bauthor{\bsnm{Kumar}, \binits{V.}}:
\bctitle{A review of multi agent perimeter defense games}.
In: \bbtitle{Decision and Game Theory for Security (GameSec 2020)},
pp. \bfpage{472}--\blpage{485}.
\bpublisher{Springer}, \blocation{???}
(\byear{2020})
\end{bchapter}
\endbibitem

\bibitem[\protect\citeauthoryear{Shishika
  et~al.}{2020}]{shishika2020cooperative}
\begin{barticle}
\bauthor{\bsnm{Shishika}, \binits{D.}},
\bauthor{\bsnm{Paulos}, \binits{J.}},
\bauthor{\bsnm{Kumar}, \binits{V.}}:
\batitle{Cooperative team strategies for multi-player perimeter-defense games}.
\bjtitle{IEEE Robotics and Automation Letters}
\bvolume{5}(\bissue{2}),
\bfpage{2738}--\blpage{2745}
(\byear{2020})
\end{barticle}
\endbibitem

\bibitem[\protect\citeauthoryear{Chung et~al.}{2011}]{chung2011search}
\begin{barticle}
\bauthor{\bsnm{Chung}, \binits{T.H.}},
\bauthor{\bsnm{Hollinger}, \binits{G.A.}},
\bauthor{\bsnm{Isler}, \binits{V.}}:
\batitle{Search and pursuit-evasion in mobile robotics: A survey}.
\bjtitle{Autonomous robots}
\bvolume{31}(\bissue{4}),
\bfpage{299}--\blpage{316}
(\byear{2011})
\end{barticle}
\endbibitem

\bibitem[\protect\citeauthoryear{Weintraub
  et~al.}{2020}]{weintraub2020introduction}
\begin{bchapter}
\bauthor{\bsnm{Weintraub}, \binits{I.E.}},
\bauthor{\bsnm{Pachter}, \binits{M.}},
\bauthor{\bsnm{Garcia}, \binits{E.}}:
\bctitle{An introduction to pursuit-evasion differential games}.
In: \bbtitle{2020 American Control Conference (ACC)},
pp. \bfpage{1049}--\blpage{1066}
(\byear{2020}).
\bcomment{IEEE}
\end{bchapter}
\endbibitem

\bibitem[\protect\citeauthoryear{Weintraub et~al.}{2023}]{weintraub2023range}
\begin{bchapter}
\bauthor{\bsnm{Weintraub}, \binits{I.E.}},
\bauthor{\bsnm{Von~Moll}, \binits{A.}},
\bauthor{\bsnm{Pachter}, \binits{M.}}:
\bctitle{Range-limited pursuit-evasion}.
In: \bbtitle{NAECON 2023-IEEE National Aerospace and Electronics Conference},
pp. \bfpage{28}--\blpage{35}
(\byear{2023}).
\bcomment{IEEE}
\end{bchapter}
\endbibitem

\bibitem[\protect\citeauthoryear{Zhou et~al.}{2025}]{zhou2025control}
\begin{bchapter}
\bauthor{\bsnm{Zhou}, \binits{M.}},
\bauthor{\bsnm{Shaikh}, \binits{M.}},
\bauthor{\bsnm{Chaubey}, \binits{V.}},
\bauthor{\bsnm{Haggerty}, \binits{P.}},
\bauthor{\bsnm{Koga}, \binits{S.}},
\bauthor{\bsnm{Panagou}, \binits{D.}},
\bauthor{\bsnm{Atanasov}, \binits{N.}}:
\bctitle{Control strategies for pursuit-evasion under occlusion using
  visibility and safety barrier functions}.
In: \bbtitle{2025 IEEE International Conference on Robotics and Automation
  (ICRA)},
pp. \bfpage{12863}--\blpage{12869}
(\byear{2025}).
\bcomment{IEEE}
\end{bchapter}
\endbibitem

\bibitem[\protect\citeauthoryear{Lee et~al.}{2023}]{lee2023gnn}
\begin{barticle}
\bauthor{\bsnm{Lee}, \binits{E.S.}},
\bauthor{\bsnm{Zhou}, \binits{L.}},
\bauthor{\bsnm{Ribeiro}, \binits{A.}},
\bauthor{\bsnm{Kumar}, \binits{V.}}:
\batitle{Graph neural networks for decentralized multi-agent perimeter
  defense}.
\bjtitle{Frontiers in Control Engineering}
\bvolume{4},
\bfpage{1104745}
(\byear{2023})
\doiurl{10.3389/fcteg.2023.1104745}
\end{barticle}
\endbibitem

\bibitem[\protect\citeauthoryear{MacDonald}{2025}]{macdonald2025swarms}
\begin{botherref}
\oauthor{\bsnm{MacDonald}, \binits{A.}}:
{AI}-Powered Drone Swarms Have Now Entered the Battlefield.
The Wall Street Journal.
Published September 2025; accessed 11 November 2026
(2025)
\end{botherref}
\endbibitem

\bibitem[\protect\citeauthoryear{Park et~al.}{2021}]{park2021antidrone}
\begin{barticle}
\bauthor{\bsnm{Park}, \binits{S.}},
\bauthor{\bsnm{Kim}, \binits{H.T.}},
\bauthor{\bsnm{Lee}, \binits{S.}},
\bauthor{\bsnm{Joo}, \binits{H.}},
\bauthor{\bsnm{Kim}, \binits{H.}}:
\batitle{Survey on anti-drone systems: components, designs, and challenges}.
\bjtitle{IEEE Access}
\bvolume{9},
\bfpage{42635}--\blpage{42659}
(\byear{2021})
\doiurl{10.1109/ACCESS.2021.3065926}
\end{barticle}
\endbibitem

\bibitem[\protect\citeauthoryear{Maity and Pourghorban}{2024}]{maity2024target}
\begin{botherref}
\oauthor{\bsnm{Maity}, \binits{D.}},
\oauthor{\bsnm{Pourghorban}, \binits{A.}}:
Cooperative target defense under communication and sensing constraints.
Preprint at https://arxiv.org/abs/2412.09939
(2024)
\end{botherref}
\endbibitem

\bibitem[\protect\citeauthoryear{Otte et~al.}{2020}]{otte2020auctions}
\begin{barticle}
\bauthor{\bsnm{Otte}, \binits{M.}},
\bauthor{\bsnm{Kuhlman}, \binits{M.J.}},
\bauthor{\bsnm{Sofge}, \binits{D.}}:
\batitle{Auctions for multi-robot task allocation in communication limited
  environments}.
\bjtitle{Autonomous Robots}
\bvolume{44},
\bfpage{547}--\blpage{584}
(\byear{2020})
\end{barticle}
\endbibitem

\bibitem[\protect\citeauthoryear{Bommasani
  et~al.}{2021}]{bommasani2021opportunities}
\begin{botherref}
\oauthor{\bsnm{Bommasani}, \binits{R.}},
\oauthor{\bsnm{Hudson}, \binits{D.A.}},
\oauthor{\bsnm{Adeli}, \binits{E.}},
\oauthor{\bsnm{Altman}, \binits{R.}},
\oauthor{\bsnm{Arora}, \binits{S.}},
\oauthor{\bsnm{Arx}, \binits{S.}},
\oauthor{\bsnm{Bernstein}, \binits{M.S.}},
\oauthor{\bsnm{Bohg}, \binits{J.}},
\oauthor{\bsnm{Bosselut}, \binits{A.}},
\oauthor{\bsnm{Brunskill}, \binits{E.}}, et al.:
On the opportunities and risks of foundation models.
arXiv preprint arXiv:2108.07258
(2021)
\end{botherref}
\endbibitem

\bibitem[\protect\citeauthoryear{Firoozi et~al.}{2025}]{firoozi2025foundation}
\begin{barticle}
\bauthor{\bsnm{Firoozi}, \binits{R.}}, \betal:
\batitle{Foundation models in robotics: applications, challenges, and the
  future}.
\bjtitle{International Journal of Robotics Research}
\bvolume{44}(\bissue{5}),
\bfpage{701}--\blpage{739}
(\byear{2025})
\doiurl{10.1177/02783649241281508}
\end{barticle}
\endbibitem

\bibitem[\protect\citeauthoryear{Hu et~al.}{2023}]{hu2023toward}
\begin{botherref}
\oauthor{\bsnm{Hu}, \binits{Y.}},
\oauthor{\bsnm{Xie}, \binits{Q.}},
\oauthor{\bsnm{Jain}, \binits{V.}},
\oauthor{\bsnm{Francis}, \binits{J.}},
\oauthor{\bsnm{Patrikar}, \binits{J.}},
\oauthor{\bsnm{Keetha}, \binits{N.}},
\oauthor{\bsnm{Kim}, \binits{S.}},
\oauthor{\bsnm{Xie}, \binits{Y.}},
\oauthor{\bsnm{Zhang}, \binits{T.}},
\oauthor{\bsnm{Fang}, \binits{H.-S.}}, et al.:
Toward general-purpose robots via foundation models: A survey and
  meta-analysis.
arXiv preprint arXiv:2312.08782
(2023)
\end{botherref}
\endbibitem

\bibitem[\protect\citeauthoryear{Ichter et~al.}{2023}]{ahn2022can}
\begin{bchapter}
\bauthor{\bsnm{Ichter}, \binits{B.}}, \betal:
\bctitle{Do as {I} can, not as {I} say: grounding language in robotic
  affordances}.
In: \bbtitle{Proc. 6th Conference on Robot Learning (CoRL), Proceedings of
  Machine Learning Research 205},
pp. \bfpage{287}--\blpage{318}
(\byear{2023})
\end{bchapter}
\endbibitem

\bibitem[\protect\citeauthoryear{Acharya et~al.}{2026}]{acharya2026capture}
\begin{botherref}
\oauthor{\bsnm{Acharya}, \binits{A.}},
\oauthor{\bsnm{Goyette}, \binits{T.}},
\oauthor{\bsnm{Ataei}, \binits{M.}},
\oauthor{\bsnm{Stoica}, \binits{A.}},
\oauthor{\bsnm{Dhiman}, \binits{V.}},
\oauthor{\bsnm{Khojasteh}, \binits{M.J.}}:
Engagement-aware agentic pursuit-evasion.
Preprint at https://arxiv.org/abs/2607.10986
(2026)
\end{botherref}
\endbibitem

\bibitem[\protect\citeauthoryear{Morari et~al.}{2015}]{morari2015predictive}
\begin{bbook}
\bauthor{\bsnm{Morari}, \binits{M.}},
\bauthor{\bsnm{Borrelli}, \binits{F.}},
\bauthor{\bsnm{Bemporad}, \binits{A.}}:
\bbtitle{Predictive Control for Linear and Hybrid Systems}.
\bpublisher{Cambridge Univ. Press, Cambridge, England, UK}, \blocation{???}
(\byear{2015})
\end{bbook}
\endbibitem

\bibitem[\protect\citeauthoryear{Mayne et~al.}{2000}]{mayne2000constrained}
\begin{barticle}
\bauthor{\bsnm{Mayne}, \binits{D.Q.}},
\bauthor{\bsnm{Rawlings}, \binits{J.B.}},
\bauthor{\bsnm{Rao}, \binits{C.V.}},
\bauthor{\bsnm{Scokaert}, \binits{P.O.M.}}:
\batitle{Constrained model predictive control: Stability and optimality}.
\bjtitle{Automatica}
\bvolume{36}(\bissue{6}),
\bfpage{789}--\blpage{814}
(\byear{2000})
\doiurl{10.1016/s0005-1098(99)00214-9}
\end{barticle}
\endbibitem

\bibitem[\protect\citeauthoryear{Dias et~al.}{2006}]{dias2006market}
\begin{barticle}
\bauthor{\bsnm{Dias}, \binits{M.B.}},
\bauthor{\bsnm{Zlot}, \binits{R.}},
\bauthor{\bsnm{Kalra}, \binits{N.}},
\bauthor{\bsnm{Stentz}, \binits{A.}}:
\batitle{Market-based multirobot coordination: a survey and analysis}.
\bjtitle{Proceedings of the IEEE}
\bvolume{94}(\bissue{7}),
\bfpage{1257}--\blpage{1270}
(\byear{2006})
\end{barticle}
\endbibitem

\bibitem[\protect\citeauthoryear{Choi et~al.}{2009}]{choi2009consensus}
\begin{barticle}
\bauthor{\bsnm{Choi}, \binits{H.-L.}},
\bauthor{\bsnm{Brunet}, \binits{L.}},
\bauthor{\bsnm{How}, \binits{J.P.}}:
\batitle{Consensus-based decentralized auctions for robust task allocation}.
\bjtitle{IEEE Transactions on Robotics}
\bvolume{25}(\bissue{4}),
\bfpage{912}--\blpage{926}
(\byear{2009})
\doiurl{10.1109/TRO.2009.2022423}
\end{barticle}
\endbibitem

\bibitem[\protect\citeauthoryear{Lott and Honary}{2026}]{lott2026decentralized}
\begin{botherref}
\oauthor{\bsnm{Lott}, \binits{J.}},
\oauthor{\bsnm{Honary}, \binits{V.}}:
Decentralized multi-robot task allocation under degraded communication: a
  benchmark of performance, reliability, and computation.
Preprint at https://arxiv.org/abs/2609.13711
(2026)
\end{botherref}
\endbibitem

\bibitem[\protect\citeauthoryear{Lee et~al.}{2025}]{lee2025prompt}
\begin{bchapter}
\bauthor{\bsnm{Lee}, \binits{D.}},
\bauthor{\bsnm{Tiwari}, \binits{M.}},
\bauthor{\bsnm{Miranda}, \binits{B.}}:
\bctitle{Prompt infection: Llm-to-llm prompt injection within multi-agent
  systems}.
In: \bbtitle{European Symposium on Research in Computer Security},
pp. \bfpage{511}--\blpage{520}
(\byear{2025}).
\bcomment{Springer}
\end{bchapter}
\endbibitem

\bibitem[\protect\citeauthoryear{Gu et~al.}{2024}]{gu2024agent}
\begin{botherref}
\oauthor{\bsnm{Gu}, \binits{X.}},
\oauthor{\bsnm{Zheng}, \binits{X.}},
\oauthor{\bsnm{Pang}, \binits{T.}},
\oauthor{\bsnm{Du}, \binits{C.}},
\oauthor{\bsnm{Liu}, \binits{Q.}},
\oauthor{\bsnm{Wang}, \binits{Y.}},
\oauthor{\bsnm{Jiang}, \binits{J.}},
\oauthor{\bsnm{Lin}, \binits{M.}}:
Agent smith: A single image can jailbreak one million multimodal llm agents
  exponentially fast.
arXiv preprint arXiv:2402.08567
(2024)
\end{botherref}
\endbibitem

\bibitem[\protect\citeauthoryear{Mittal}{2026}]{mittal2026gerbera}
\begin{botherref}
\oauthor{\bsnm{Mittal}, \binits{V.}}:
The {R}ussian {G}erbera Drone Is Evolving Beyond Being A Simple Decoy.
Forbes.
Published 2 August 2026
(2026).
\url{https://www.forbes.com/sites/vikrammittal/2026/08/02/the-russian-gerbera-drone-is-evolving-beyond-being-a-simple-decoy/}
\end{botherref}
\endbibitem

\bibitem[\protect\citeauthoryear{Athalye
  et~al.}{2018}]{athalye2018synthesizing}
\begin{bchapter}
\bauthor{\bsnm{Athalye}, \binits{A.}},
\bauthor{\bsnm{Engstrom}, \binits{L.}},
\bauthor{\bsnm{Ilyas}, \binits{A.}},
\bauthor{\bsnm{Kwok}, \binits{K.}}:
\bctitle{Synthesizing robust adversarial examples}.
In: \bbtitle{International Conference on Machine Learning},
pp. \bfpage{284}--\blpage{293}
(\byear{2018}).
\bcomment{PMLR}
\end{bchapter}
\endbibitem

\bibitem[\protect\citeauthoryear{Khan et~al.}{2023}]{khan2023taxidermy}
\begin{bchapter}
\bauthor{\bsnm{Khan}, \binits{M.M.}},
\bauthor{\bsnm{Gee}, \binits{P.}},
\bauthor{\bsnm{Upshaw}, \binits{J.}},
\bauthor{\bsnm{Hassanalian}, \binits{M.}}:
\bctitle{Taxidermy birds as platform for flapping wing drones: a bioinspired
  mechanism for wildlife monitoring}.
In: \bbtitle{AIAA SCITECH 2023 Forum},
p. \bfpage{0836}
(\byear{2023})
\end{bchapter}
\endbibitem

\bibitem[\protect\citeauthoryear{Vosbein
  et~al.}{2026}]{vosbein2026advancements}
\begin{bchapter}
\bauthor{\bsnm{Vosbein}, \binits{D.}},
\bauthor{\bsnm{Upshaw}, \binits{J.}},
\bauthor{\bsnm{Maimako}, \binits{S.}},
\bauthor{\bsnm{Hassanalian}, \binits{M.}}:
\bctitle{Advancements in flapping wing drone technology and wildlife
  observation via taxidermy}.
In: \bbtitle{AIAA SCITECH 2026 Forum},
p. \bfpage{2082}
(\byear{2026})
\end{bchapter}
\endbibitem

\bibitem[\protect\citeauthoryear{Bondar}{2025}]{bondar2025ukraine}
\begin{botherref}
\oauthor{\bsnm{Bondar}, \binits{K.}}:
Ukraine's future vision and current capabilities for waging {AI}-enabled
  autonomous warfare.
Technical report,
Center for Strategic and International Studies (CSIS)
(March 2025).
\url{https://www.csis.org/analysis/ukraines-future-vision-and-current-capabilities-waging-ai-enabled-autonomous-warfare}
\end{botherref}
\endbibitem

\bibitem[\protect\citeauthoryear{Lillis and Cohen}{2026}]{cnn2026aierror}
\begin{botherref}
\oauthor{\bsnm{Lillis}, \binits{K.B.}},
\oauthor{\bsnm{Cohen}, \binits{Z.}}:
{US} military had close call after using {AI} for false intelligence report,
  sources say.
CNN.
Published 18 September 2026
(2026).
\url{https://www.cnn.com/2026/09/18/politics/us-military-ai-false-intelligence-china-ship}
\end{botherref}
\endbibitem

\bibitem[\protect\citeauthoryear{Papadopoulos
  et~al.}{2026}]{papadopoulos2026mind}
\begin{botherref}
\oauthor{\bsnm{Papadopoulos}, \binits{V.}},
\oauthor{\bsnm{Shah}, \binits{M.}},
\oauthor{\bsnm{Zimmerman}, \binits{S.}},
\oauthor{\bsnm{Lindsey}, \binits{J.}}:
Mind viruses: Self-propagating ideas in multi-agent llm systems.
arXiv preprint arXiv:2608.10218
(2026)
\end{botherref}
\endbibitem

\bibitem[\protect\citeauthoryear{Belcak et~al.}{2025}]{belcak2025small}
\begin{botherref}
\oauthor{\bsnm{Belcak}, \binits{P.}},
\oauthor{\bsnm{Heinrich}, \binits{G.}},
\oauthor{\bsnm{Diao}, \binits{S.}},
\oauthor{\bsnm{Fu}, \binits{Y.}},
\oauthor{\bsnm{Dong}, \binits{X.}},
\oauthor{\bsnm{Muralidharan}, \binits{S.}},
\oauthor{\bsnm{Lin}, \binits{Y.C.}},
\oauthor{\bsnm{Molchanov}, \binits{P.}}:
Small language models are the future of agentic {AI}.
Preprint at https://arxiv.org/abs/2506.02153
(2025)
\end{botherref}
\endbibitem

\bibitem[\protect\citeauthoryear{Yang et~al.}{2024}]{yang2024plug}
\begin{bchapter}
\bauthor{\bsnm{Yang}, \binits{Z.}},
\bauthor{\bsnm{Raman}, \binits{S.S.}},
\bauthor{\bsnm{Shah}, \binits{A.}},
\bauthor{\bsnm{Tellex}, \binits{S.}}:
\bctitle{Plug in the safety chip: enforcing constraints for {LLM}-driven robot
  agents}.
In: \bbtitle{Proc. IEEE International Conference on Robotics and Automation
  (ICRA)}
(\byear{2024})
\end{bchapter}
\endbibitem

\bibitem[\protect\citeauthoryear{Yu et~al.}{2023}]{yu2023language}
\begin{botherref}
\oauthor{\bsnm{Yu}, \binits{W.}},
\oauthor{\bsnm{Gileadi}, \binits{N.}},
\oauthor{\bsnm{Fu}, \binits{C.}},
\oauthor{\bsnm{Kirmani}, \binits{S.}},
\oauthor{\bsnm{Lee}, \binits{K.-H.}},
\oauthor{\bsnm{Arenas}, \binits{M.G.}},
\oauthor{\bsnm{Chiang}, \binits{H.-T.L.}},
\oauthor{\bsnm{Erez}, \binits{T.}},
\oauthor{\bsnm{Hasenclever}, \binits{L.}},
\oauthor{\bsnm{Humplik}, \binits{J.}}, et al.:
Language to rewards for robotic skill synthesis.
arXiv preprint arXiv:2306.08647
(2023)
\end{botherref}
\endbibitem

\bibitem[\protect\citeauthoryear{Cemri et~al.}{2025}]{cemri2025why}
\begin{bchapter}
\bauthor{\bsnm{Cemri}, \binits{M.}},
\bauthor{\bsnm{Pan}, \binits{M.Z.}},
\bauthor{\bsnm{Yang}, \binits{S.}},
\bauthor{\bsnm{Agrawal}, \binits{L.A.}},
\bauthor{\bsnm{Chopra}, \binits{B.}},
\bauthor{\bsnm{Tiwari}, \binits{R.}},
\bauthor{\bsnm{Keutzer}, \binits{K.}},
\bauthor{\bsnm{Parameswaran}, \binits{A.}},
\bauthor{\bsnm{Klein}, \binits{D.}},
\bauthor{\bsnm{Ramchandran}, \binits{K.}},
\bauthor{\bsnm{Zaharia}, \binits{M.}},
\bauthor{\bsnm{Gonzalez}, \binits{J.E.}},
\bauthor{\bsnm{Stoica}, \binits{I.}}:
\bctitle{Why do multi-agent {LLM} systems fail?}
In: \bbtitle{Advances in Neural Information Processing Systems (NeurIPS),
  Datasets and Benchmarks Track}
(\byear{2025})
\end{bchapter}
\endbibitem

\bibitem[\protect\citeauthoryear{Huang et~al.}{2024}]{huang2024resilience}
\begin{botherref}
\oauthor{\bsnm{Huang}, \binits{J.-t.}},
\oauthor{\bsnm{Zhou}, \binits{J.}},
\oauthor{\bsnm{Jin}, \binits{T.}},
\oauthor{\bsnm{Zhou}, \binits{X.}},
\oauthor{\bsnm{Chen}, \binits{Z.}},
\oauthor{\bsnm{Wang}, \binits{W.}},
\oauthor{\bsnm{Yuan}, \binits{Y.}},
\oauthor{\bsnm{Lyu}, \binits{M.R.}},
\oauthor{\bsnm{Sap}, \binits{M.}}:
On the Resilience of {LLM}-Based Multi-Agent Collaboration with Faulty Agents.
arXiv:2408.00989
(2024)
\end{botherref}
\endbibitem

\bibitem[\protect\citeauthoryear{Chen}{2018}]{chen2018dove}
\begin{botherref}
\oauthor{\bsnm{Chen}, \binits{S.}}:
China takes surveillance to new heights with flock of robotic {D}oves.
South China Morning Post.
Published 24 June 2018
(2018)
\end{botherref}
\endbibitem

\bibitem[\protect\citeauthoryear{Andersson et~al.}{2019}]{andersson2019casadi}
\begin{barticle}
\bauthor{\bsnm{Andersson}, \binits{J.A.E.}},
\bauthor{\bsnm{Gillis}, \binits{J.}},
\bauthor{\bsnm{Horn}, \binits{G.}},
\bauthor{\bsnm{Rawlings}, \binits{J.B.}},
\bauthor{\bsnm{Diehl}, \binits{M.}}:
\batitle{{CasADi}: a software framework for nonlinear optimization and optimal
  control}.
\bjtitle{Mathematical Programming Computation}
\bvolume{11}(\bissue{1}),
\bfpage{1}--\blpage{36}
(\byear{2019})
\doiurl{10.1007/s12532-018-0139-4}
\end{barticle}
\endbibitem

\bibitem[\protect\citeauthoryear{Stellato et~al.}{2020}]{stellato2020osqp}
\begin{barticle}
\bauthor{\bsnm{Stellato}, \binits{B.}},
\bauthor{\bsnm{Banjac}, \binits{G.}},
\bauthor{\bsnm{Goulart}, \binits{P.}},
\bauthor{\bsnm{Bemporad}, \binits{A.}},
\bauthor{\bsnm{Boyd}, \binits{S.}}:
\batitle{{OSQP}: an operator splitting solver for quadratic programs}.
\bjtitle{Mathematical Programming Computation}
\bvolume{12}(\bissue{4}),
\bfpage{637}--\blpage{672}
(\byear{2020})
\doiurl{10.1007/s12532-020-00179-2}
\end{barticle}
\endbibitem

\bibitem[\protect\citeauthoryear{Wächter and Biegler}{2006}]{wachter2006ipopt}
\begin{barticle}
\bauthor{\bsnm{Wächter}, \binits{A.}},
\bauthor{\bsnm{Biegler}, \binits{L.T.}}:
\batitle{On the implementation of an interior-point filter line-search
  algorithm for large-scale nonlinear programming}.
\bjtitle{Mathematical Programming}
\bvolume{106}(\bissue{1}),
\bfpage{25}--\blpage{57}
(\byear{2006})
\doiurl{10.1007/s10107-004-0559-y}
\end{barticle}
\endbibitem

\bibitem[\protect\citeauthoryear{Ames et~al.}{2019}]{ames2019control}
\begin{bchapter}
\bauthor{\bsnm{Ames}, \binits{A.D.}},
\bauthor{\bsnm{Coogan}, \binits{S.}},
\bauthor{\bsnm{Egerstedt}, \binits{M.}},
\bauthor{\bsnm{Notomista}, \binits{G.}},
\bauthor{\bsnm{Sreenath}, \binits{K.}},
\bauthor{\bsnm{Tabuada}, \binits{P.}}:
\bctitle{Control barrier functions: Theory and applications}.
In: \bbtitle{2019 18th European Control Conference (ECC)},
pp. \bfpage{3420}--\blpage{3431}.
\bpublisher{IEEE}, \blocation{???}
(\byear{2019}).
\doiurl{10.23919/ecc.2019.8796030}
\end{bchapter}
\endbibitem

\bibitem[\protect\citeauthoryear{Dhiman et~al.}{2023}]{dhiman2021control}
\begin{barticle}
\bauthor{\bsnm{Dhiman}, \binits{V.}},
\bauthor{\bsnm{Khojasteh}, \binits{M.J.}},
\bauthor{\bsnm{Franceschetti}, \binits{M.}},
\bauthor{\bsnm{Atanasov}, \binits{N.}}:
\batitle{Control barriers in bayesian learning of system dynamics}.
\bjtitle{IEEE transactions on automatic control}
\bvolume{68}(\bissue{1}),
\bfpage{214}--\blpage{229}
(\byear{2023})
\end{barticle}
\endbibitem

\bibitem[\protect\citeauthoryear{Cheng et~al.}{2020}]{cheng2020safe}
\begin{bchapter}
\bauthor{\bsnm{Cheng}, \binits{R.}},
\bauthor{\bsnm{Khojasteh}, \binits{M.J.}},
\bauthor{\bsnm{Ames}, \binits{A.D.}},
\bauthor{\bsnm{Burdick}, \binits{J.W.}}:
\bctitle{Safe multi-agent interaction through robust control barrier functions
  with learned uncertainties}.
In: \bbtitle{2020 59th IEEE Conference on Decision and Control (CDC)},
pp. \bfpage{777}--\blpage{783}
(\byear{2020}).
\bcomment{IEEE}
\end{bchapter}
\endbibitem

\bibitem[\protect\citeauthoryear{Gilbert}{1960}]{gilbert1960burst}
\begin{barticle}
\bauthor{\bsnm{Gilbert}, \binits{E.N.}}:
\batitle{Capacity of a burst-noise channel}.
\bjtitle{Bell System Technical Journal}
\bvolume{39}(\bissue{5}),
\bfpage{1253}--\blpage{1265}
(\byear{1960})
\doiurl{10.1002/j.1538-7305.1960.tb03959.x}
\end{barticle}
\endbibitem

\bibitem[\protect\citeauthoryear{Elliott}{1963}]{elliott1963burst}
\begin{barticle}
\bauthor{\bsnm{Elliott}, \binits{E.O.}}:
\batitle{Estimates of error rates for codes on burst-noise channels}.
\bjtitle{Bell System Technical Journal}
\bvolume{42}(\bissue{5}),
\bfpage{1977}--\blpage{1997}
(\byear{1963})
\doiurl{10.1002/j.1538-7305.1963.tb00955.x}
\end{barticle}
\endbibitem

\bibitem[\protect\citeauthoryear{Miao et~al.}{2016}]{miao2016coding}
\begin{barticle}
\bauthor{\bsnm{Miao}, \binits{F.}},
\bauthor{\bsnm{Zhu}, \binits{Q.}},
\bauthor{\bsnm{Pajic}, \binits{M.}},
\bauthor{\bsnm{Pappas}, \binits{G.J.}}:
\batitle{Coding schemes for securing cyber-physical systems against stealthy
  data injection attacks}.
\bjtitle{IEEE Transactions on Control of Network Systems}
\bvolume{4}(\bissue{1}),
\bfpage{106}--\blpage{117}
(\byear{2016})
\end{barticle}
\endbibitem

\bibitem[\protect\citeauthoryear{{Google DeepMind}}{2026}]{gemini31flashlite}
\begin{botherref}
\oauthor{\bsnm{{Google DeepMind}}}:
Gemini 3.1 {Flash-Lite} Model Card.
\url{https://deepmind.google/models/model-cards/gemini-3-1-flash-lite/}.
Published 3 March 2026; accessed 5 October 2026
(2026)
\end{botherref}
\endbibitem

\bibitem[\protect\citeauthoryear{{OpenAI}}{2026}]{gpt61sol}
\begin{botherref}
\oauthor{\bsnm{{OpenAI}}}:
{GPT-6.1 Sol} Model.
\url{https://developers.openai.com/api/docs/models/gpt-6.1-sol}.
OpenAI API documentation; accessed 17 October 2026
(2026)
\end{botherref}
\endbibitem

\bibitem[\protect\citeauthoryear{Coluccia et~al.}{2021}]{coluccia2021drone}
\begin{barticle}
\bauthor{\bsnm{Coluccia}, \binits{A.}},
\bauthor{\bsnm{Fascista}, \binits{A.}},
\bauthor{\bsnm{Schumann}, \binits{A.}},
\bauthor{\bsnm{Sommer}, \binits{L.}},
\bauthor{\bsnm{Dimou}, \binits{A.}},
\bauthor{\bsnm{Zarpalas}, \binits{D.}},
\bauthor{\bsnm{M{\'e}ndez}, \binits{M.}},
\bauthor{\bsnm{Iglesia}, \binits{D.}},
\bauthor{\bsnm{Gonz{\'a}lez}, \binits{I.}},
\bauthor{\bsnm{Mercier}, \binits{J.-P.}},
\bauthor{\bsnm{Gagn{\'e}}, \binits{G.}},
\bauthor{\bsnm{Mitra}, \binits{A.}},
\bauthor{\bsnm{Rajashekar}, \binits{S.}}:
\batitle{Drone vs. bird detection: deep learning algorithms and results from a
  grand challenge}.
\bjtitle{Sensors}
\bvolume{21},
\bfpage{2824}
(\byear{2021})
\doiurl{10.3390/s21082824}
\end{barticle}
\endbibitem

\bibitem[\protect\citeauthoryear{Upshaw and
  Hassanalian}{2023}]{upshaw2023application}
\begin{bchapter}
\bauthor{\bsnm{Upshaw}, \binits{J.}},
\bauthor{\bsnm{Hassanalian}, \binits{M.}}:
\bctitle{Application of camouflaging fixed-wing, quadcopter, and ornithopter
  uavs}.
In: \bbtitle{2023 Regional Student Conferences},
p. \bfpage{72279}
(\byear{2023})
\end{bchapter}
\endbibitem

\bibitem[\protect\citeauthoryear{Vosbein et~al.}{2025}]{vosbein2025swimming}
\begin{bchapter}
\bauthor{\bsnm{Vosbein}, \binits{D.}},
\bauthor{\bsnm{McDonagh}, \binits{K.}},
\bauthor{\bsnm{Goodyear}, \binits{S.}},
\bauthor{\bsnm{Hassanalian}, \binits{M.}}:
\bctitle{Swimming robotic taxidermy duck for wildlife monitoring: Development
  and initial testing}.
In: \bbtitle{AIAA AVIATION FORUM AND ASCEND 2025},
p. \bfpage{3527}
(\byear{2025})
\end{bchapter}
\endbibitem

\bibitem[\protect\citeauthoryear{Glasserman and Yao}{1992}]{glasserman1992crn}
\begin{barticle}
\bauthor{\bsnm{Glasserman}, \binits{P.}},
\bauthor{\bsnm{Yao}, \binits{D.D.}}:
\batitle{Some guidelines and guarantees for common random numbers}.
\bjtitle{Management Science}
\bvolume{38}(\bissue{6}),
\bfpage{884}--\blpage{908}
(\byear{1992})
\doiurl{10.1287/mnsc.38.6.884}
\end{barticle}
\endbibitem

\bibitem[\protect\citeauthoryear{Newcombe}{1998}]{newcombe1998paired}
\begin{barticle}
\bauthor{\bsnm{Newcombe}, \binits{R.G.}}:
\batitle{Improved confidence intervals for the difference between binomial
  proportions based on paired data}.
\bjtitle{Statistics in Medicine}
\bvolume{17}(\bissue{22}),
\bfpage{2635}--\blpage{2650}
(\byear{1998})
\doiurl{10.1002/(sici)1097-0258(19981130)17:22<2635::aid-sim954>3.0.co;2-c}
\end{barticle}
\endbibitem

\bibitem[\protect\citeauthoryear{Abaid~Ullah
  et~al.}{2026a}]{abaidullah2026code}
\begin{botherref}
\oauthor{\bsnm{Abaid~Ullah}, \binits{K.}},
\oauthor{\bsnm{Acharya}, \binits{A.}},
\oauthor{\bsnm{Khojasteh}, \binits{M.J.}}:
{Code for: Consensus Within, Competition Between: Decentralized LLM Teams for
  Perimeter Defense}.
Zenodo https://doi.org/10.5281/zenodo.24209157
(2026)
\end{botherref}
\endbibitem

\bibitem[\protect\citeauthoryear{Abaid~Ullah
  et~al.}{2026b}]{abaidullah2026data}
\begin{botherref}
\oauthor{\bsnm{Abaid~Ullah}, \binits{K.}},
\oauthor{\bsnm{Acharya}, \binits{A.}},
\oauthor{\bsnm{Khojasteh}, \binits{M.J.}}:
{Data for: Consensus Within, Competition Between: Decentralized LLM Teams for
  Perimeter Defense}.
Zenodo https://doi.org/10.5281/zenodo.24208099
(2026)
\end{botherref}
\endbibitem

\end{thebibliography}

\clearpage
\setcounter{section}{0}\setcounter{figure}{0}\setcounter{table}{0}
\renewcommand{\thesection}{S\arabic{section}}
\renewcommand{\theHsection}{S\arabic{section}}
\renewcommand{\theHfigure}{S\arabic{figure}}
\renewcommand{\theHtable}{S\arabic{table}}
\renewcommand{\figurename}{Supplementary Fig.}
\renewcommand{\tablename}{Supplementary Table}
\vspace*{0.3\textheight}
\begin{center}
{\LARGE\bfseries Supplementary Information}\par\bigskip
{\large Consensus Within, Competition Between:\\ Decentralized LLM Teams for Perimeter Defense}\par\bigskip\bigskip
\parbox{0.8\linewidth}{\centering\small Supplementary Data~1 and Supplementary Videos~1--3 are the ancillary files of this arXiv submission.}
\end{center}
\clearpage
\section{Contents}
{\small\begin{tabular}{@{}p{3.8cm}@{}p{\dimexpr\linewidth-3.8cm\relax}@{}}
\multicolumn{2}{@{}l}{\textbf{Supplementary Figures}}\\
Supplementary Fig.~\ref{fig:breachgrid} & Breach outcome of every episode in the main study \dotfill p.~\pageref{fig:breachgrid}\\
Supplementary Fig.~\ref{fig:ep-consensus} & An episode of the consensus team \dotfill p.~\pageref{fig:ep-consensus}\\
Supplementary Fig.~\ref{fig:ep-independent} & An episode of the independent team \dotfill p.~\pageref{fig:ep-independent}\\
Supplementary Fig.~\ref{fig:ep-limited-range} & Commander orders that never arrived \dotfill p.~\pageref{fig:ep-limited-range}\\
Supplementary Fig.~\ref{fig:ep-airspace-lowcost} & Low-cost commander, birds and civilian drones \dotfill p.~\pageref{fig:ep-airspace-lowcost}\\
Supplementary Fig.~\ref{fig:ep-airspace-nearfrontier} & Near-frontier commander, birds and civilian drones \dotfill p.~\pageref{fig:ep-airspace-nearfrontier}\\
Supplementary Fig.~\ref{fig:reason-birds} & Bird images and the LLMs' reasoning \dotfill p.~\pageref{fig:reason-birds}\\
Supplementary Fig.~\ref{fig:reason-civilian} & Civilian-drone images and the LLMs' reasoning \dotfill p.~\pageref{fig:reason-civilian}\\
Supplementary Fig.~\ref{fig:reason-attack} & Attack-drone images and the LLMs' reasoning \dotfill p.~\pageref{fig:reason-attack}\\
Supplementary Fig.~\ref{fig:ep-disguise-nowarning} & Disguised intruders, without the forewarning \dotfill p.~\pageref{fig:ep-disguise-nowarning}\\
Supplementary Fig.~\ref{fig:ep-disguise-warning} & Disguised intruders, with the forewarning \dotfill p.~\pageref{fig:ep-disguise-warning}\\
Supplementary Fig.~\ref{fig:reason-disguise-u} & Bird-drone images and the LLMs' reasoning, without the forewarning \dotfill p.~\pageref{fig:reason-disguise-u}\\
Supplementary Fig.~\ref{fig:reason-disguise-w} & Bird-drone images and the LLMs' reasoning, with the forewarning \dotfill p.~\pageref{fig:reason-disguise-w}\\
Supplementary Fig.~\ref{fig:filter} & The barrier filter in head-on states \dotfill p.~\pageref{fig:filter}\\[3pt]
\multicolumn{2}{@{}l}{\textbf{Supplementary Tables}}\\
Supplementary Table~\ref{tab:cells} & Outcomes of every cell \dotfill p.~\pageref{tab:cells}\\
Supplementary Table~\ref{tab:arch} & Breach-rate differences between the defender teams \dotfill p.~\pageref{tab:arch}\\
Supplementary Table~\ref{tab:coord} & Allocation measures by team and network condition \dotfill p.~\pageref{tab:coord}\\
Supplementary Table~\ref{tab:cond} & Change in breach rate under each condition, by intruder count \dotfill p.~\pageref{tab:cond}\\
Supplementary Table~\ref{tab:exposure} & Exposure to falsified reports \dotfill p.~\pageref{tab:exposure}\\
Supplementary Table~\ref{tab:conn} & Network connectivity under limited radio range \dotfill p.~\pageref{tab:conn}\\
Supplementary Table~\ref{tab:weights} & Bid weights and pursuit multipliers chosen at high authority \dotfill p.~\pageref{tab:weights}\\
Supplementary Table~\ref{tab:authority} & High authority and model capability: comparisons \dotfill p.~\pageref{tab:authority}\\
Supplementary Table~\ref{tab:decoyseeds} & Episodes with birds and civilian drones, by seed \dotfill p.~\pageref{tab:decoyseeds}\\
Supplementary Table~\ref{tab:decoyanswers} & Answers about each kind of contact with birds and civilian drones present \dotfill p.~\pageref{tab:decoyanswers}\\
Supplementary Table~\ref{tab:airspace} & Outcomes and treatment of contacts with birds and civilian drones, and with disguised intruders \dotfill p.~\pageref{tab:airspace}\\
Supplementary Table~\ref{tab:notation} & Notation \dotfill p.~\pageref{tab:notation}\\
Supplementary Table~\ref{tab:profiles} & Model profiles, calls, tokens, latency and cost \dotfill p.~\pageref{tab:profiles}\\
Supplementary Table~\ref{tab:repro} & Execution identities and re-simulation checks \dotfill p.~\pageref{tab:repro}\\
Supplementary Table~\ref{tab:models} & Low-cost models compared in the selection study \dotfill p.~\pageref{tab:models}\\
Supplementary Table~\ref{tab:commandercheck} & Commander check before the main study \dotfill p.~\pageref{tab:commandercheck}\\[3pt]
\multicolumn{2}{@{}l}{\textbf{Supplementary Notes}}\\
Supplementary Note~\ref{note:prompts} & Prompts, response schemas and example calls \dotfill p.~\pageref{note:prompts}\\
Supplementary Note~\ref{note:models} & Choice of the two LLMs \dotfill p.~\pageref{note:models}\\[3pt]
\multicolumn{2}{@{}l}{\textbf{Supplementary Data} (separate file; title and legend in this document)}\\
Supplementary Data~\ref{data:contrasts} & Every paired comparison with its per-seed values \dotfill p.~\pageref{data:contrasts}\\[3pt]
\multicolumn{2}{@{}l}{\textbf{Supplementary Videos} (separate files; titles and legends in this document)}\\
Supplementary Videos~\ref{vid:commander}--\ref{vid:independent} & One example episode of each team in the main study \dotfill p.~\pageref{vid:commander}\\
\end{tabular}}%
\clearpage
\section{Terms used in this document}

\textbf{Game.} Four defenders protect a region against two or four intruders in a simulated arena, a 100 by 100 square in abstract units, in which one unit can be read as 10\,m. Each agent's sensor covers a 60\textdegree{} field of view (FOV) and reaches 35 units, a distance called the sensor range, and an agent sees opponents only inside this area. Every agent on both teams consults a large language model (LLM) once per decision epoch of 1\,sec. Unless stated otherwise the LLM is Gemini 3.1 Flash-Lite, the low-cost LLM. GPT-6.1 Sol is the near-frontier LLM. A breach is an intruder entering the region within the 40\,sec episode, and a team's breach rate is the share of its episodes with a breach. An agent is active until it is destroyed, disabled or leaves the arena. A contact is any object a defender detects, whose kind it is not told. Each agent acts on its beliefs: for each opponent it knows of, the last observed position and velocity, from its own sensor or a teammate's message, projected forward from the time of that observation. A known intruder is one that some defender holds a belief about. In the main study every LLM works from a state-based observation, a text description of positions, velocities and the values computed from them, with no images. A seed is one of 30 random starting configurations (4001 to 4030), and episodes that share a seed and an intruder count are matched across teams and conditions.

\textbf{Defender teams.}
In the commander team, the LLM of one defender assigns a task to every defender, and its orders travel over the network. In the consensus team, defenders bid for tasks and agree on who takes which through the consensus-based auction algorithm (CBAA), an algorithm rather than a discussion among their LLMs. The independent team runs the same agents with the auction messages switched off, so each defender commits to its own highest bid. An LLM's authority is how much of each decision it makes itself rather than leaving it to fixed rules. At low authority, each defender scores every task with a fixed formula. Its LLM adjusts the scores only by choosing a weight profile, a priority for each intruder and up to three preferred tasks, so it can make a known intruder less attractive but never drop it. At high authority, the consensus team's LLMs instead enter their own value for each of the 12 weights of the formula, within allowed ranges, and a multiplier from 0 to 2 on each intruder's pursuit value. They can thus set any intruder's pursuit value to zero, and the bids enter the same auction. High authority does not include every low-authority choice (Supplementary Table~\ref{tab:notation}). The independent team always runs at low authority. The consensus team runs at low authority unless stated otherwise. The intruders always use the low-cost LLM and are the same for every team.

\textbf{Network conditions}, always in this order: the ideal network (complete and lossless, with one round of latency per link), independent loss, burst loss, falsified reports from one compromised defender, and limited radio range (links of 15 units, with relaying). Under independent loss each link loses each 0.1\,sec exchange with a 30\% chance, independently of other links and exchanges, and under burst loss the rate is the same on average, but losses come in runs.

\textbf{Airspace challenges.} One bird and two civilian drones, together the non-hostile contacts, fly among four intruders on the ideal network (seeds 4001 to 4005). The defender and commander LLMs then receive a multimodal observation, the same text plus one image of every contact, and their prompts gain the airspace paragraph. In the disguise runs each intruder is shown as an image of a flapping-wing drone built from a preserved bird, which makes it a disguised intruder. These runs are made without and with a one-sentence forewarning that some intruders may be disguised as birds. An LLM drops a contact when its answer removes the contact's pursuit: no defender assigned by a commander, or a pursuit multiplier of zero at high authority. At low authority it can only deprioritize a contact, by giving it a low priority, which halves the value of its pursuit but keeps the pursuit possible. A contact-epoch is one contact in one epoch.

\textbf{Barrier filter.} Every agent's acceleration passes through a barrier filter on the candidate barrier $h = \lVert r \rVert^2 - D^2$, where $r$ is the relative position of two agents and $D = 3$ units is the minimum separation. A defender's filter ignores the intruder it pursues, so that it can intercept it. The filter guarantees one step of separation only when its constraints are feasible.

\textbf{Statistics.} The seed is the unit of analysis. A comparison is the mean over 30 seeds of each seed's paired difference, with a two-sided 95\% confidence interval (CI) from the $t$ distribution (29 degrees of freedom). Each comparison is read from its mean; the CI shows its precision, and no result is classified by whether its CI includes zero. Results with birds, civilian drones and disguised intruders (five seeds) are counts and ranges, without CIs or tests. Full definitions are in the Methods of the paper.

\clearpage

\clearpage
\section{Supplementary Figures}
Supplementary Figs. 2--6, 10 and 11 show screenshots from single re-simulated episodes, with times counted from the start. Each episode was chosen by a rule fixed before any image was drawn, usually the first seed that shows the behavior in question. The moments within each episode were picked by hand, to show how it unfolded. These screenshots are illustrative only. They show that a behavior can occur and what it looks like, but not how often it occurs or whether it is typical, because other seeds end differently. For how often each outcome occurs, see the Supplementary Tables.

\begin{figure}[ht]
\centering
\includegraphics[width=\linewidth]{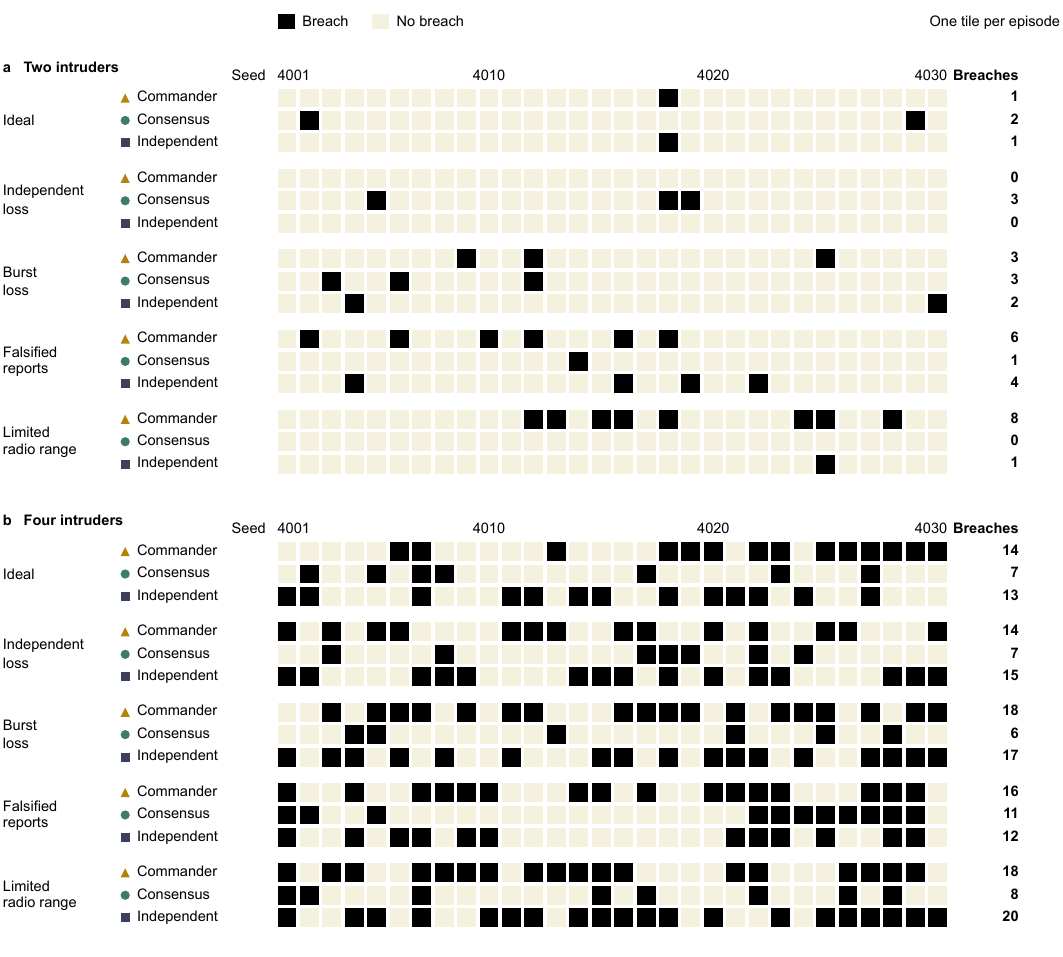}
\refstepcounter{figure}\label{fig:breachgrid}
\silegendstart
\textbf{Supplementary Fig. 1\sicapsep{}Breach outcome of every episode in the main study.}
Each tile is one episode, marked by whether an intruder breached the region within the 40 sec episode. Columns are seeds 4001 to 4030; rows are the three defender teams under each network condition. \textbf{(a)} Two intruders. \textbf{(b)} Four intruders. Bold numbers are breaches of 30 per row.
\end{figure}
\clearpage

\begin{figure}[ht]
\centering
\includegraphics[width=\linewidth]{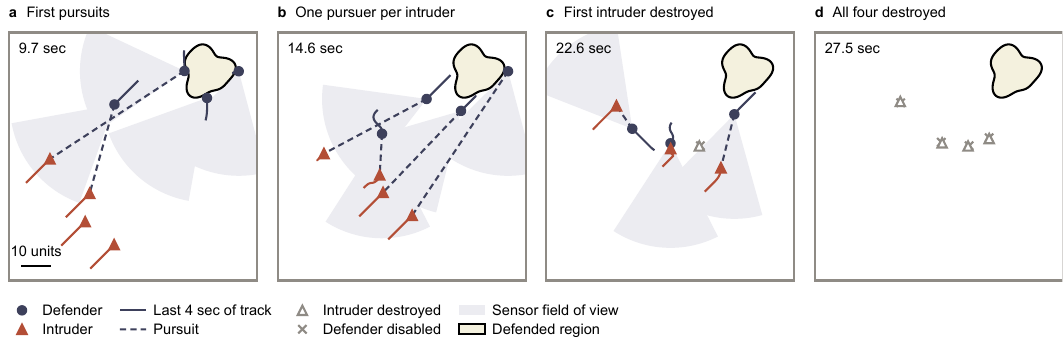}
\refstepcounter{figure}\label{fig:ep-consensus}
\silegendstart
\textbf{Supplementary Fig. 2\sicapsep{}An episode of the consensus team.}
Ideal network, seed 4003. \textbf{(a)} Two defenders pursue the first two intruders seen. \textbf{(b)} Every intruder has its own pursuer. \textbf{(c)} The first intruder is destroyed, and its pursuer is disabled with it. \textbf{(d)} All four intruders are destroyed, and none breached.
 \end{figure}

\begin{figure}[ht]
\centering
\includegraphics[width=\linewidth]{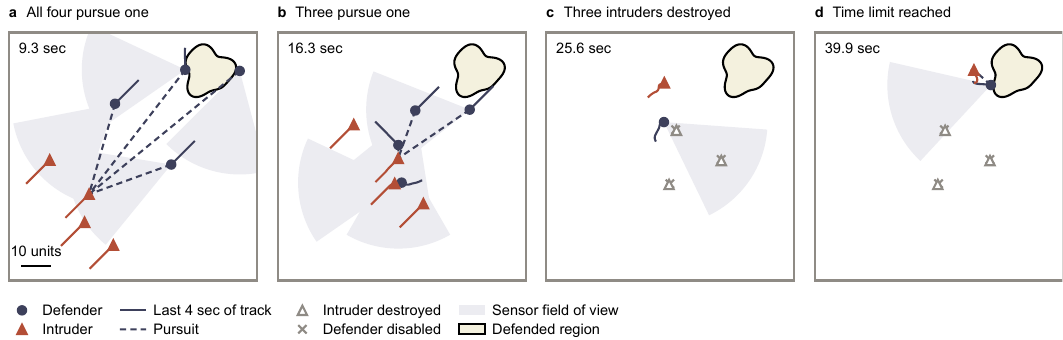}
\refstepcounter{figure}\label{fig:ep-independent}
\silegendstart
\textbf{Supplementary Fig. 3\sicapsep{}An episode of the independent team.}
Ideal network, seed 4003, the same starts as Supplementary Fig.~\ref{fig:ep-consensus}. \textbf{(a)} All four defenders pursue the same intruder. \textbf{(b)} Three pursue one intruder while two others go unpursued. \textbf{(c)} Three intruders are destroyed, each with a defender. \textbf{(d)} The episode reaches the 40 sec limit with one intruder alive.
 \end{figure}

\begin{figure}[ht]
\centering
\includegraphics[width=\linewidth]{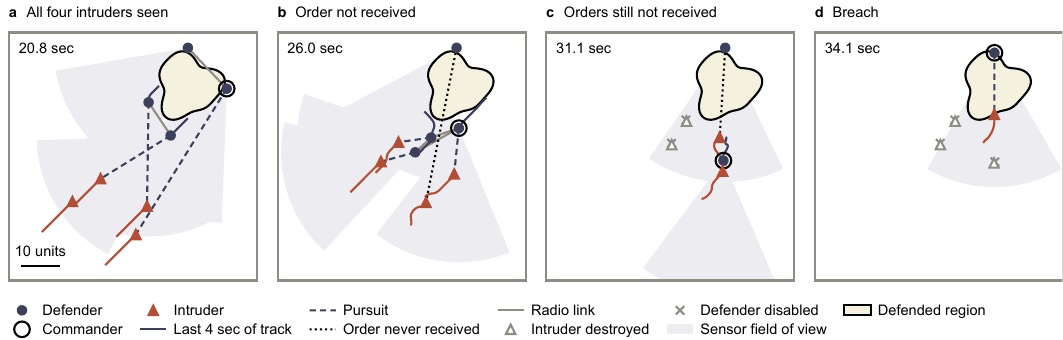}
\refstepcounter{figure}\label{fig:ep-limited-range}
\silegendstart
\textbf{Supplementary Fig. 4\sicapsep{}An episode of the commander team under limited radio range.}
Seed 4001. Grey lines link defenders within radio range. \textbf{(a)} All four intruders are seen, and three defenders pursue. \textbf{(b)} The commander's order sending the defender at the top after an intruder never arrives, because no teammate is within radio range of it. \textbf{(c)} Its orders to pursue the intruder that later breaches never arrive either. \textbf{(d)} The commander was disabled at 32.2 sec, the defender at the top took command and pursued that intruder from 33.0 sec, and the intruder breached at 34.1 sec.
 \end{figure}

\begin{figure}[ht]
\centering
\includegraphics[width=\linewidth]{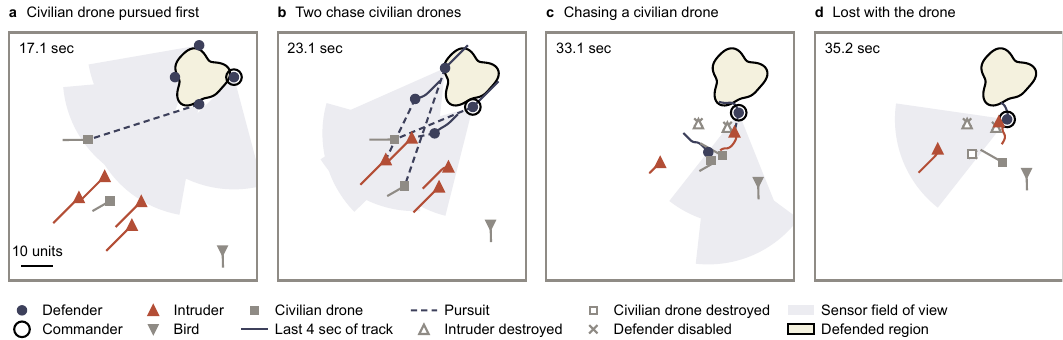}
\refstepcounter{figure}\label{fig:ep-airspace-lowcost}
\silegendstart
\textbf{Supplementary Fig. 5\sicapsep{}An episode of the low-cost commander among birds and civilian drones.}
Ideal network, one bird and two civilian drones, seed 4001. \textbf{(a)} The first pursuit the commander assigns is of a civilian drone. \textbf{(b)} Two defenders chase civilian drones. \textbf{(c)} A defender chases a civilian drone. \textbf{(d)} It collided with the civilian drone at 35.1 s: the civilian drone was destroyed and the defender disabled. Two intruders were destroyed and none breached.
 \end{figure}

\begin{figure}[ht]
\centering
\includegraphics[width=\linewidth]{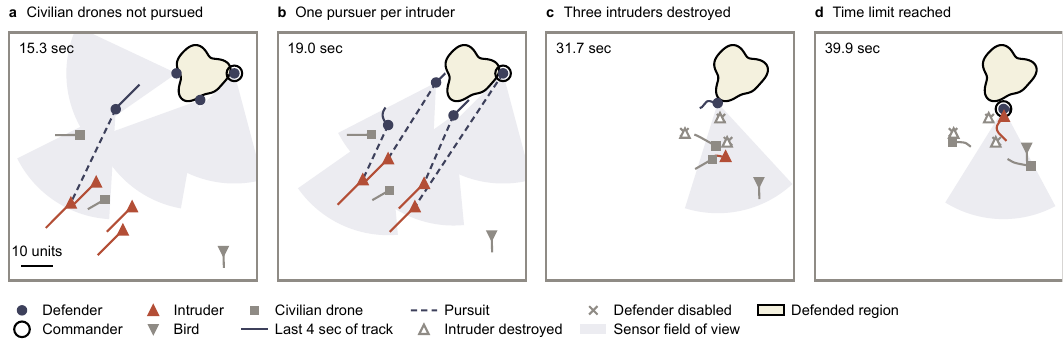}
\refstepcounter{figure}\label{fig:ep-airspace-nearfrontier}
\silegendstart
\textbf{Supplementary Fig. 6\sicapsep{}An episode of the near-frontier commander among birds and civilian drones.}
The setting and seed of Supplementary Fig.~\ref{fig:ep-airspace-lowcost}. \textbf{(a)} The civilian drones are seen but not pursued. \textbf{(b)} Every intruder has its own pursuer. \textbf{(c)} Three intruders are destroyed. \textbf{(d)} The episode reaches the 40 sec limit without a breach. No defender pursued a bird or civilian drone.
 \end{figure}

\begin{figure}[ht]
\centering
\begin{minipage}[t]{0.31\linewidth}\centering
\includegraphics[width=\linewidth,height=0.62\linewidth,keepaspectratio]{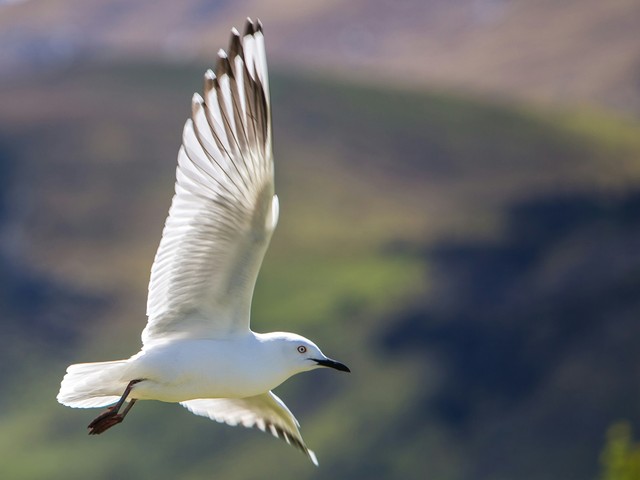}\\[2pt]
\textbf{(a) Bird 1}\par\smallskip
\leftskip0pt\raggedright\figurecaptionfont
\textbf{Low-cost model} (contact 4): \llmquote{``With only 3 seconds left in the episode (30 steps), and given that the optimistic interception forecasts for blue 1 against the approaching contacts 2, 3, and 4 are all beyond the episode duration (5.4s, 12.7s, and 5.8s respectively), it is impossible for blue 1 to neutralize these threats before the episode concludes. \ldots{}''}\par\smallskip
\textbf{Near-frontier model} (contact 4): \llmquote{``Contacts 0, 1, and 3 appear to be intruder drones; contact 4 is a bird, and 5 and 6 appear civilian. \ldots{}''}
\end{minipage}\hfill
\begin{minipage}[t]{0.31\linewidth}\centering
\includegraphics[width=\linewidth,height=0.62\linewidth,keepaspectratio]{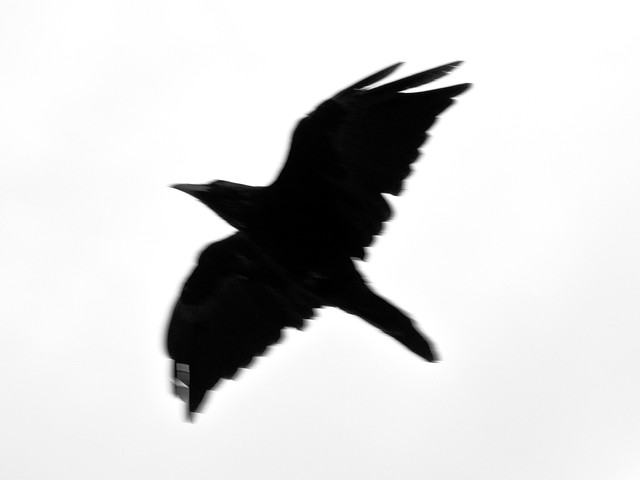}\\[2pt]
\textbf{(b) Bird 2}\par\smallskip
\leftskip0pt\raggedright\figurecaptionfont
\textbf{Low-cost model} (contact 4): \llmquote{``\ldots{} Blue 3 is closest to contact 1, Blue 2 is closest to contact 0, Blue 0 is directed to intercept contact 5, and Blue 1 is assigned to contact 4. \ldots{}''}\par\smallskip
\textbf{Near-frontier model} (contact 4): \llmquote{``\ldots{} Contact 4 is a bird and contact 5 appears to be a civilian consumer drone, so avoid sacrificing defenders on them. \ldots{}''}
\end{minipage}\hfill
\begin{minipage}[t]{0.31\linewidth}\centering
\includegraphics[width=\linewidth,height=0.62\linewidth,keepaspectratio]{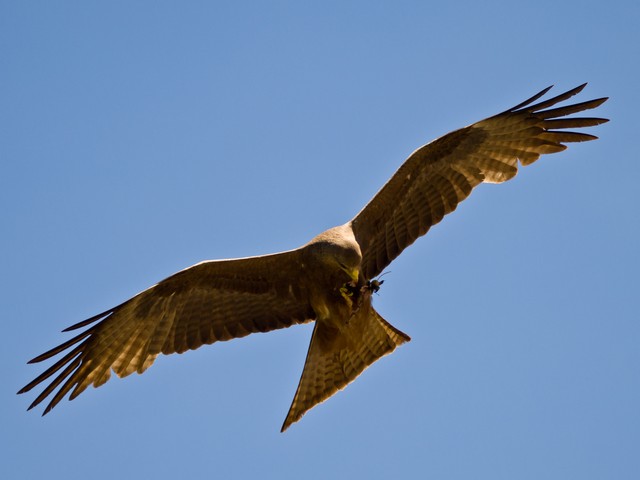}\\[2pt]
\textbf{(c) Bird 3}\par\smallskip
\leftskip0pt\raggedright\figurecaptionfont
\textbf{Low-cost model} (contact 4): \llmquote{``Contact 4 is the only detected intruder and is approaching the defended region. \ldots{}''}\par\smallskip
\textbf{Near-frontier model} (contact 4): \llmquote{``Contact 4 is visibly a bird, so intercepting it would needlessly disable a defender. \ldots{}''}
\end{minipage}
 \refstepcounter{figure}\label{fig:reason-birds}
\silegendstart[\bigskip]
\textbf{Supplementary Fig.~\thefigure\sicapsep{}The bird images and the commander's reasoning about them.}
Each image is shown as the LLMs received it, with the reasoning of the commander team's LLM in the airspace challenge (low-cost model, Gemini 3.1 Flash-Lite; near-frontier model, GPT-6.1 Sol). For each image and model the quote comes from the first commander call, seeds in order from 4001, that attached the image to a contact and named that contact in its reasoning. Every sentence naming the contact is quoted, with ellipses for skipped sentences. Contact numbers are the identifiers the LLM saw, and the prompts call the defenders \texttt{blue} and the intruders \texttt{red}. Images: (a) R. Long, Unsplash License; (b) CGP Grey, CC BY 2.0; (c) W. Warby, CC BY 2.0; cropped and resized.
\end{figure}
\clearpage

\begin{figure}[ht]
\centering
\begin{minipage}[t]{0.31\linewidth}\centering
\includegraphics[width=\linewidth,height=0.62\linewidth,keepaspectratio]{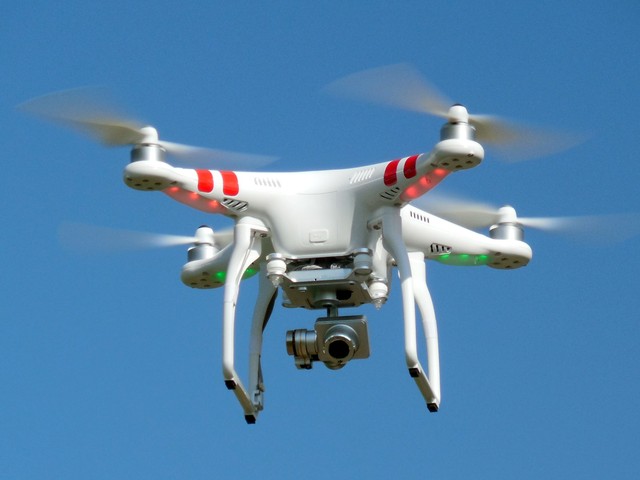}\\[2pt]
\textbf{(a) Civilian drone 1}\par\smallskip
\leftskip0pt\raggedright\figurecaptionfont
\textbf{Low-cost model} (contact 5): \llmquote{``Contact 5 is identified as an approaching threat. \ldots{}''}\par\smallskip
\textbf{Near-frontier model} (contact 5): \llmquote{``Contact 5 appears to be a civilian camera drone, so intercepting it would risk sacrificing a defender without stopping an intruder. \ldots{}''}
\end{minipage}\hfill
\begin{minipage}[t]{0.31\linewidth}\centering
\includegraphics[width=\linewidth,height=0.62\linewidth,keepaspectratio]{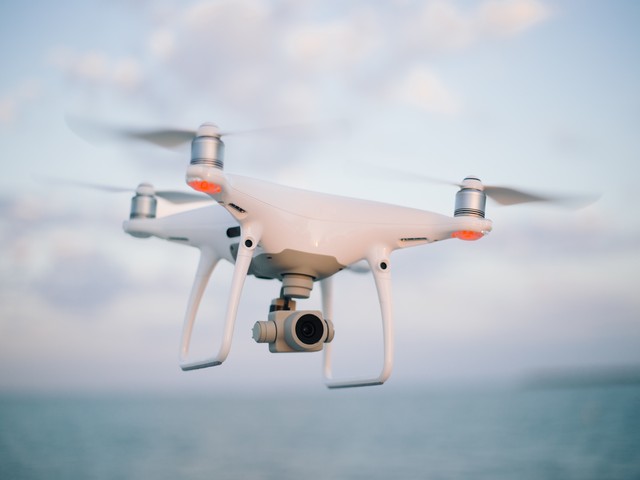}\\[2pt]
\textbf{(b) Civilian drone 2}\par\smallskip
\leftskip0pt\raggedright\figurecaptionfont
\textbf{Low-cost model} (contact 6): \llmquote{``\ldots{} Finally, Contact 6 is the next most urgent (18.1s breach time), and Blue 1 is the best-placed agent to handle it (13.3s). \ldots{}''}\par\smallskip
\textbf{Near-frontier model} (contact 6): \llmquote{``Contacts 0--3 appear to be intruders; contacts 5 and 6 appear to be civilian camera drones and should not consume defenders. \ldots{}''}
\end{minipage}\hfill
\begin{minipage}[t]{0.31\linewidth}\centering
\includegraphics[width=\linewidth,height=0.62\linewidth,keepaspectratio]{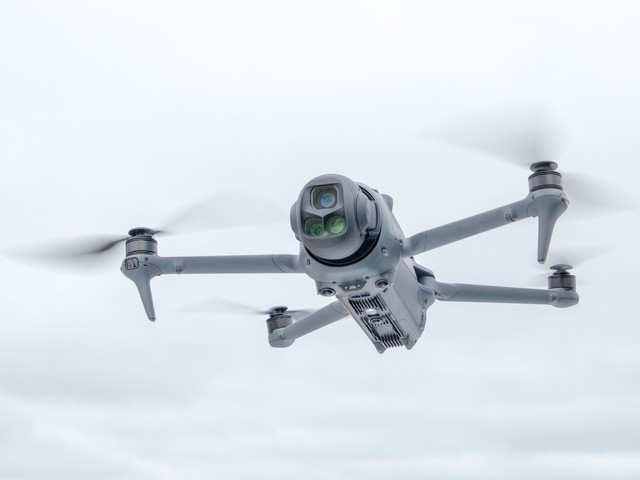}\\[2pt]
\textbf{(c) Civilian drone 3}\par\smallskip
\leftskip0pt\raggedright\figurecaptionfont
\textbf{Low-cost model} (contact 5): \llmquote{``Contact 5 is the primary threat with a predicted breach time of 17.5 seconds. \ldots{}''}\par\smallskip
\textbf{Near-frontier model} (contact 5): \llmquote{``Contact 5 appears to be a civilian camera drone, so avoid spending a defender on interception. \ldots{}''}
\end{minipage}
 \refstepcounter{figure}\label{fig:reason-civilian}
\silegendstart[\bigskip]
\textbf{Supplementary Fig.~\thefigure\sicapsep{}The civilian-drone images and the commander's reasoning about them.}
As in Supplementary Fig.~\ref{fig:reason-birds}. Images: (a) succo, Pixabay, CC0 1.0; (b) J. Sorenson, Pexels, CC0 1.0; (c) H. Soloviev, CC BY 4.0; brand names removed, cropped and resized.
\end{figure}

\begin{figure}[ht]
\centering
\begin{minipage}[t]{0.31\linewidth}\centering
\includegraphics[width=\linewidth,height=0.62\linewidth,keepaspectratio]{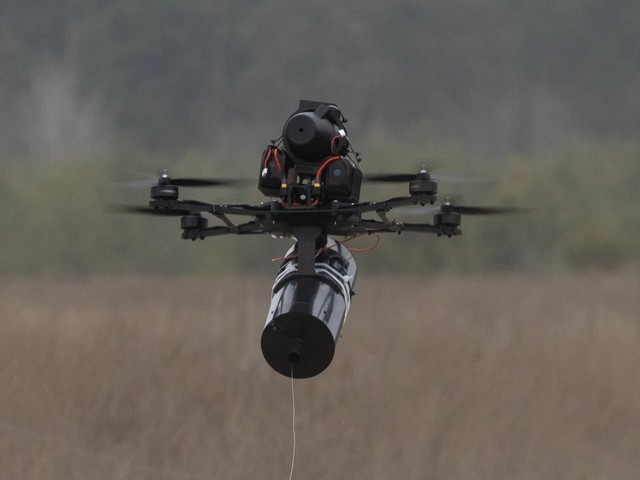}\\[2pt]
\textbf{(a) Attack drone 1}\par\smallskip
\leftskip0pt\raggedright\figurecaptionfont
\textbf{Low-cost model} (contact 0): \llmquote{``\ldots{} Blue 3 is assigned to intercept contact 0 (6.6s optimistic interception), Blue 0 is assigned to contact 1 (9.4s), and Blue 2 is assigned to contact 5 (10.0s). \ldots{}''}\par\smallskip
\textbf{Near-frontier model} (contact 0): \llmquote{``Contacts 0 and 2 appear armed, while 5 and 6 appear to be civilian camera drones. Keep blue 1 pursuing contact 2 and send blue 3 against the more urgent contact 0; their optimistic interception times are both before forecast breach, though continued tracking is essential. \ldots{}''}
\end{minipage}\hfill
\begin{minipage}[t]{0.31\linewidth}\centering
\includegraphics[width=\linewidth,height=0.62\linewidth,keepaspectratio]{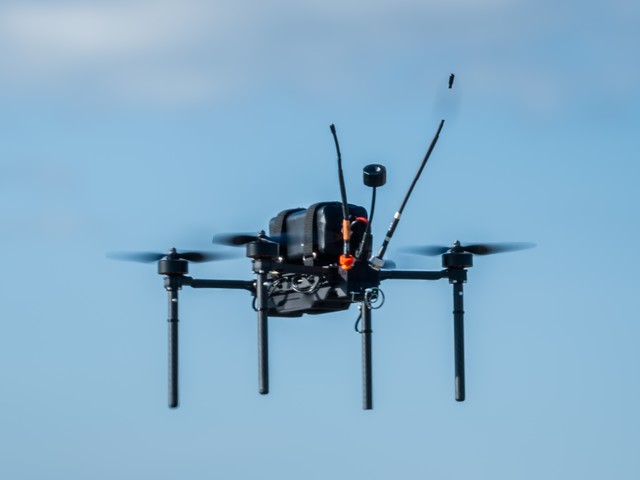}\\[2pt]
\textbf{(b) Attack drone 2}\par\smallskip
\leftskip0pt\raggedright\figurecaptionfont
\textbf{Low-cost model} (contact 1): \llmquote{``\ldots{} Blue 3 is assigned to intercept contact 0 (6.6s optimistic interception), Blue 0 is assigned to contact 1 (9.4s), and Blue 2 is assigned to contact 5 (10.0s). \ldots{}''}\par\smallskip
\textbf{Near-frontier model} (contact 1): \llmquote{``Contacts 0, 1, and 2 appear weapon-equipped; 5 and 6 appear to be civilian camera drones and should not consume defenders. \ldots{} Assign blue 2 to contact 0 and blue 3 to contact 1: their optimistic interception times are 8.0 and 7.4 seconds, both before forecast breaches and with better minimum margin than the reverse pairing. \ldots{}''}
\end{minipage}\hfill
\begin{minipage}[t]{0.31\linewidth}\centering
\includegraphics[width=\linewidth,height=0.62\linewidth,keepaspectratio]{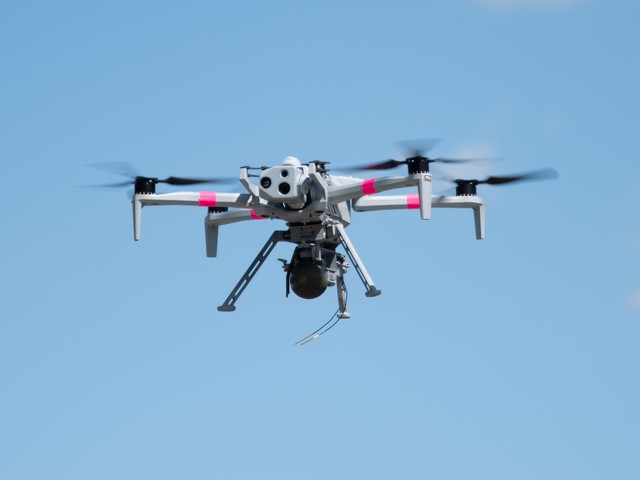}\\[2pt]
\textbf{(c) Attack drone 3}\par\smallskip
\leftskip0pt\raggedright\figurecaptionfont
\textbf{Low-cost model} (contact 2): \llmquote{``\ldots{} Contact 0 (urgency 0.88), Contact 1 (urgency 0.86), Contact 2 (urgency 0.83), and Contact 3 (urgency 0.83) represent the most critical threats. \ldots{} Blue 3 is assigned to intercept Contact 0, Blue 0 to Contact 1, Blue 2 to Contact 2, and Blue 1 to Contact 3, ensuring all four high-urgency threats are addressed by the available team members.''}\par\smallskip
\textbf{Near-frontier model} (contact 2): \llmquote{``Contact 2 appears to carry a munition, while contact 5 appears to be a civilian camera drone. Assign blue 1 to intercept contact 2: its optimistic interception time is 7.7 sec, comfortably before the forecast breach at 17.1 sec. \ldots{}''}
\end{minipage}
 \refstepcounter{figure}\label{fig:reason-attack}
\silegendstart[\bigskip]
\textbf{Supplementary Fig.~\thefigure\sicapsep{}The attack-drone images of the intruders and the commander's reasoning about them.}
As in Supplementary Fig.~\ref{fig:reason-birds}. Images: (a) ArmyInform (armyinform.com.ua), CC BY 4.0, cropped; (b) U.S. Marine Corps photo by Cpl.\ J. C. Dela Torre, public domain; (c) U.S. Army photo by Sgt.\ C. Mackall, public domain; markings removed, cropped and resized.
\end{figure}

\begin{figure}[ht]
\centering
\includegraphics[width=\linewidth]{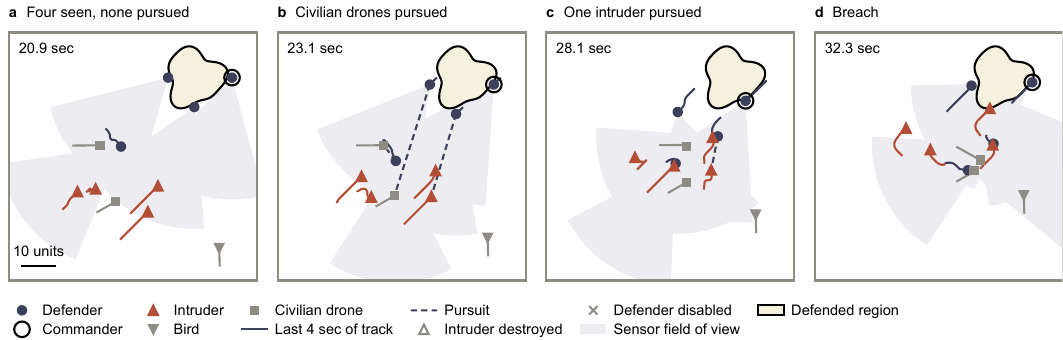}
\refstepcounter{figure}\label{fig:ep-disguise-nowarning}
\silegendstart
\textbf{Supplementary Fig. 10\sicapsep{}An episode of the near-frontier commander with disguised intruders, without the forewarning.}
The setting and seed of Supplementary Fig. 6, but the LLMs saw each intruder as an image of a taxidermy bird drone. \textbf{(a)} All four intruders are seen, and none is pursued. \textbf{(b)} Defenders pursue both civilian drones and one intruder. \textbf{(c)} One intruder is pursued. \textbf{(d)} An intruder breaches, and no intruder was destroyed.
 \end{figure}
 
\begin{figure}[ht]
\centering
\includegraphics[width=\linewidth]{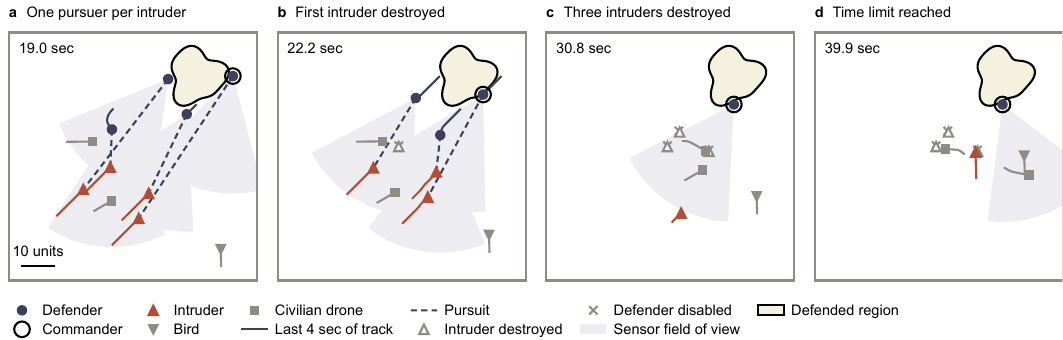}
\refstepcounter{figure}\label{fig:ep-disguise-warning}
\silegendstart
\textbf{Supplementary Fig. 11\sicapsep{}An episode of the near-frontier commander with disguised intruders, with the forewarning.}
The episode of Supplementary Fig. 10 with the one-sentence forewarning in the prompt. \textbf{(a)} Every intruder has its own pursuer. \textbf{(b)} The first intruder is destroyed. \textbf{(c)} Three intruders are destroyed. \textbf{(d)} The episode reaches the 40 sec limit without a breach. No defender pursued a bird or civilian drone.
\end{figure}
\clearpage

\begin{figure}[ht]
\centering
\begin{minipage}[t]{0.47\linewidth}\centering
\includegraphics[width=\linewidth,height=0.62\linewidth,keepaspectratio]{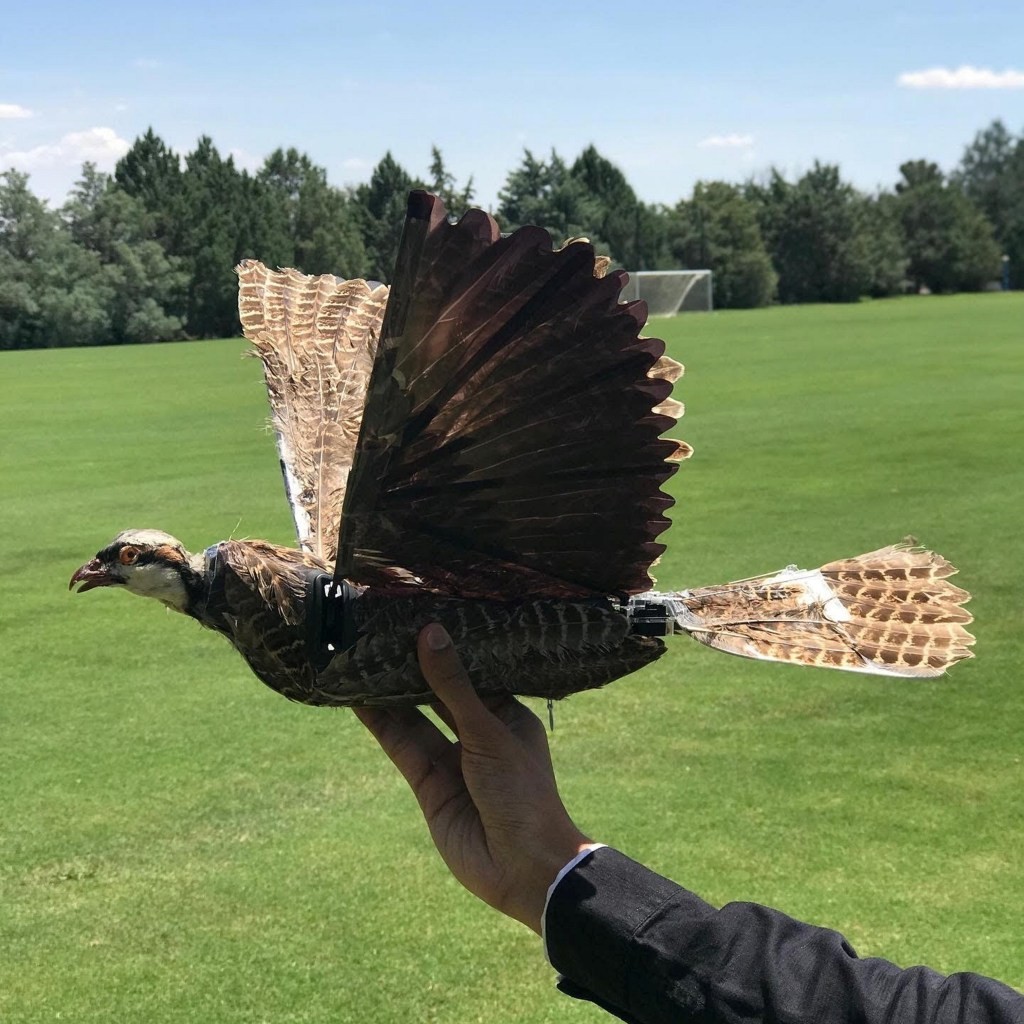}\\[2pt]
\textbf{(a) Bird drone 1}\par\smallskip
\leftskip0pt\raggedright\figurecaptionfont
\textbf{Low-cost model} (contact 0): \llmquote{``\ldots{} Blue 2 is assigned to NEUTRALIZE:0:0 to address the breach threat from Contact 0. \ldots{}''}\par\smallskip
\textbf{Near-frontier model} (contact 0): \llmquote{``Contacts 0--3 appear to be artificial bird-like drones; contact 5 appears to be a civilian quadcopter and should be avoided. \ldots{} Blue 1 can intercept contact 0 much sooner than the other blues, completing coverage of the suspected intruders.''}
\end{minipage}\hfill
\begin{minipage}[t]{0.47\linewidth}\centering
\includegraphics[width=\linewidth,height=0.62\linewidth,keepaspectratio]{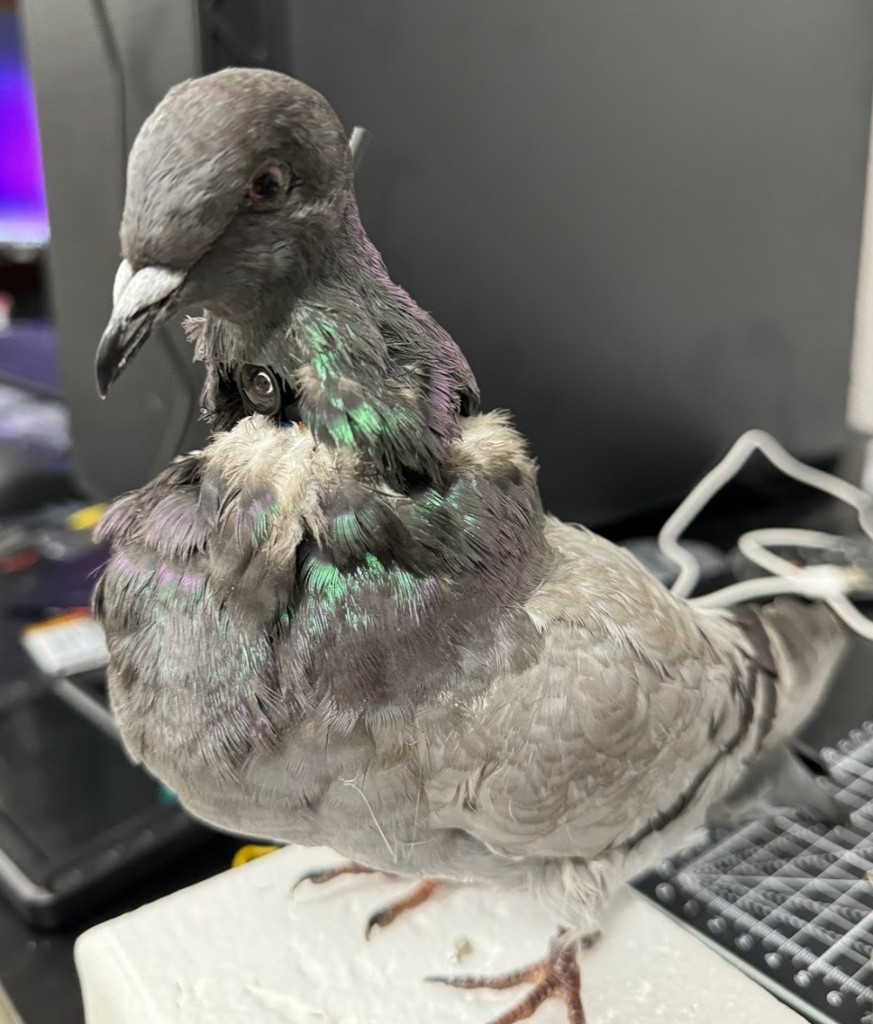}\\[2pt]
\textbf{(b) Bird drone 2}\par\smallskip
\leftskip0pt\raggedright\figurecaptionfont
\textbf{Low-cost model} (contact 1): \llmquote{``\ldots{} Blue 0 is tasked with NEUTRALIZE:1:0 to address Contact 1. \ldots{}''}\par\smallskip
\textbf{Near-frontier model} (contact 1): \llmquote{``Contacts 1 and 2 are birds, and contact 5 appears to be a civilian camera quadcopter. \ldots{}''}
\end{minipage}

\medskip
\begin{minipage}[t]{0.47\linewidth}\centering
\includegraphics[width=\linewidth,height=0.62\linewidth,keepaspectratio]{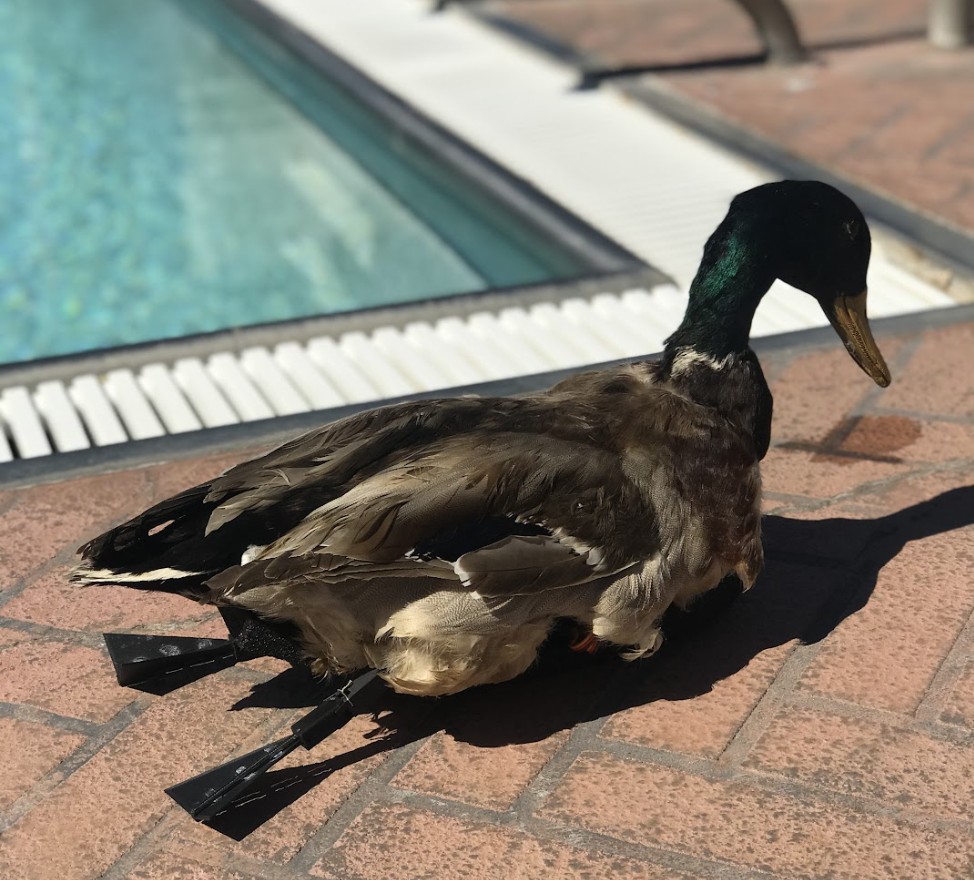}\\[2pt]
\textbf{(c) Bird drone 3}\par\smallskip
\leftskip0pt\raggedright\figurecaptionfont
\textbf{Low-cost model} (contact 2): \llmquote{``\ldots{} Blue 3 is assigned to Contact 0, Blue 0 to Contact 1, Blue 2 to Contact 2, and Blue 1 to Contact 5, maximizing the likelihood of neutralization before any potential breach.''}\par\smallskip
\textbf{Near-frontier model} (contact 2): \llmquote{``Contact 2 is a bird and contact 5 appears to be a civilian camera drone; neither warrants sacrificing a defender. \ldots{}''}
\end{minipage}\hfill
\begin{minipage}[t]{0.47\linewidth}\centering
\includegraphics[width=\linewidth,height=0.62\linewidth,keepaspectratio]{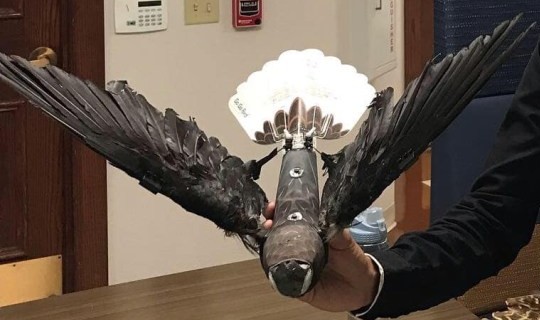}\\[2pt]
\textbf{(d) Bird drone 4}\par\smallskip
\leftskip0pt\raggedright\figurecaptionfont
\textbf{Low-cost model} (contact 3): \llmquote{``The identified mechanical red intruders (drones) are contacts 3, 5, and 6. \ldots{} I am reallocating the blue agents to intercept the drones based on their current proximity and calculated interception times: Blue 3 is closest to contact 5, Blue 0 to contact 3, and Blue 2 to contact 6. \ldots{}''}\par\smallskip
\textbf{Near-frontier model} (contact 3): \llmquote{``Contacts 0--3 appear to be artificial bird-like drones; contact 5 appears to be a civilian quadcopter and should be avoided. Assign blue 3 to the most urgent contact 1, blue 2 to contact 2, and blue 0 to contact 3; their optimistic interception times all precede forecast breaches. \ldots{}''}
\end{minipage}
 \refstepcounter{figure}\label{fig:reason-disguise-u}
\silegendstart[\bigskip]
\textbf{Supplementary Fig.~\thefigure\sicapsep{}The bird-drone images of the disguised intruders and the commander's reasoning about them, without the forewarning.}
As in Supplementary Fig.~\ref{fig:reason-birds}, for the disguise runs: each intruder was shown as one of these images of taxidermy bird drones, and nothing else changed. Images: M. Hassanalian, New Mexico Tech.
\end{figure}
\clearpage
\begin{figure}[ht]
\centering
\begin{minipage}[t]{0.47\linewidth}\centering
\includegraphics[width=\linewidth,height=0.62\linewidth,keepaspectratio]{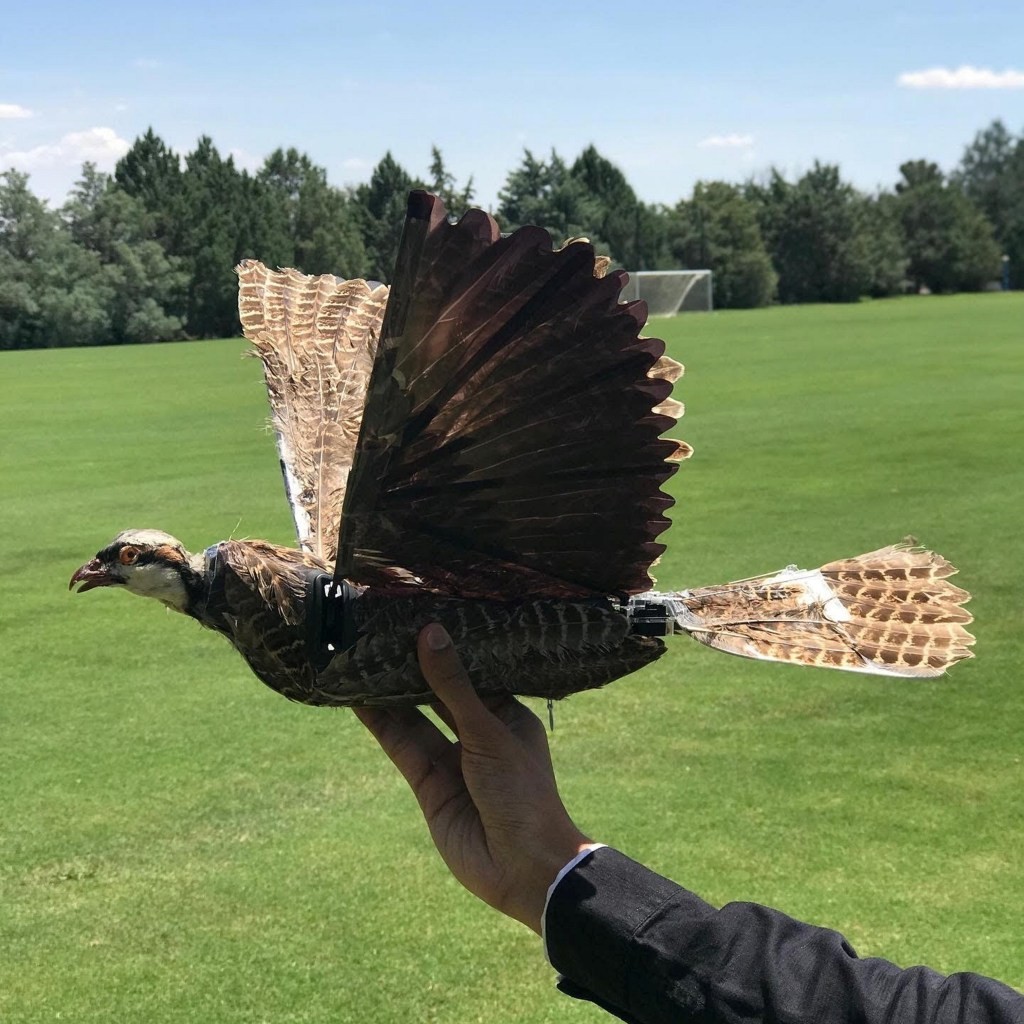}\\[2pt]
\textbf{(a) Bird drone 1}\par\smallskip
\leftskip0pt\raggedright\figurecaptionfont
\textbf{Low-cost model} (contact 0): \llmquote{``Contact 5 (a drone) is the only confirmed threat, as contacts 0 and 1 appear to be birds based on their visual identification. \ldots{}''}\par\smallskip
\textbf{Near-frontier model} (contact 0): \llmquote{``Contacts 0 and 2 appear to be artificial bird decoys and share a direct closing velocity, making them the priority suspected intruders. Keep blue 1 pursuing contact 2 and send blue 3 against the more urgent contact 0; both forecast interceptions precede breach. \ldots{}''}
\end{minipage}\hfill
\begin{minipage}[t]{0.47\linewidth}\centering
\includegraphics[width=\linewidth,height=0.62\linewidth,keepaspectratio]{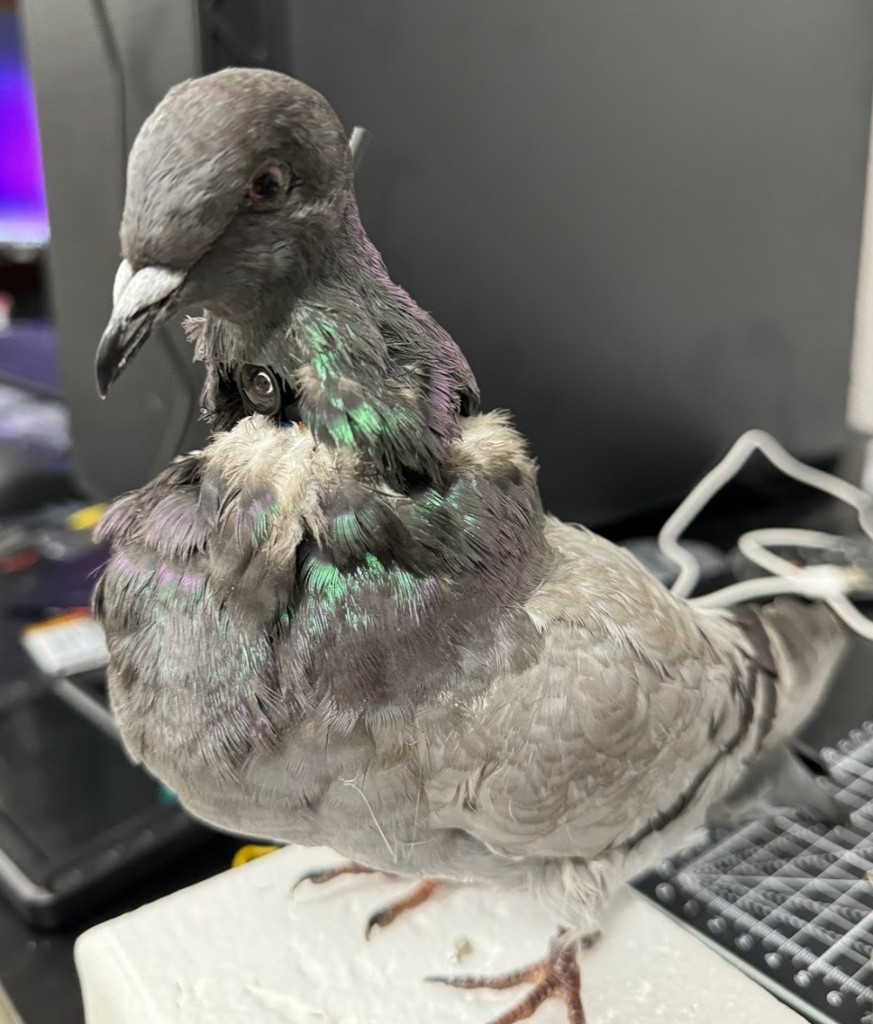}\\[2pt]
\textbf{(b) Bird drone 2}\par\smallskip
\leftskip0pt\raggedright\figurecaptionfont
\textbf{Low-cost model} (contact 1): \llmquote{``Contact 5 (a drone) is the only confirmed threat, as contacts 0 and 1 appear to be birds based on their visual identification. \ldots{}''}\par\smallskip
\textbf{Near-frontier model} (contact 1): \llmquote{``Contacts 0, 1 and 2 show artificial-looking bird features and identical closing velocities, making them the priority suspected intruders. Assign blue 1 to the most urgent contact 0, blue 3 to contact 1, and blue 2 to contact 2; all have forecast interceptions before breach. \ldots{}''}
\end{minipage}

\medskip
\begin{minipage}[t]{0.47\linewidth}\centering
\includegraphics[width=\linewidth,height=0.62\linewidth,keepaspectratio]{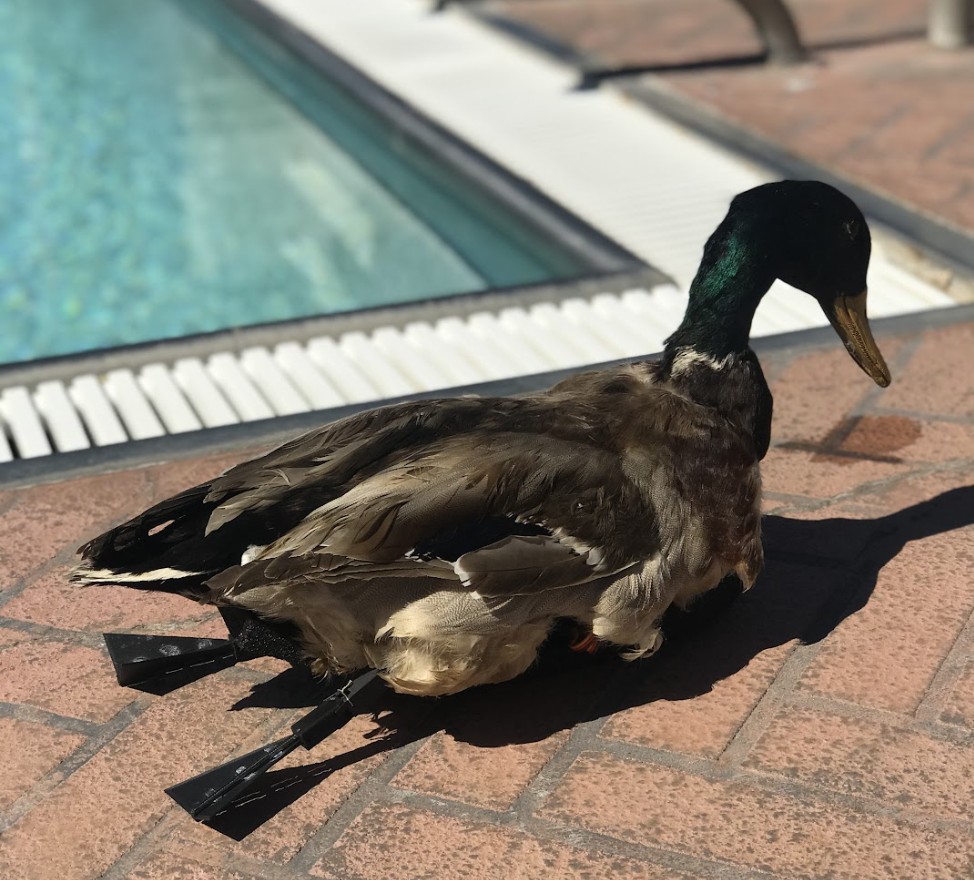}\\[2pt]
\textbf{(c) Bird drone 3}\par\smallskip
\leftskip0pt\raggedright\figurecaptionfont
\textbf{Low-cost model} (contact 2): \llmquote{``\ldots{} Blue 0 is assigned to contact 2 and Blue 1 is assigned to contact 5. \ldots{}''}\par\smallskip
\textbf{Near-frontier model} (contact 2): \llmquote{``Contact 2 appears bird-disguised but has mechanical-looking appendages, making it the stronger intruder candidate. \ldots{}''}
\end{minipage}\hfill
\begin{minipage}[t]{0.47\linewidth}\centering
\includegraphics[width=\linewidth,height=0.62\linewidth,keepaspectratio]{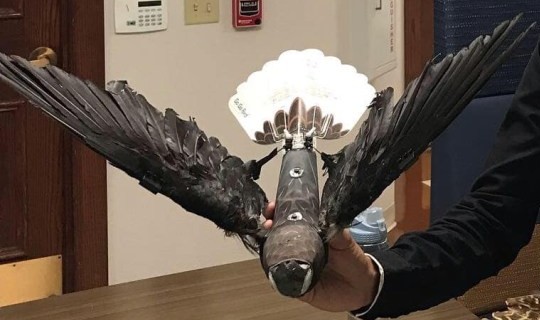}\\[2pt]
\textbf{(d) Bird drone 4}\par\smallskip
\leftskip0pt\raggedright\figurecaptionfont
\textbf{Low-cost model} (contact 3): \llmquote{``\ldots{} Contact 6 and 3 have longer breach times and will be managed as the tactical situation evolves.''}\par\smallskip
\textbf{Near-frontier model} (contact 3): \llmquote{``Contacts 0--3 show artificial or mechanical bird features and are credible disguised intruders; 5 and 6 appear to be civilian quadcopters. \ldots{} Deploy blue 0 against contact 3: its forecast interception precedes breach by 2.7 seconds. \ldots{}''}
\end{minipage}
 \refstepcounter{figure}\label{fig:reason-disguise-w}
\silegendstart[\bigskip]
\textbf{Supplementary Fig.~\thefigure\sicapsep{}The same images and the commander's reasoning about them, with the forewarning.}
As in Supplementary Fig.~\ref{fig:reason-disguise-u}, with the one-sentence forewarning \texttt{Some red intruders may be disguised as birds.} in the prompt. Images: M. Hassanalian, New Mexico Tech.
\end{figure}
\clearpage

\begin{figure}[ht]
\centering
\includegraphics[width=88mm]{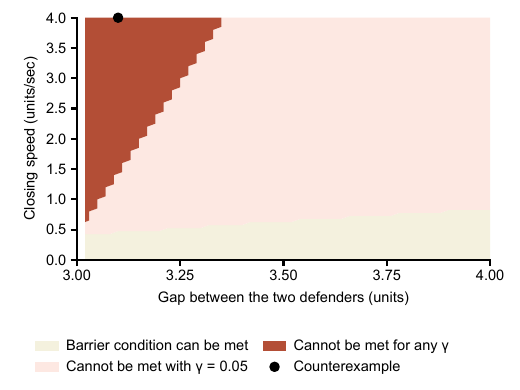}
\refstepcounter{figure}\label{fig:filter}
\silegendstart
\textbf{Supplementary Fig. 14\sicapsep{}The barrier filter's condition cannot be met in some head-on states.}
Each state is the gap and closing speed of two defenders approaching each other head-on just outside the minimum separation of 3 units, so the barrier h, the squared gap minus the squared minimum separation, is positive in every state shown. With both defenders braking as hard as the limits allow, the key shows whether the filter's condition can be met at the next step. In the marked counterexample the gap is 3.1 units and the defenders close at 4 units per second, so h is positive, yet the largest attainable h at the next step is \textminus{}1.49, below the 0.58 the condition requires. On the grid, 612 of 68,850 states have a positive h but no admissible acceleration that keeps h at or above zero, so the condition fails for every choice of $\gamma$. An exact computation over a grid of states, with no simulation.
 \end{figure}
\clearpage
 
\clearpage
\section{Supplementary Tables}

\begingroup
\setstretch{1}
\refstepcounter{table}\label{tab:cells}
\noindent Supplementary Table~\thetable: \textbf{Outcomes of every cell.}
Each row is one defender team with one intruder count and network condition, with one episode per seed (seeds 4001 to 4030).
A breach is an intruder entering the region within the 40\,sec episode.
``Ended at 40\,sec'' counts the episodes without a breach that ran to the time limit with an intruder alive, out of the episodes without a breach.
Each intruder alive at that point is classified over its last 10\,sec as in a stand-off (evading on at least half of the steps, with a defender within 6 units on at least half), closing (otherwise, if its distance to the region fell by more than 1 unit) or other.
Over the main study, intruders breached in 35 of 450 episodes with two intruders and 196 of 450 with four, and 78 of 415 and 126 of 254 episodes without a breach ended at 40\,sec.
Of the intruders alive at 40\,sec, 43 of 85 were closing with two intruders and 113 of 161 with four.
Endings come from deterministic re-simulation of every episode from its cached model answers, and every re-simulated episode matched its record (Methods; Supplementary Table~\ref{tab:repro}).
The episodes of the consensus team at high authority with the near-frontier model are the valid episodes of two runs (Methods).
\par\medskip
\tablebodyfont
\setlength{\tabcolsep}{3.5pt}
\addtocounter{table}{-1}%
\begin{longtable}{@{}lclccc@{}}
\toprule
Team & Intruders & Network condition & Breaches & \makecell{Ended at 40\,sec\\(no breach)} & \makecell{Alive at 40\,sec: stand-off,\\closing, other} \\
\midrule
\endfirsthead
\multicolumn{6}{@{}l}{\normalsize Supplementary Table~\thetable{} (continued)} \\[3pt]
\toprule
Team & Intruders & Network condition & Breaches & \makecell{Ended at 40\,sec\\(no breach)} & \makecell{Alive at 40\,sec: stand-off,\\closing, other} \\
\midrule
\endhead
\bottomrule
\endlastfoot
\multicolumn{6}{@{}l}{\textit{Main study, low-cost LLM}} \\*
Commander & 2 & Ideal network & 1 of 30 & 7 of 29 & 3, 3, 1 \\
 & 2 & Independent loss & 0 of 30 & 9 of 30 & 3, 6, 1 \\
 & 2 & Burst loss & 3 of 30 & 5 of 27 & 1, 3, 2 \\
 & 2 & Falsified reports & 6 of 30 & 10 of 24 & 2, 5, 4 \\
 & 2 & Limited radio range & 8 of 30 & 3 of 22 & 0, 2, 1 \\
 & 4 & Ideal network & 14 of 30 & 6 of 16 & 1, 5, 0 \\
 & 4 & Independent loss & 14 of 30 & 6 of 16 & 1, 5, 2 \\
 & 4 & Burst loss & 18 of 30 & 6 of 12 & 4, 3, 1 \\
 & 4 & Falsified reports & 16 of 30 & 10 of 14 & 5, 9, 1 \\
 & 4 & Limited radio range & 18 of 30 & 7 of 12 & 1, 7, 1 \\
Consensus & 2 & Ideal network & 2 of 30 & 5 of 28 & 2, 2, 1 \\
 & 2 & Independent loss & 3 of 30 & 3 of 27 & 2, 1, 0 \\
 & 2 & Burst loss & 3 of 30 & 7 of 27 & 3, 3, 1 \\
 & 2 & Falsified reports & 1 of 30 & 3 of 29 & 1, 3, 0 \\
 & 2 & Limited radio range & 0 of 30 & 4 of 30 & 3, 1, 0 \\
 & 4 & Ideal network & 7 of 30 & 11 of 23 & 0, 12, 0 \\
 & 4 & Independent loss & 7 of 30 & 7 of 23 & 3, 5, 0 \\
 & 4 & Burst loss & 6 of 30 & 7 of 24 & 2, 3, 2 \\
 & 4 & Falsified reports & 11 of 30 & 12 of 19 & 5, 8, 3 \\
 & 4 & Limited radio range & 8 of 30 & 7 of 22 & 2, 7, 0 \\
Independent & 2 & Ideal network & 1 of 30 & 2 of 29 & 0, 2, 1 \\
 & 2 & Independent loss & 0 of 30 & 10 of 30 & 3, 6, 2 \\
 & 2 & Burst loss & 2 of 30 & 5 of 28 & 2, 4, 0 \\
 & 2 & Falsified reports & 4 of 30 & 3 of 26 & 1, 1, 1 \\
 & 2 & Limited radio range & 1 of 30 & 2 of 29 & 0, 1, 1 \\
 & 4 & Ideal network & 13 of 30 & 11 of 17 & 3, 14, 0 \\
 & 4 & Independent loss & 15 of 30 & 14 of 15 & 3, 14, 2 \\
 & 4 & Burst loss & 17 of 30 & 9 of 13 & 1, 9, 0 \\
 & 4 & Falsified reports & 12 of 30 & 9 of 18 & 3, 10, 0 \\
 & 4 & Limited radio range & 20 of 30 & 4 of 10 & 1, 2, 1 \\
\addlinespace
\multicolumn{6}{@{}l}{\textit{Consensus team at high authority, low-cost LLM}} \\*
Consensus & 2 & Ideal network & 10 of 30 & 9 of 20 & 8, 2, 0 \\
 & 2 & Independent loss & 10 of 30 & 6 of 20 & 4, 2, 1 \\
 & 2 & Burst loss & 6 of 30 & 10 of 24 & 4, 5, 3 \\
 & 2 & Falsified reports & 12 of 30 & 6 of 18 & 1, 2, 5 \\
 & 2 & Limited radio range & 8 of 30 & 8 of 22 & 1, 3, 4 \\
 & 4 & Ideal network & 23 of 30 & 3 of 7 & 1, 2, 0 \\
 & 4 & Independent loss & 21 of 30 & 4 of 9 & 3, 4, 0 \\
 & 4 & Burst loss & 23 of 30 & 5 of 7 & 1, 4, 1 \\
 & 4 & Falsified reports & 20 of 30 & 2 of 10 & 2, 2, 0 \\
 & 4 & Limited radio range & 25 of 30 & 3 of 5 & 0, 5, 0 \\
\addlinespace
\multicolumn{6}{@{}l}{\textit{Consensus team at high authority, near-frontier LLM}} \\*
Consensus & 4 & Ideal network & 7 of 30 & 6 of 23 & 3, 4, 0 \\
\addlinespace
\multicolumn{6}{@{}l}{\textit{Commander team, near-frontier LLM}} \\*
Commander & 2 & Ideal network & 0 of 30 & 7 of 30 & 1, 6, 1 \\
 & 2 & Falsified reports & 1 of 30 & 4 of 29 & 2, 2, 0 \\
 & 4 & Ideal network & 8 of 30 & 10 of 22 & 2, 11, 0 \\
 & 4 & Falsified reports & 3 of 30 & 18 of 27 & 4, 17, 3 \\
\end{longtable}
\endgroup
\clearpage
\begin{table}[ht]
\setstretch{1}
\caption{\textbf{Breach-rate differences between the defender teams.}
Each row is the mean over 30 seeds of each seed's paired difference in breach rate, first team minus second, in percentage points, with episodes matched by seed and intruder count.
Pairs per seed is the number of episode pairs that each seed contributes.
The pooled rows cover the ideal network, independent loss, burst loss and limited radio range with two and four intruders.
Falsified reports enter the comparisons within each team (Supplementary Table~\ref{tab:cond}).
CIs are two-sided 95\% CIs from the $t$ distribution over the 30 seed values (29 degrees of freedom).
The pooled rows form the prespecified primary family (Methods), and the rows by intruder count and condition are secondary.
a, not estimable, because every seed's difference was zero.}
\label{tab:arch}
\centering
\tablebodyfont
\setlength{\tabcolsep}{4pt}
\renewcommand{\arraystretch}{0.9}
\begin{tabular}{@{}lccc@{}}
\toprule
Episodes & \makecell{Pairs\\per seed} & \makecell{Estimate\\(pp)} & 95\% CI ($t$) \\
\midrule
\multicolumn{4}{@{}l}{\textit{Commander minus consensus}} \\
Pooled & 8 & +17 & +9 to +24 \\
2 intruders, pooled & 4 & +3 & \textminus4 to +11 \\
4 intruders, pooled & 4 & +30 & +18 to +42 \\
2 intruders, ideal network & 1 & \textminus3 & \textminus15 to +9 \\
2 intruders, independent loss & 1 & \textminus10 & \textminus21 to +1 \\
2 intruders, burst loss & 1 & 0 & \textminus14 to +14 \\
2 intruders, limited radio range & 1 & +27 & +10 to +43 \\
4 intruders, ideal network & 1 & +23 & \textminus2 to +49 \\
4 intruders, independent loss & 1 & +23 & \textminus2 to +49 \\
4 intruders, burst loss & 1 & +40 & +15 to +65 \\
4 intruders, limited radio range & 1 & +33 & +11 to +56 \\
\addlinespace
\multicolumn{4}{@{}l}{\textit{Commander minus independent}} \\
Pooled & 8 & +3 & \textminus5 to +11 \\
2 intruders, pooled & 4 & +7 & +1 to +13 \\
4 intruders, pooled & 4 & \textminus1 & \textminus15 to +13 \\
2 intruders, ideal network & 1 & 0 & not estimable\textsuperscript{a} \\
2 intruders, independent loss & 1 & 0 & not estimable\textsuperscript{a} \\
2 intruders, burst loss & 1 & +3 & \textminus12 to +19 \\
2 intruders, limited radio range & 1 & +23 & +7 to +39 \\
4 intruders, ideal network & 1 & +3 & \textminus25 to +32 \\
4 intruders, independent loss & 1 & \textminus3 & \textminus34 to +27 \\
4 intruders, burst loss & 1 & +3 & \textminus23 to +30 \\
4 intruders, limited radio range & 1 & \textminus7 & \textminus32 to +19 \\
\addlinespace
\multicolumn{4}{@{}l}{\textit{Consensus minus independent}} \\
Pooled & 8 & \textminus14 & \textminus22 to \textminus6 \\
2 intruders, pooled & 4 & +3 & \textminus2 to +9 \\
4 intruders, pooled & 4 & \textminus31 & \textminus45 to \textminus17 \\
2 intruders, ideal network & 1 & +3 & \textminus9 to +15 \\
2 intruders, independent loss & 1 & +10 & \textminus1 to +21 \\
2 intruders, burst loss & 1 & +3 & \textminus12 to +19 \\
2 intruders, limited radio range & 1 & \textminus3 & \textminus10 to +3 \\
4 intruders, ideal network & 1 & \textminus20 & \textminus45 to +5 \\
4 intruders, independent loss & 1 & \textminus27 & \textminus52 to \textminus1 \\
4 intruders, burst loss & 1 & \textminus37 & \textminus62 to \textminus12 \\
4 intruders, limited radio range & 1 & \textminus40 & \textminus63 to \textminus17 \\
\addlinespace
\bottomrule
\end{tabular}
\end{table}
\clearpage

\begin{sidewaystable}
\setstretch{1}
\caption{\textbf{Allocation measures by team and network condition.}
Each entry is pooled over the 60 episodes of a team and condition (two and four intruders, 30 seeds).
Duplicate commitments is the share of once-per-second checks, at the end of each decision epoch, at which two or more active defenders were committed to the same task.
For the commander team a defender's commitment is the newest order it has accepted, read at the same checks by re-simulation.
Unpursued is the share of known-intruder time, over the steps at which some active defender held a belief about an active intruder, in which no active defender was executing its pursuit.
Unseen is the share of active intruder time in which no defender had yet seen the intruder.
Scouting is the share of defender-steps before the team's first contact in which a defender executed a scouting task.
The last three columns give the share of a pursuer's control steps on which the barrier filter changed its acceleration by more than 0.01, in total and by the kind of agent whose constraint was binding (teammates or disabled defenders, or intruders).
Pooled over the ideal network, independent loss, burst loss and limited radio range, known intruders went unpursued for 14\% of their time in the commander team, 13\% in the consensus team and 42\% in the independent team, and intruders were unseen for 59\% and 41\% of their time in the commander and consensus teams.
Pooled over all five conditions, the filter changed a pursuer's acceleration because of teammates or disabled defenders on 10\%, 7\% and 19\% of its control steps in the three teams.
Coverage, scouting and filter measures come from deterministic re-simulation, with every episode matching its record (Methods).
Descriptive, without CIs.}
\label{tab:coord}
\centering
\tablebodyfont
\setlength{\tabcolsep}{3.5pt}
\begin{tabular}{@{}lccccccc@{}}
\toprule
& & & & & \multicolumn{3}{c}{Filter changed pursuer acceleration} \\
\cmidrule(l){6-8}
Network condition & \makecell{Duplicate\\commitments} & Unpursued & Unseen & \makecell{Scouting before\\first contact} & All & \makecell{Teammates or\\disabled} & Intruders \\
\midrule

\multicolumn{8}{@{}l}{\textit{Commander}} \\*
Ideal network & 0\% & 11\% & 60\% & 0\% & 34\% & 9\% & 26\% \\
Independent loss & 0\% & 15\% & 58\% & 0\% & 36\% & 10\% & 26\% \\
Burst loss & 0\% & 14\% & 59\% & 0\% & 34\% & 8\% & 25\% \\
Falsified reports & 0\% & 15\% & 58\% & 0\% & 37\% & 10\% & 28\% \\
Limited radio range & 3\% & 17\% & 59\% & 0\% & 37\% & 12\% & 25\% \\
\addlinespace
\multicolumn{8}{@{}l}{\textit{Consensus}} \\*
Ideal network & 0\% & 7\% & 40\% & 36\% & 18\% & 6\% & 12\% \\
Independent loss & 0\% & 8\% & 40\% & 40\% & 17\% & 6\% & 10\% \\
Burst loss & 1\% & 9\% & 42\% & 38\% & 18\% & 6\% & 11\% \\
Falsified reports & 0\% & 7\% & 39\% & 36\% & 17\% & 6\% & 10\% \\
Limited radio range & 27\% & 29\% & 42\% & 85\% & 23\% & 10\% & 13\% \\
\addlinespace
\multicolumn{8}{@{}l}{\textit{Independent}} \\*
Ideal network & 68\% & 43\% & 41\% & 40\% & 24\% & 17\% & 7\% \\
Independent loss & 72\% & 42\% & 38\% & 43\% & 28\% & 21\% & 7\% \\
Burst loss & 71\% & 41\% & 40\% & 43\% & 27\% & 19\% & 8\% \\
Falsified reports & 71\% & 40\% & 39\% & 44\% & 24\% & 17\% & 7\% \\
Limited radio range & 48\% & 42\% & 41\% & 87\% & 30\% & 18\% & 11\% \\
\bottomrule
\end{tabular}
\end{sidewaystable}
\clearpage

\clearpage
\begingroup
\setstretch{1}
\refstepcounter{table}\label{tab:cond}
\noindent Supplementary Table~\thetable: \textbf{Change in breach rate under each network condition, by intruder count.}
Each row is the mean over 30 seeds of each seed's paired difference in breach rate, in percentage points, with episodes matched by seed and intruder count, and its two-sided 95\% CI from the $t$ distribution (29 degrees of freedom). Within each condition, the first row pools all three teams, the next rows give each team, and the last rows give the difference between two teams' changes. Intruders ``Both'' pools two and four intruders. The pooled change under falsified reports belongs to the prespecified primary family, and the commander team's change and the commander-minus-consensus difference under falsified reports were added during the study (Methods). The differences between the two loss conditions come from a post hoc analysis. The study stored between-team differences only for falsified reports, limited radio range and the difference between the two loss conditions. Breach rates by cell are in Supplementary Table~\ref{tab:cells}.
\par\medskip
\tablebodyfont
\setlength{\tabcolsep}{4pt}
\addtocounter{table}{-1}%
\begin{longtable}{@{}lcccc@{}}
\toprule
Comparison & Intruders & \makecell{Pairs\\per seed} & \makecell{Estimate\\(pp)} & 95\% CI ($t$) \\
\midrule
\endfirsthead
\multicolumn{5}{@{}l}{\normalsize Supplementary Table~\thetable{} (continued)} \\[3pt]
\toprule
Comparison & Intruders & \makecell{Pairs\\per seed} & \makecell{Estimate\\(pp)} & 95\% CI ($t$) \\
\midrule
\endhead
\bottomrule
\endlastfoot
\multicolumn{5}{@{}l}{\textit{Independent loss minus ideal network}} \\*
All teams & Both & 6 & +1 & \textminus6 to +8 \\
Commander & 2 & 1 & \textminus3 & \textminus10 to +3 \\
 & 4 & 1 & 0 & \textminus26 to +26 \\
 & Both & 2 & \textminus2 & \textminus16 to +13 \\
Consensus & 2 & 1 & +3 & \textminus12 to +19 \\
 & 4 & 1 & 0 & \textminus22 to +22 \\
 & Both & 2 & +2 & \textminus13 to +17 \\
Independent & 2 & 1 & \textminus3 & \textminus10 to +3 \\
 & 4 & 1 & +7 & \textminus17 to +31 \\
 & Both & 2 & +2 & \textminus11 to +14 \\
\addlinespace
\multicolumn{5}{@{}l}{\textit{Burst loss minus ideal network}} \\*
All teams & Both & 6 & +6 & \textminus3 to +16 \\
Commander & 2 & 1 & +7 & \textminus7 to +20 \\
 & 4 & 1 & +13 & \textminus12 to +39 \\
 & Both & 2 & +10 & \textminus6 to +26 \\
Consensus & 2 & 1 & +3 & \textminus12 to +19 \\
 & 4 & 1 & \textminus3 & \textminus26 to +20 \\
 & Both & 2 & 0 & \textminus15 to +15 \\
Independent & 2 & 1 & +3 & \textminus9 to +15 \\
 & 4 & 1 & +13 & \textminus10 to +37 \\
 & Both & 2 & +8 & \textminus6 to +23 \\
\addlinespace
\multicolumn{5}{@{}l}{\textit{Burst loss minus independent loss}} \\*
Commander & Both & 2 & +12 & \textminus3 to +26 \\
Consensus & Both & 2 & \textminus2 & \textminus17 to +13 \\
Independent & Both & 2 & +7 & \textminus7 to +20 \\
Commander minus consensus & Both & 2 & +13 & \textminus9 to +35 \\
Commander minus independent & Both & 2 & +5 & \textminus16 to +26 \\
Independent minus consensus & Both & 2 & +8 & \textminus10 to +27 \\
\addlinespace
\multicolumn{5}{@{}l}{\textit{Falsified reports minus ideal network}} \\*
All teams & Both & 6 & +7 & \textminus2 to +15 \\
Commander & 2 & 1 & +17 & +3 to +31 \\
 & 4 & 1 & +7 & \textminus21 to +34 \\
 & Both & 2 & +12 & \textminus4 to +27 \\
Consensus & 2 & 1 & \textminus3 & \textminus15 to +9 \\
 & 4 & 1 & +13 & \textminus8 to +35 \\
 & Both & 2 & +5 & \textminus6 to +16 \\
Independent & 2 & 1 & +10 & \textminus5 to +25 \\
 & 4 & 1 & \textminus3 & \textminus32 to +25 \\
 & Both & 2 & +3 & \textminus14 to +21 \\
Commander minus consensus & 2 & 1 & +20 & \textminus1 to +41 \\
 & 4 & 1 & \textminus7 & \textminus46 to +32 \\
 & Both & 2 & +7 & \textminus15 to +28 \\
Commander minus independent & 2 & 1 & +7 & \textminus13 to +26 \\
 & 4 & 1 & +10 & \textminus30 to +50 \\
 & Both & 2 & +8 & \textminus14 to +31 \\
Consensus minus independent & 2 & 1 & \textminus13 & \textminus32 to +6 \\
 & 4 & 1 & +17 & \textminus16 to +49 \\
 & Both & 2 & +2 & \textminus18 to +22 \\
\addlinespace
\multicolumn{5}{@{}l}{\textit{Limited radio range minus ideal network}} \\*
All teams & Both & 6 & +9 & +1 to +18 \\
Commander & 2 & 1 & +23 & +7 to +39 \\
 & 4 & 1 & +13 & \textminus16 to +42 \\
 & Both & 2 & +18 & +1 to +36 \\
Consensus & 2 & 1 & \textminus7 & \textminus16 to +3 \\
 & 4 & 1 & +3 & \textminus17 to +24 \\
 & Both & 2 & \textminus2 & \textminus13 to +10 \\
Independent & 2 & 1 & 0 & \textminus10 to +10 \\
 & 4 & 1 & +23 & \textminus2 to +49 \\
 & Both & 2 & +12 & \textminus3 to +26 \\
Commander minus consensus & 2 & 1 & +30 & +13 to +47 \\
 & 4 & 1 & +10 & \textminus23 to +43 \\
 & Both & 2 & +20 & +2 to +38 \\
Commander minus independent & 2 & 1 & +23 & +7 to +39 \\
 & 4 & 1 & \textminus10 & \textminus51 to +31 \\
 & Both & 2 & +7 & \textminus16 to +30 \\
Consensus minus independent & 2 & 1 & \textminus7 & \textminus20 to +7 \\
 & 4 & 1 & \textminus20 & \textminus55 to +15 \\
 & Both & 2 & \textminus13 & \textminus32 to +6 \\
\end{longtable}
\endgroup
\clearpage

\begin{table}[ht]
\setstretch{1}
\caption{\textbf{Exposure to falsified reports.}
Attacked episodes are the episodes with falsified reports (both intruder counts, 30 seeds).
From its first direct observation of an intruder, the compromised defender shifted that intruder's position by 10 units toward the south-west in the first-hand reports it sent (Methods).
A defender accepted a falsified report when the ordinary merge rule stored it in place of its own record.
Defender 0 is the commander in the commander teams.
A pursuit on a falsified position counts an episode in which a defender pursued the targeted intruder while holding the falsified record of it.
Breaches equal the falsified-reports cells of Supplementary Table~\ref{tab:cells}.
Counts come from deterministic re-simulation of the attacked episodes, with every episode matching its record (Methods).
Descriptive, without CIs.}
\label{tab:exposure}
\centering
\tablebodyfont
\setlength{\tabcolsep}{2pt}
\begin{tabular}{@{}>{\raggedright\arraybackslash}p{4.4cm}cccc@{}}%
\toprule
& \multicolumn{3}{c}{Low-cost LLM} & Near-frontier LLM \\
\cmidrule(lr){2-4}\cmidrule(l){5-5}
& Commander & Consensus & Independent & Commander \\
\midrule

Attacked episodes & 60 & 60 & 60 & 60 \\
Episodes with falsified reports sent & 60 & 55 & 60 & 53 \\
Episodes with an acceptance & 56 & 50 & 50 & 41 \\
Episodes with an acceptance by defender 0 & 55 & 50 & 50 & 41 \\
Reports sent per episode (median) & 83.5 & 80 & 89.5 & 68.5 \\
Acceptances per episode (median) & 53 & 93 & 53.5 & 95.5 \\
Episodes with a pursuit on a falsified position & 30 & 24 & 41 & 14 \\
Episodes with a breach & 22 & 12 & 16 & 4 \\
Breaches by the targeted intruder (episodes) & 7 & 3 & 6 & 1 \\
\bottomrule
\end{tabular}
\end{table}
\clearpage

\begin{table}[ht]
\setstretch{1}
\caption{\textbf{Network connectivity under limited radio range.}
Links are limited to 15 units, and messages reach distant defenders by rebroadcast (Methods).
A communication round is connected when the active defenders form one group over the links within radio range.
Connected is the share of rounds that were connected: the mean over episodes, and its 10th, 50th and 90th percentiles over episodes.
Never connected counts the episodes in which no round was connected.
Largest group is the mean share of active defenders in the largest group, and groups is the mean number of groups per round.
A partition is a run of consecutive rounds that were not connected, including a run still open at the end of the episode.
Mean partition is the mean over episodes of each episode's mean partition length (an episode without one counts as 0\,sec), and longest partition is the median over episodes of each episode's longest.
Intruders ``Both'' pools the 30 episodes with two intruders and the 30 with four.
All measures come from deterministic re-simulation, with every episode matching its record (Methods).
Descriptive, without CIs.}
\label{tab:conn}
\centering
\tablebodyfont
\setlength{\tabcolsep}{4pt}
\begin{tabular}{@{}lccccccc@{}}
\toprule
& \multicolumn{2}{c}{Connected (\% of rounds)} & & & & \multicolumn{2}{c}{Partition length (s)} \\
\cmidrule(lr){2-3}\cmidrule(l){7-8}
Intruders & Mean & \makecell{Percentiles\\10 / 50 / 90} & \makecell{Never\\connected} & \makecell{Largest\\group (\%)} & Groups & Mean & Longest \\
\midrule

\multicolumn{8}{@{}l}{\textit{Commander}} \\*
2 & 55 & 47 / 54 / 63 & 0 of 30 & 79 & 1.7 & 7.2 & 10.5 \\
4 & 67 & 47 / 71 / 85 & 0 of 30 & 83 & 1.6 & 5.6 & 4.8 \\
Both & 61 & 47 / 56 / 81 & 0 of 60 & 81 & 1.6 & 6.4 & 8.8 \\
\addlinespace
\multicolumn{8}{@{}l}{\textit{Consensus}} \\*
2 & 2 & 0 / 0 / 9 & 23 of 30 & 43 & 3.1 & 25.6 & 27.2 \\
4 & 27 & 14 / 27 / 41 & 1 of 30 & 57 & 2.5 & 19.6 & 23.0 \\
Both & 14 & 0 / 11 / 35 & 24 of 60 & 50 & 2.8 & 22.6 & 25.2 \\
\addlinespace
\multicolumn{8}{@{}l}{\textit{Independent}} \\*
2 & 8 & 0 / 0 / 26 & 22 of 30 & 46 & 3.0 & 26.3 & 25.4 \\
4 & 31 & 8 / 31 / 51 & 2 of 30 & 60 & 2.5 & 22.7 & 23.1 \\
Both & 19 & 0 / 16 / 51 & 24 of 60 & 53 & 2.8 & 24.5 & 24.2 \\
\bottomrule
\end{tabular}
\end{table}
\clearpage

\begin{sidewaystable}
\setstretch{1}
\caption{\textbf{Bid weights and pursuit multipliers chosen by the LLMs at high authority.}
Each answer at high authority sets the 12 weights of the bid formula (symbols and order as in Supplementary Table~\ref{tab:notation}), each within its allowed range, and a multiplier from 0 to 2 on the pursuit value of each intruder known to the defender (Methods).
Entries are the weights applied after validation and budget scaling, as medians with 25th to 75th percentiles over answers.
Fig.~6a plots five of these weights against the values available at low authority, which Supplementary Table~\ref{tab:notation} lists.
Both teams ran on the ideal network with four intruders and seeds 4001 to 4030 (30 episodes each).
Before contact counts a defender's answers before it first knew of an intruder, and after contact the rest.
The five value weights may sum to at most 3.2, their default total, and the multipliers to at most the number of known intruders; a budget overrun is scaled down in proportion.
A multiplier of 0 leaves the intruder's pursuit bid with only its costs.
Answers were read from the stored responses of the valid episodes, without simulation.
Descriptive, without CIs.}
\label{tab:weights}
\centering
\tablebodyfont
\setlength{\tabcolsep}{3pt}
\begin{tabular}{@{}lccccc@{}}
\toprule
& & \multicolumn{2}{c}{Low-cost LLM} & \multicolumn{2}{c}{Near-frontier LLM} \\
\cmidrule(lr){3-4}\cmidrule(l){5-6}
Weight & \makecell{Default\\(allowed range)} & Before contact & After contact & Before contact & After contact \\
\midrule
Threat value ($w_T$) & 1.20 (0--3) & 1.20 (1.20--1.20) & 1.50 (1.50--1.50) & 0.30 (0.20--0.40) & 0.80 (0.65--1.00) \\
Engagement value ($w_E$) & 1.00 (0--3) & 1.00 (1.00--1.00) & 1.20 (1.20--1.50) & 0.35 (0.25--0.45) & 1.65 (1.40--1.90) \\
Station value ($w_H$) & 0.05 (0--1.5) & 0.10 (0.05--0.50) & 0.10 (0.05--0.10) & 0.20 (0.15--0.25) & 0.05 (0.05--0.15) \\
Station coverage ($w_C$) & 0.50 (0--3) & 0.50 (0.50--0.50) & 0.20 (0.10--0.30) & 1.05 (0.80--1.35) & 0.25 (0.15--0.50) \\
Scouting information ($w_I$) & 0.45 (0--3) & 0.45 (0.10--0.45) & 0.10 (0.05--0.20) & 1.20 (0.90--1.60) & 0.20 (0.15--0.30) \\
Travel cost ($w_D$) & 0.15 (0--0.6) & 0.15 (0.15--0.15) & 0.20 (0.15--0.30) & 0.20 (0.20--0.20) & 0.20 (0.15--0.30) \\
Scouting travel cost ($w_R$) & 0.05 (0--0.3) & 0.05 (0.05--0.05) & 0.05 (0.05--0.10) & 0.18 (0.12--0.25) & 0.10 (0.10--0.15) \\
Belief-age cost ($w_A$) & 0.10 (0--1) & 0.10 (0.10--0.10) & 0.10 (0.10--0.20) & 0.20 (0.15--0.20) & 0.15 (0.10--0.20) \\
Attrition cost ($w_X$) & 0.50 (0--2) & 0.50 (0.50--0.50) & 0.50 (0.50--0.80) & 0.50 (0.50--0.50) & 0.25 (0.15--0.50) \\
Leaving-station cost ($w_L$) & 0.25 (0--1.5) & 0.25 (0.25--0.25) & 0.25 (0.25--0.50) & 0.25 (0.25--0.35) & 0.25 (0.15--0.25) \\
Switching cost ($w_W$) & 0.05 (0--0.5) & 0.05 (0.05--0.05) & 0.10 (0.05--0.20) & 0.08 (0.05--0.12) & 0.08 (0.04--0.12) \\
Continuity cost ($w_P$) & 0.15 (0--1) & 0.15 (0.15--0.15) & 0.20 (0.15--0.50) & 0.15 (0.15--0.15) & 0.25 (0.15--0.40) \\
\midrule
Threat $+$ engagement & 2.20 & 2.20 (2.20--2.20) & 2.70 (2.70--3.00) & 0.70 (0.50--0.85) & 2.60 (2.25--2.80) \\
Station $+$ coverage & 0.55 & 0.60 (0.55--1.00) & 0.30 (0.20--0.40) & 1.30 (0.90--1.60) & 0.35 (0.20--0.65) \\
\midrule
Answers & & 1,823 & 2,022 & 1,140 & 2,345 \\
Value budget scaled & & 0 & 0 & 0 & 0 \\
Multiplier budget scaled & & 0 & 43 & 0 & 0 \\
Pursuit multipliers set & & -- & 6,276 & -- & 6,916 \\
\quad equal to 0 & & -- & 135 & -- & 44 \\
\quad below 1 & & -- & 2,876 & -- & 3,049 \\
\quad above 1 & & -- & 2,093 & -- & 3,216 \\
\bottomrule
\end{tabular}
\end{sidewaystable}
\clearpage

\begin{table}[ht]
\setstretch{1}
\caption{\textbf{High authority and model capability: comparisons.}
Each row is a paired difference, first team minus second, in percentage points, with episodes matched by seed and intruder count. In the first two blocks, breach rates and unseen time pool the ideal network, independent loss, burst loss and limited radio range. Rows with several pairs per seed give the mean over 30 seeds of each seed's difference with a two-sided 95\% CI from the $t$ distribution. Rows with one pair per seed (the near-frontier model at high authority, and the near-frontier commander by intruder count) give Newcombe's hybrid score CI (method 10), marked N. On the ideal network with four intruders, intruders were unseen for 39\%, 54\% and 40\% of their active time for the consensus team at high authority with the near-frontier and the low-cost model and at low authority, and known intruders went unpursued for 24\%, 11\% and 8\% of their time. Over the four conditions, intruders were unseen for 53\% of their time at high authority and 41\% at low authority, both with the low-cost model. The 30 episodes of the near-frontier model at high authority come from two runs, 22 seeds from the first and 8 from the second (Methods).}
\label{tab:authority}
\centering
\tablebodyfont
\setlength{\tabcolsep}{3pt}
\begin{tabular}{@{}llccc@{}}
\toprule
Comparison & Intruders & \makecell{Pairs\\per seed} & \makecell{Estimate\\(pp)} & 95\% CI \\
\midrule

\multicolumn{5}{@{}l}{\textit{High minus low authority, consensus team, low-cost LLM}} \\*
Breach rate & Both & 8 & +38 & +30 to +45 \\
 & 2 & 4 & +22 & +12 to +31 \\
 & 4 & 4 & +53 & +44 to +62 \\
Change under falsified reports & Both & 2 & \textminus7 & \textminus28 to +15 \\
 & 2 & 1 & +10 & \textminus18 to +38 \\
 & 4 & 1 & \textminus23 & \textminus55 to +9 \\
Intruder time unseen & Both & 8 & +12 & +10 to +14 \\
\addlinespace
\multicolumn{5}{@{}l}{\textit{Consensus team at high authority, low-cost LLM, minus the other main-study teams}} \\*
Minus commander & Both & 8 & +21 & +12 to +29 \\
Minus independent & Both & 8 & +24 & +17 to +30 \\
\addlinespace
\multicolumn{5}{@{}l}{\textit{Consensus team at high authority, near-frontier LLM, ideal network, four intruders}} \\*
Minus the low-cost LLM at high authority & 4 & 1 & \textminus53 & \textminus70 to \textminus28\textsuperscript{N} \\
Minus the low-cost LLM at low authority & 4 & 1 & 0 & \textminus24 to +24\textsuperscript{N} \\
Minus near-frontier commander & 4 & 1 & \textminus3 & \textminus23 to +17\textsuperscript{N} \\
\addlinespace
\multicolumn{5}{@{}l}{\textit{Near-frontier minus low-cost commander}} \\*
Ideal network & Both & 2 & \textminus12 & \textminus26 to +3 \\
 & 2 & 1 & \textminus3 & \textminus17 to +8\textsuperscript{N} \\
 & 4 & 1 & \textminus20 & \textminus42 to +6\textsuperscript{N} \\
Falsified reports & Both & 2 & \textminus30 & \textminus42 to \textminus18 \\
 & 2 & 1 & \textminus17 & \textminus34 to +1\textsuperscript{N} \\
 & 4 & 1 & \textminus43 & \textminus62 to \textminus19\textsuperscript{N} \\
Change under falsified reports & Both & 2 & \textminus18 & \textminus38 to +1 \\
 & 2 & 1 & \textminus13 & \textminus30 to +3 \\
 & 4 & 1 & \textminus23 & \textminus60 to +13 \\
\bottomrule
\end{tabular}
\end{table}
\clearpage

\begin{table}[ht]
\setstretch{1}
\caption{\textbf{Episodes with birds and civilian drones, by seed.}
Ideal network, four intruders, one bird and two civilian drones per episode, seeds 4001 to 4005. Without refers to the matched episode of the same team without birds and civilian drones (the main study, or the team's own runs for the consensus team at high authority and the near-frontier commander). An episode ends at a breach, at the time limit of 40\,sec, or when every intruder has been destroyed or has left the arena. Every defender that intercepts an intruder or collides with a bird or civilian drone is disabled, so the defenders lost to intruders equal the intruders destroyed. The low-cost teams lost 3 to 4 defenders to birds and civilian drones each, and the near-frontier model lost 0 at high authority and 0 as commander. Counts at five seeds are descriptive, without CIs.}
\label{tab:decoyseeds}
\centering
\tablebodyfont
\setlength{\tabcolsep}{4pt}
\begin{tabular}{@{}lcclccc@{}}
\toprule
& \multicolumn{2}{c}{Breach} & & & \multicolumn{2}{c}{Defenders lost to} \\
\cmidrule(lr){2-3}\cmidrule(l){6-7}
Seed & Without & With & Ending with them & \makecell{Intruders\\destroyed} & Birds & \makecell{Civilian\\drones} \\
\midrule
\multicolumn{7}{@{}l}{\textit{Consensus at low authority, low-cost LLM}} \\*
4001 & no & yes & breach & 2 & 0 & 1 \\
4002 & yes & yes & breach & 1 & 0 & 0 \\
4003 & no & no & time limit & 3 & 0 & 1 \\
4004 & no & yes & breach & 2 & 0 & 1 \\
4005 & yes & yes & breach & 2 & 0 & 1 \\
\textit{Total} & 2 & 4 & & 10 & 0 & 4 \\
\addlinespace
\multicolumn{7}{@{}l}{\textit{Commander, low-cost LLM}} \\*
4001 & no & no & time limit & 2 & 0 & 1 \\
4002 & no & no & time limit & 2 & 0 & 1 \\
4003 & no & yes & breach & 1 & 0 & 1 \\
4004 & no & no & time limit & 2 & 1 & 0 \\
4005 & no & no & time limit & 2 & 0 & 1 \\
\textit{Total} & 0 & 1 & & 9 & 1 & 4 \\
\addlinespace
\multicolumn{7}{@{}l}{\textit{Consensus at high authority, low-cost LLM}} \\*
4001 & yes & yes & breach & 0 & 0 & 0 \\
4002 & yes & no & time limit & 2 & 0 & 1 \\
4003 & yes & no & time limit & 3 & 0 & 0 \\
4004 & yes & no & time limit & 3 & 0 & 1 \\
4005 & yes & no & time limit & 1 & 0 & 1 \\
\textit{Total} & 5 & 1 & & 9 & 0 & 3 \\
\addlinespace
\multicolumn{7}{@{}l}{\textit{Consensus at high authority, near-frontier LLM}} \\*
4001 & yes & no & all intruders resolved & 4 & 0 & 0 \\
4002 & no & no & all intruders resolved & 4 & 0 & 0 \\
4003 & yes & no & all intruders resolved & 4 & 0 & 0 \\
4004 & no & no & all intruders resolved & 4 & 0 & 0 \\
4005 & no & no & all intruders resolved & 4 & 0 & 0 \\
\textit{Total} & 2 & 0 & & 20 & 0 & 0 \\
\addlinespace
\multicolumn{7}{@{}l}{\textit{Commander, near-frontier LLM}} \\*
4001 & no & no & time limit & 3 & 0 & 0 \\
4002 & no & no & time limit & 2 & 0 & 0 \\
4003 & yes & no & time limit & 3 & 0 & 0 \\
4004 & no & no & time limit & 3 & 0 & 0 \\
4005 & no & no & time limit & 3 & 0 & 0 \\
\textit{Total} & 1 & 0 & & 14 & 0 & 0 \\
\bottomrule
\end{tabular}

\end{table}
\clearpage

\begin{table}[ht]
\setstretch{1}
\caption{\textbf{Answers about each kind of contact with birds and civilian drones present.}
Counts over the five episodes with birds and civilian drones of each team (ideal network, four intruders, seeds 4001 to 4005) that underlie the shares of contacts dropped or deprioritized in Fig.~7c (main paper) and Supplementary Table~\ref{tab:airspace}. For the consensus team at low authority, each count is one contact the defender held a belief about, in one answer, with the priority the answer gave it (medium if the answer left the contact out). At high authority, each count is one contact the defender held a belief about, in one answer, with the multiplier the answer wrote (1 if left out), before validation and budget scaling. For the commanders, each count is one contact in the commander's catalog in one epoch, counted as assigned if the answer gave any defender a pursuit task on it. The near-frontier commander assigned a defender to a bird or civilian drone in 2 of the 361 contact-epochs in which it saw one, and the near-frontier model at high authority dropped 100\% of birds and 91\% of civilian drones in its answers. Answers were tallied from the stored responses as written, before validation. Counts at five seeds are descriptive, without CIs.}
\label{tab:decoyanswers}
\centering
\tablebodyfont
\begin{tabular}{@{}lccc@{}}
\toprule
Answer & Intruders & Birds & Civilian drones \\
\midrule
\multicolumn{4}{@{}l}{\textit{Consensus at low authority, low-cost LLM}} \\*
High priority & 459 & 19 & 172 \\
Medium priority & 475 & 41 & 222 \\
Low priority & 83 & 235 & 55 \\
\addlinespace
\multicolumn{4}{@{}l}{\textit{Commander, low-cost LLM}} \\*
Defender assigned & 192 & 14 & 76 \\
No defender assigned & 120 & 88 & 107 \\
\addlinespace
\multicolumn{4}{@{}l}{\textit{Consensus at high authority, low-cost LLM}} \\*
Multiplier above 1 & 551 & 22 & 164 \\
Multiplier 1 & 366 & 84 & 239 \\
Multiplier below 1 & 244 & 200 & 204 \\
Multiplier 0 & 6 & 50 & 18 \\
\addlinespace
\multicolumn{4}{@{}l}{\textit{Consensus at high authority, near-frontier LLM}} \\*
Multiplier above 1 & 893 & 0 & 4 \\
Multiplier 1 & 37 & 0 & 15 \\
Multiplier below 1 & 103 & 0 & 34 \\
Multiplier 0 & 6 & 269 & 550 \\
\addlinespace
\multicolumn{4}{@{}l}{\textit{Commander, near-frontier LLM}} \\*
Defender assigned & 280 & 0 & 2 \\
No defender assigned & 55 & 123 & 236 \\
\bottomrule
\end{tabular}
\end{table}
\clearpage

\begin{table}[ht]
\setstretch{1}
\caption{\textbf{Outcomes and treatment of contacts in airspace with birds and civilian drones, and with disguised intruders.}
Five seeds (4001 to 4005), ideal network, four intruders, one bird and two civilian drones per episode. In the lower block the intruders were disguised intruders, each shown to the LLMs as an image of a bird drone (main paper's Fig.~7b), without and with the forewarning sentence (Methods); birds, civilian drones and everything else were as in the upper block. Breaches are counts of five episodes, without birds and civilian drones (the matched episodes of the main study) and with them. Defenders lost to birds and civilian drones are summed over the five episodes; without the disguise no episode lost more than one. Intruders destroyed are counted of the 20 intruders in the five episodes. ``Dropped or deprioritized'' is the share of the LLMs' answers or epochs that dropped a contact or deprioritized it, by contact kind: deprioritized for the consensus team at low authority (a low priority, which halves the value of the contact's pursuit but keeps its pursuit task); dropped at high authority (a pursuit multiplier of zero) and for the commanders (a contact-epoch in which no defender was assigned to the contact). Counts and shares are descriptive; no confidence intervals are given at five seeds.}
\label{tab:airspace}
\centering
\tablebodyfont
\setlength{\tabcolsep}{3pt}
\begin{tabular*}{\linewidth}{@{\extracolsep{\fill}}p{0.24\linewidth}ccccc@{}}
\toprule
& \multicolumn{3}{c}{Low-cost LLM} & \multicolumn{2}{c}{Near-frontier LLM} \\
\cmidrule(lr){2-4}\cmidrule(l){5-6}
& \shortstack{Consensus,\\low authority} & Commander & \shortstack{Consensus,\\high authority} & \shortstack{Consensus,\\high authority} & Commander \\
\midrule
\raggedright Breaches, main study (of~5) & 2 & 0 & 5 & 2 & 1 \\
\raggedright Breaches with birds and civilian drones (of~5) & 4 & 1 & 1 & 0 & 0 \\
\raggedright Defenders lost to birds & 0 & 1 & 0 & 0 & 0 \\
\raggedright Defenders lost to civilian drones & 4 & 4 & 3 & 0 & 0 \\
\midrule
\raggedright \textit{Dropped or deprioritized (\%)} & & & & & \\
\raggedright \quad Intruders & 8 & 38 & 1 & 1 & 16 \\
\raggedright \quad Birds & 80 & 86 & 14 & 100 & 100 \\
\raggedright \quad Civilian drones & 12 & 58 & 3 & 91 & 99 \\
\raggedright Intruders destroyed (of~20) & 10 & 9 & 9 & 20 & 14 \\
\midrule
\multicolumn{6}{@{}l}{\textit{Disguised intruders: without / with forewarning}} \\
\raggedright Breaches (of~5) & 5 / 3 & 3 / 5 & 5 / 5 & 3 / 0 & 3 / 1 \\
\raggedright Intruders destroyed (of~20) & 4 / 4 & 8 / 10 & 4 / 5 & 4 / 14 & 2 / 14 \\
\raggedright Defenders lost to birds and civilian drones & 5 / 11 & 2 / 4 & 1 / 4 & 2 / 0 & 2 / 1 \\
\raggedright \textit{Dropped or deprioritized (\%)} & & & & & \\
\raggedright \quad Intruders & 44 / 56 & 51 / 42 & 10 / 3 & 59 / 0 & 61 / 25 \\
\raggedright \quad Birds & 48 / 72 & 75 / 74 & 18 / 5 & 100 / 23 & 100 / 100 \\
\raggedright \quad Civilian drones & 18 / 7 & 47 / 42 & 5 / 1 & 72 / 33 & 77 / 93 \\
\bottomrule
\end{tabular*}
\end{table}

\clearpage

\clearpage
\begingroup
\setstretch{1}
\refstepcounter{table}\label{tab:notation}
\noindent Supplementary Table~\thetable: \textbf{Notation.}
Every symbol in the Methods of the paper, grouped by the part of Methods that introduces it.
Value or default gives each run setting and the default of each weight, which is that of the balanced profile.
The last two columns give the values that the consensus team's LLM can set.
At low authority it changes $w_C$, $w_I$, $w_X$, $w_L$ and $T$ only together, by choosing one of four weight profiles, and sets a priority for each known intruder (Methods).
At high authority it enters each value within the range shown, the five value weights may sum to at most 3.2, their default total, and the multipliers to at most the number of known intruders.
A dash marks a value that the LLM cannot change, and an empty cell a quantity that the defender computes or an index.
\par\medskip
\tablebodyfont
\setlength{\tabcolsep}{4pt}
\addtocounter{table}{-1}%
\begin{longtable}{@{}lp{0.31\linewidth}lll@{}}
\toprule
Symbol & Meaning & \makecell[l]{Value or\\default} & \makecell[l]{Low\\authority} & \makecell[l]{High\\authority} \\
\midrule
\endfirsthead
\multicolumn{5}{@{}l}{\normalsize Supplementary Table~\thetable{} (continued)} \\[3pt]
\toprule
Symbol & Meaning & \makecell[l]{Value or\\default} & \makecell[l]{Low\\authority} & \makecell[l]{High\\authority} \\
\midrule
\endhead
\bottomrule
\endlastfoot

\multicolumn{5}{@{}l}{\emph{Game and dynamics}}\\
$\Delta t$ & \raggedright simulation step & 0.1\,sec &  &  \tabularnewline
$k$ & \raggedright step index &  &  &  \tabularnewline
$x$ & \raggedright state of an agent, $[p^\top, v^\top, \theta]^\top$ &  &  &  \tabularnewline
$p$ & \raggedright planar position &  &  &  \tabularnewline
$v$; $v_x$, $v_y$ & \raggedright velocity; its components & $|v_x|, |v_y| \le \,2$\,units\,sec$^{-1}$ &  &  \tabularnewline
$\theta$ & \raggedright sensor heading &  &  &  \tabularnewline
$u$ & \raggedright input of an agent, $[a^\top, \omega]^\top$ &  &  &  \tabularnewline
$a$; $a_x$, $a_y$ & \raggedright translational acceleration; its components & $|a_x|, |a_y| \le \,2$\,units\,sec$^{-2}$ &  &  \tabularnewline
$\omega$ & \raggedright yaw rate & $|\omega| \le 2\pi$\,rad\,sec$^{-1}$ &  &  \tabularnewline
\addlinespace
\multicolumn{5}{@{}l}{\emph{Barrier filter}}\\
$r$ & \raggedright relative position of two agents &  &  &  \tabularnewline
$D$ & \raggedright minimum separation & 3 units &  &  \tabularnewline
$h$; $h_k$ & \raggedright barrier, $\lVert r \rVert^2 - D^2$; its value at step $k$ &  &  &  \tabularnewline
$\gamma$ & \raggedright rate in the barrier condition & 0.05 &  &  \tabularnewline
\addlinespace
\multicolumn{5}{@{}l}{\emph{Tasks and bids}}\\
$q$ & \raggedright candidate task &  &  &  \tabularnewline
$j$ & \raggedright intruder &  &  &  \tabularnewline
$s$ & \raggedright station &  &  &  \tabularnewline
$g$ & \raggedright cell of the coverage grid &  &  &  \tabularnewline
$b_q$ & \raggedright bid for task $q$ &  &  &  \tabularnewline
$\mathbf{1}[\cdot]$ & \raggedright indicator: 1 if the condition in brackets holds, else 0 &  &  &  \tabularnewline
$V_q$, $C_q$ & \raggedright value and cost of task $q$ &  &  &  \tabularnewline
$K_q$ & \raggedright change cost, part of $C_q$ &  &  &  \tabularnewline
$Q$ & \raggedright tasks marked as preferred & empty & up to three & up to three \tabularnewline
$\beta$ & \raggedright bonus for each preferred task & 0.2 & -- & -- \tabularnewline
\addlinespace
\multicolumn{5}{@{}l}{\emph{Constants}}\\
$L$ & \raggedright length scale, a quarter of the arena & 25 units & -- & -- \tabularnewline
$T$ & \raggedright urgency time scale & 60\,sec & 60 or 90\,sec & -- \tabularnewline
\addlinespace
\multicolumn{5}{@{}l}{\emph{Computed by the defender for each task}}\\
$d_q$ & \raggedright distance to task $q$ &  &  &  \tabularnewline
$\tau_j$ & \raggedright forecast time for intruder $j$ to reach the region &  &  &  \tabularnewline
$\delta_j$ & \raggedright distance from intruder $j$ to the region &  &  &  \tabularnewline
$U_j$ & \raggedright urgency of intruder $j$ &  &  &  \tabularnewline
$O_j$ & \raggedright opportunity to intercept intruder $j$ in time &  &  &  \tabularnewline
$m_s$ & \raggedright unwatched approach sector that station $s$ would watch &  &  &  \tabularnewline
$I_g$ & \raggedright unwatched approach sector swept on the way to cell $g$ &  &  &  \tabularnewline
$\alpha_j$ & \raggedright age of the belief about intruder $j$, as a distance &  &  &  \tabularnewline
$n$ & \raggedright number of active teammates &  &  &  \tabularnewline
$\ell_q$, $m_0$ & \raggedright leaving the held station, and its coverage &  &  &  \tabularnewline
$\sigma_q$ & \raggedright switching to a different task &  &  &  \tabularnewline
$c_q$ & \raggedright dropping the pursuit of a known intruder &  &  &  \tabularnewline
\addlinespace
\multicolumn{5}{@{}l}{\emph{Value weights}}\\
$w_T$ & \raggedright threat value of every known intruder & 1.2 & -- & 0--3 \tabularnewline
$w_E$ & \raggedright engagement value, on $\pi_j U_j O_j$ & 1 & -- & 0--3 \tabularnewline
$w_H$ & \raggedright station value & 0.05 & -- & 0--1.5 \tabularnewline
$w_C$ & \raggedright weight on station coverage $m_s$ & 0.5 & 0.5 or 0.8 & 0--3 \tabularnewline
$w_I$ & \raggedright weight on scouting information $I_g$ & 0.45 & 0.45 or 0.75 & 0--3 \tabularnewline
\addlinespace
\multicolumn{5}{@{}l}{\emph{Cost weights}}\\
$w_D$ & \raggedright travel cost of pursuits and stations & 0.15 & -- & 0--0.6 \tabularnewline
$w_R$ & \raggedright travel cost of scouting & 0.05 & -- & 0--0.3 \tabularnewline
$w_A$ & \raggedright belief-age cost, on $\alpha_j$ & 0.1 & -- & 0--1 \tabularnewline
$w_X$ & \raggedright attrition cost & 0.5 & 0.2, 0.5 or 0.9 & 0--2 \tabularnewline
$w_L$ & \raggedright leaving-station cost & 0.25 & 0.25 or 0.4 & 0--1.5 \tabularnewline
$w_W$ & \raggedright switching cost & 0.05 & -- & 0--0.5 \tabularnewline
$w_P$ & \raggedright continuity cost & 0.15 & -- & 0--1 \tabularnewline
\addlinespace
\multicolumn{5}{@{}l}{\emph{Set for each intruder}}\\
$\pi_j$ & \raggedright priority of intruder $j$ & 1 & 0.5, 1 or 2 & -- \tabularnewline
$\rho_j$ & \raggedright pursuit multiplier of intruder $j$ & 1 & -- & 0--2 \tabularnewline
\addlinespace
\multicolumn{5}{@{}l}{\emph{Airspace challenges}}\\
$N$ & \raggedright identifier of a contact, in the label of its image &  &  &  \tabularnewline
\end{longtable}
\endgroup
\clearpage

\begin{table}[ht]
\setstretch{1}
\caption{\textbf{Model profiles, calls and cost.}
Upper part: the two frozen model profiles, called through OpenRouter.
The endpoint revision is the endpoint's name as listed when the profile was frozen; the provider tier is the one requested, with fallback to other providers disabled.
Every call used a strict JSON response schema whose first field is the reasoning.
The near-frontier endpoint takes no temperature, so none was sent.
List prices are for prompt / completion tokens.
Each call allowed retries after a rate limit (up to five), a connection error, a server error, an empty answer or a provider abort (up to two each) and a timeout (one).
Lower part: calls per run, counted from the episode records of the valid episodes, by the interface that made them, and the provider-reported cost written to the records.
Calls are synchronous, so their latency does not affect the game; the median latency of a commander call was 2.0\,sec for the low-cost model and 5.7\,sec for the near-frontier model on the ideal network and under falsified reports.
In the run of the near-frontier model at high authority, the 6 invalid episodes of the first run and their calls are excluded (Methods).}
\label{tab:profiles}
\centering
\tablebodyfont
\setlength{\tabcolsep}{3pt}
\begin{tabular}{@{}lll@{}}
\toprule
& Low-cost LLM & Near-frontier LLM \\
\midrule

Model id & \texttt{google/gemini-3.1-flash-lite} & \texttt{openai/gpt-6.1-sol} \\
Endpoint revision & \texttt{gemini-3.1-flash-lite-20260507} & \texttt{gpt-6.1-sol-20260929} \\
Provider tier & flex & standard \\
Reasoning effort & medium & medium \\
Temperature & 1.0 & not sent \\
Output allowance (tokens) & 8,000 & 25,000 \\
Timeout (s) & 300 & 300 \\
List price (US\$ per million tokens) & 0.125 / 0.75 & 2 / 10 \\
\bottomrule
\end{tabular}

\bigskip
\begin{tabular}{@{}>{\raggedright\arraybackslash}p{3.4cm}cccccc@{}}%
\toprule
& & \multicolumn{3}{c}{Calls by interface} & & \\
\cmidrule(lr){3-5}
Run & Episodes & Defenders & Commander & Intruders & All calls & Cost (US\$) \\
\midrule
Main study (three teams) & 900 & 70,033 & 10,657 & 80,098 & 160,788 & 104.63 \\
Consensus team at high authority, low-cost LLM & 300 & 39,804 & -- & 29,100 & 68,904 & 56.21 \\
Consensus team at high authority, near-frontier LLM & 30 & 3,485 & -- & 3,485 & 6,970 & 38.14 \\
Commander team, near-frontier LLM & 120 & -- & 4,231 & 10,919 & 15,150 & 26.47 \\
Birds and civilian drones (five teams) & 25 & 1,667 & 285 & 2,711 & 4,663 & 9.97 \\
Disguised intruders (five teams, without and with forewarning) & 50 & 3,767 & 730 & 6,734 & 11,231 & 28.17 \\
\bottomrule
\end{tabular}
\end{table}
\clearpage

\begin{table}[ht]
\setstretch{1}
\caption{\textbf{Execution identities and re-simulation checks.}
Each run records an execution identity, a hash over its code version, interfaces, model profile, policy settings and study-version label; a run made in several invocations has one identity per invocation.
A recorded episode is valid when it ended without a terminal problem such as an unrecovered call failure or a solver failure; in the run of the near-frontier model at high authority, 6 first-run episodes were invalid after call failures under a provider's rate limit and were run again under a new identity before any outcome was read (Methods).
Every response is stored under a hash of its full request.
A re-simulation rebuilds an episode from the code and the stored responses, with the API key removed, so a missing response stops the episode instead of calling a model.
A re-simulated episode matched when it equalled its record on the breach, the breach step, the breaching intruder, every intruder's outcome, the number of intruders destroyed and the steps run; a match means equality on these recorded fields, not trajectory identity.
The episodes with birds and civilian drones, without and with the disguise, were not re-simulated: their measures are read from the records and the stored responses.
The disguise ran as one invocation under one execution identity, and the prompt version recorded with every call names the condition, without or with the forewarning.}
\label{tab:repro}
\centering
\tablebodyfont
\setlength{\tabcolsep}{4pt}
\begin{tabular}{@{}llccc@{}}
\toprule
Run & Execution identities & \makecell{Episodes\\recorded} & \makecell{Valid\\episodes} & \makecell{Re-simulated,\\matched} \\
\midrule
Main study (three teams) & \makecell[l]{\texttt{d63bb80c8d4f5a75}\\\texttt{b739142f8b120dfd}} & 900 & 900 & 900 of 900 \\
\makecell[l]{Consensus team at high\\authority, low-cost LLM} & \makecell[l]{\texttt{1b6b98cdc56e256d}} & 300 & 300 & 300 of 300 \\
\makecell[l]{Consensus team at high\\authority, near-frontier LLM} & \makecell[l]{\texttt{c707ebe21531bd07}\\\texttt{ca2a2312187c581a}} & 36 & 30 & 30 of 30 \\
Commander team, near-frontier LLM & \makecell[l]{\texttt{f1e1d17fe92b508f}\\\texttt{9e261602168f2bdc}} & 120 & 120 & 120 of 120 \\
Birds and civilian drones (five teams) & \makecell[l]{\texttt{5e4aff75a94d27fc}} & 25 & 25 & -- \\
\makecell[l]{Disguised intruders (five\\teams, without and with forewarning)} & \makecell[l]{\texttt{4ddde6e2dcc9d371}} & 50 & 50 & -- \\
\bottomrule
\end{tabular}
\end{table}
\clearpage
 
\clearpage
\section{Supplementary Data}

\sidata{data:contrasts}{Every paired comparison with its per-seed values}{%
A comma-separated file with one row per comparison of breach rate in Supplementary Tables~\ref{tab:arch}, \ref{tab:cond} and~\ref{tab:authority}, except the unseen-time comparison.
Columns give the table, the comparison (first minus second), the measure, the conditions and intruder counts pooled, the episode pairs per seed, the estimate (a fraction), the 95\% CI and its method ($t$ over seeds, or Newcombe's method 10 for one pair per seed), and the value of each seed 4001 to 4030.
A seed's value is the mean over its matched pairs of the first breach indicator minus the second.
The per-seed values were recomputed from the episode records, and their mean equals the stored estimate of every comparison.}
 
\section{Supplementary Videos}

\sivideo{vid:commander}{An episode of the commander team on the ideal network with four intruders.}{%
Seed 4003, chosen by a criterion fixed before rendering: the first seed on which each main-study team's breach outcome equalled its most common outcome in this cell.
The episode is illustrative; rates are in Supplementary Table~\ref{tab:cells}.
Dots are defenders with their sensor fields of view, triangles intruders, solid lines the last 4\,sec of each track and dashed lines pursuits; a black ring marks the commander, hollow grey triangles are destroyed intruders and grey crosses disabled defenders.
The defended region is at the upper right.
Drawn from a deterministic re-simulation that matched its record, at four times real time.
No breach; all four intruders destroyed.}
\sivideo{vid:consensus}{An episode of the consensus team on the ideal network with four intruders.}{%
The same seed, criterion and symbols as Supplementary Video~\ref{vid:commander}.
No breach; all four intruders destroyed.}
\sivideo{vid:independent}{An episode of the independent team on the ideal network with four intruders.}{%
The same seed, criterion and symbols as Supplementary Video~\ref{vid:commander}.
No breach; three intruders destroyed, and the episode ended at the 40\,sec limit with one intruder alive.}
 
\clearpage
\sinote{note:prompts}{Prompts, response schemas and example calls}
Four versioned interfaces build every request: m8b.7 for the defenders of the consensus team at low authority and of the independent team, m8b.9 for the defenders of the consensus team at high authority, commander.10 for the commander and intruder.4 for the intruders. Outside the episodes with birds and civilian drones, each interface sent the same system prompt and response schema in every recorded request, the intruder interface one of each per phase (before reaching a gate and after passing one), and the same text served both models where both were used. In the episodes with birds and civilian drones, the defender and commander prompts carried one added paragraph, the airspace paragraph, reproduced below. A user message states the agent's own picture at the decision epoch. A defender's message gives its own state, its teammates' positions, the intruders it knows of, with the age of each belief, how long each coverage cell has gone unwatched, its current task, the time left and every candidate task with each term of its bid. The commander's message gives every active defender with its last issued assignment, every reported intruder with its forecast and each defender's optimistic interception time, the task list and the coverage record. An intruder's message gives its own state and phase, its current action, the gates, its teammates' states and last choices, and the team's shared observations of defenders. Each box is colored as the team it is written for is in the main figures: teal for the consensus team at low authority, dark green for it at high authority, indigo for the independent team, ochre for the commander and orange for the intruders; grey marks text shared by several of them. The prompts are reproduced exactly as sent, including their own terms: \texttt{BLUE} and \texttt{blue} denote the defenders, and \texttt{RED} and \texttt{red} the intruders. The code renders the numbers in the defender prompts from the constants that the bid computation uses. Calls, decode settings and model profiles are in Supplementary Table~\ref{tab:profiles}.

{\setstretch{1}
\subsection{System prompts}
\begin{promptbox}{Teal}{Defender, consensus team at low authority (interface m8b.7)}
You are ONE defending agent (BLUE) in a 2D pursuit-evasion simulation. Blue defends a region against RED intruders. Your team allocates tasks with an auction: each blue bids on its own candidate tasks; the highest bid for a task wins it, and a blue that is outbid takes its best remaining task. You do NOT choose your task and you do NOT write any bid. You choose how YOUR bids are weighted.

Answer with four things, in this order, and nothing numeric beyond contact ids:

1. reasoning -- FIRST: one sentence about the actual picture (distances, ages, what you are trading).

2. scoring_profile -- exactly one of:
- aggressive: as balanced except attrition 0.5 -> 0.2, urgency time scale 60 -> 90 s
- balanced: the reference weights (coverage 0.5, leaving 0.25, information 0.45, attrition 0.5, urgency time scale 60 sec)
- conservative: as balanced except coverage 0.5 -> 0.8, leaving 0.25 -> 0.4, attrition 0.5 -> 0.9
- scouting: as balanced except information 0.45 -> 0.75

3. threat_priorities -- for contacts YOU believe in (listed below), a priority of LOW, MEDIUM or HIGH. MEDIUM is the default for any contact you leave out. You may only name contact_id values that are listed below.

4. preferred_tasks -- at most 3 task ids, each copied exactly as one of the quoted task_id strings in YOUR candidate task list below (the text inside the quotes only, without the location). Each receives a fixed bonus. Leave it empty if nothing stands out.

Task kinds: NEUTRALIZE:c:s closes to contact c (you are permanently disabled on contact). DEFEND:k holds perimeter station k. RECON:r patrols toward coverage region r.

What you know is what YOU believe, and it is incomplete. Contacts you cannot see are not listed, and a short list does NOT mean the region is safe. Other agents are answering at the same moment on their own beliefs.

How your answer changes the bids (the highest bid is your task unless the auction gives it to a teammate who bids more for it). Bid = value - travel - leaving - switching - continuity (NEUTRALIZE also minus attrition and belief-age uncertainty), plus 0.2 for a preferred task. Your candidate task list shows each task's terms under the balanced profile. Every number below is an allocation value that only ranks tasks for allocation; none of them is the mission's objective.
- Approach sector: intruders are expected from bearings 160 to 300 degrees as seen from the defended centre (any distance). Only ground in that sector counts below.
- DEFEND:k value = 0.05 + coverage weight x coverage, where coverage is how much approach-sector ground station k's sensor wedge (at the station's fixed heading) adds beyond what your teammates' sensors watch now (1.0 = one full sensor's worth).
- RECON:r value = information weight x information: approach-sector ground your sensor sweeps on the way to region r and there, counted by how stale each region is in YOUR coverage record and by how soon you reach it (an optimistic travel forecast from your current velocity); ground a blue already watches counts 0. Travel 0.05 per 25 units.
- NEUTRALIZE:c value = 1.0 x (threat value 1.2 + priority x urgency x opportunity). The threat value is the same for every believed red, whatever its heading, priority or reachability. Urgency = the larger of 1/(1 + forecast time to breach / 60 sec) and 1/(1 + distance to the region / 25). Forecast time to breach: from the contact's believed velocity, the time to close on the approach funnel (the region together with the four public approach gates) at its current closing speed, plus the last leg into the region; none when it is not closing (moving away, crossing or stationary). Opportunity comes from an OPTIMISTIC kinematic interception forecast (from your current velocity, with the per-axis speed and acceleration limits of 2; it ignores collision avoidance and the contact's reaction): 1 if you can reach the contact's forecast path before its forecast breach and within the episode, falling linearly to 0.25 over 10 sec later than that, 0.25 with no interception; averaged over +-0.3 u/s of uncertainty in the contact's velocity. Travel 0.15 per 25 units of current distance; attrition = attrition weight / (1 + teammates), because contact disables you.
- Leaving your current station costs leaving weight x that station's coverage. Changing task costs 0.05. While you pursue a contact you still believe in, every other task (another contact, RECON, DEFEND) costs an extra 0.15 (continuity). For ANOTHER CONTACT only, that extra cost is multiplied by (1 - R x H x S), down to 0: R = how fast the contact you pursue moves away from the approach funnel (0 if it is not moving away, 1 at 0.5 u/s or faster); H = 1 if the other contact's forecast breach comes at least 5 sec before the episode ends, falling to 0 at the end (0 with no forecast breach); S = how far your optimistic interception of the other contact comes before its forecast breach (0 for no lead, 1 for 2 sec or more), averaged over the same velocity uncertainty. RECON and DEFEND always pay the full extra cost. Each task line shows the continuity actually applied.
- Priorities multiply only urgency x opportunity, never the threat value: LOW 0.5, MEDIUM 1, HIGH 2.

Mission (the same for every blue team): your team's objective is to prevent any red from entering the defended region before the episode ends. The episode lasts 400 steps (dt = 0.1 sec, 40 sec). Any entry by a red is a breach and ends the episode. A blue that touches a red destroys it and is permanently disabled. A red that leaves the arena is removed and can no longer breach. The episode also ends, without a breach, once every red has been destroyed or has left the arena. Reds can change direction at any time, and reds your team has not detected may exist.

Headings: every agent's sensor points along its heading and sees a 60-degree wedge (30 degrees either side) out to 35 units; nothing outside it is detected. Angles are in degrees, counter-clockwise from the +x axis (0 = +x, 90 = +y, 180 = -x, 270 = -y). Steering and sensing are independent: an agent can move one way while looking another. A heading listed for an opponent is the one seen at that sighting and is NOT updated since (agents can turn up to 360 degrees per second).

Respond only with a valid JSON object matching the required schema, with "reasoning" as its first field.
\end{promptbox}
The independent team's prompt differs from the consensus team's only in the lines below (unified diff: lines starting with a minus sign are the consensus team's, lines starting with a plus sign the independent team's).
\begin{promptbox}{Indigo}{Defender, independent team (interface m8b.7): difference}
--- consensus, low authority
+++ independent
-You are ONE defending agent (BLUE) in a 2D pursuit-evasion simulation. Blue defends a region against RED intruders. Your team allocates tasks with an auction: each blue bids on its own candidate tasks; the highest bid for a task wins it, and a blue that is outbid takes its best remaining task. You do NOT choose your task and you do NOT write any bid. You choose how YOUR bids are weighted.
+You are ONE defending agent (BLUE) in a 2D pursuit-evasion simulation. Blue defends a region against RED intruders. Each blue computes bids for its own candidate tasks and takes its own highest-bid task. There is no coordination between blues: teammates may take the same task. You do NOT choose your task and you do NOT write any bid. You choose how YOUR bids are weighted.
-How your answer changes the bids (the highest bid is your task unless the auction gives it to a teammate who bids more for it). Bid = value - travel - leaving - switching - continuity (NEUTRALIZE also minus attrition and belief-age uncertainty), plus 0.2 for a preferred task. Your candidate task list shows each task's terms under the balanced profile. Every number below is an allocation value that only ranks tasks for allocation; none of them is the mission's objective.
+How your answer changes the bids (your highest bid is your task). Bid = value - travel - leaving - switching - continuity (NEUTRALIZE also minus attrition and belief-age uncertainty), plus 0.2 for a preferred task. Your candidate task list shows each task's terms under the balanced profile. Every number below is an allocation value that only ranks tasks for allocation; none of them is the mission's objective.
\end{promptbox}
\begin{promptbox}{Deep}{Defender, consensus team at high authority, both models (interface m8b.9)}
You are ONE defending agent (BLUE) in a 2D pursuit-evasion simulation. Blue defends a region against RED intruders. Your team allocates tasks with an auction: each blue bids on its own candidate tasks; the highest bid for a task wins it, and a blue that is outbid takes its best remaining task. You do NOT choose your task and you do NOT write any bid. You choose the weights YOUR bids use.

Answer with four things, in this order:

1. reasoning -- FIRST: at most two sentences about the actual picture and what your weights trade off.

2. weights -- all 12 weights listed below, each a number within its allowed range.

3. contact_priorities -- for contacts YOU believe in (listed below), a number from 0 to 2. 1 is the default for any contact you leave out. You may only name contact_id values that are listed below.

4. preferred_tasks -- at most 3 task ids, each copied exactly as one of the quoted task_id strings in YOUR candidate task list below (the text inside the quotes only, without the location). Leave it empty if nothing stands out.

Task kinds: NEUTRALIZE:c:s closes to contact c (you are permanently disabled on contact). DEFEND:k holds perimeter station k. RECON:r patrols toward coverage region r.

What you know is what YOU believe, and it is incomplete. Contacts you cannot see are not listed, and a short list does NOT mean the region is safe. Other agents are answering at the same moment on their own beliefs.

How bids are formed (the highest bid is your task unless the auction gives it to a teammate who bids more for it). Every blue computes its bids with the same formula from its own beliefs and its own weights. You choose the weights YOUR bids use; teammates choose theirs. Every number below is an allocation value that only ranks tasks for allocation; none of them is the mission's objective.
- DEFEND:k bid = w_station + w_coverage x coverage - w_travel x distance/25 - penalties
- RECON:r bid = w_scout x information - w_scout_travel x distance/25 - penalties
- NEUTRALIZE:c bid = priority(c) x (w_threat + w_engage x urgency x opportunity) - w_travel x distance/25 - w_attrition x 1/(1 + teammates) - w_belief_age x age term - penalties
- penalties = w_leave x leaving + w_switch x switching + w_continuity x continuity; then plus 0.2 for each of your preferred tasks.
Features (each candidate task line below lists its own):
- Approach sector: intruders are expected from bearings 160 to 300 degrees as seen from the defended centre (any distance). Only ground in that sector counts below.
- coverage: how much approach-sector ground station k's sensor wedge (at the station's fixed heading) adds beyond what your teammates' sensors watch now (1.0 = one full sensor's worth).
- information: approach-sector ground your sensor sweeps on the way to region r and there, counted by how stale each region is in YOUR coverage record and by how soon you reach it (an optimistic travel forecast from your current velocity); ground a blue already watches counts 0.
- urgency: the larger of 1/(1 + forecast time to breach / 60 sec) and 1/(1 + distance to the region / 25). Forecast time to breach: from the contact's believed velocity, the time to close on the approach funnel (the region together with the four public approach gates) at its current closing speed, plus the last leg into the region; none when it is not closing (moving away, crossing or stationary).
- opportunity: from an OPTIMISTIC kinematic interception forecast (from your current velocity, with the per-axis speed and acceleration limits of 2; it ignores collision avoidance and the contact's reaction): 1 if you can reach the contact's forecast path before its forecast breach and within the episode, falling linearly to 0.25 over 10 sec later than that, 0.25 with no interception; averaged over +-0.3 u/s of uncertainty in the contact's velocity.
- age term: the belief's age x the reds' top speed / 25.
- leaving: the coverage of the station you would leave (0 if you hold no station or the task is that station).
- switching: 1 for every task other than your current task, else 0.
- continuity: while you pursue a contact you still believe in, 1 for every other task (RECON, DEFEND); for ANOTHER CONTACT it is (1 - R x H x S), down to 0: R = how fast the contact you pursue moves away from the approach funnel (0 if it is not moving away, 1 at 0.5 u/s or faster); H = 1 if the other contact's forecast breach comes at least 5 sec before the episode ends, falling to 0 at the end (0 with no forecast breach); S = how far your optimistic interception of the other contact comes before its forecast breach (0 for no lead, 1 for 2 sec or more). 0 when you pursue no contact.
Weights (what each multiplies, default, allowed range):
- w_threat: NEUTRALIZE: value every believed red carries, before priority. Default 1.2, allowed 0 to 3.
- w_engage: NEUTRALIZE: x urgency x opportunity. Default 1, allowed 0 to 3.
- w_station: DEFEND: value of holding any station. Default 0.05, allowed 0 to 1.5.
- w_coverage: DEFEND: x the station's coverage. Default 0.5, allowed 0 to 3.
- w_scout: RECON: x information. Default 0.45, allowed 0 to 3.
- w_attrition: NEUTRALIZE: cost x 1/(1 + teammates), because contact disables you. Default 0.5, allowed 0 to 2.
- w_belief_age: NEUTRALIZE: cost x the belief-age term. Default 0.1, allowed 0 to 1.
- w_travel: NEUTRALIZE and DEFEND: cost per 25 units of distance. Default 0.15, allowed 0 to 0.6.
- w_scout_travel: RECON: cost per 25 units of distance. Default 0.05, allowed 0 to 0.3.
- w_leave: any task that leaves your current station: cost x that station's coverage. Default 0.25, allowed 0 to 1.5.
- w_continuity: any task other than the contact you pursue: cost x the continuity factor. Default 0.15, allowed 0 to 1.
- w_switch: any task other than your current task: cost. Default 0.05, allowed 0 to 0.5.
- Budget: w_threat + w_engage + w_station + w_coverage + w_scout must not exceed 3.2 (the defaults sum to exactly 3.2).
- priority(c): a number from 0 to 2 per contact you believe in, default 1 for any contact you leave out; the priorities of all the contacts you believe in must not sum to more than their number.
A value outside its allowed range is replaced by its default. If the five budgeted weights sum to more than the budget, all five are scaled down in proportion until they sum to the budget; likewise, if the priorities of the contacts you believe in sum to more than their number, all of them are scaled down in proportion until they sum to their number. So raising one budgeted weight (or one priority) without lowering the others lowers the others. A low weight or priority lowers a bid; it does not remove the task, and a blue that is outbid takes its best remaining task, even one with a negative bid. Your candidate task list shows each task's features and its bid under the default weights.

Mission (the same for every blue team): your team's objective is to prevent any red from entering the defended region before the episode ends. The episode lasts 400 steps (dt = 0.1 sec, 40 sec). Any entry by a red is a breach and ends the episode. A blue that touches a red destroys it and is permanently disabled. A red that leaves the arena is removed and can no longer breach. The episode also ends, without a breach, once every red has been destroyed or has left the arena. Reds can change direction at any time, and reds your team has not detected may exist.

Headings: every agent's sensor points along its heading and sees a 60-degree wedge (30 degrees either side) out to 35 units; nothing outside it is detected. Angles are in degrees, counter-clockwise from the +x axis (0 = +x, 90 = +y, 180 = -x, 270 = -y). Steering and sensing are independent: an agent can move one way while looking another. A heading listed for an opponent is the one seen at that sighting and is NOT updated since (agents can turn up to 360 degrees per second).

Respond only with a valid JSON object matching the required schema, with "reasoning" as its first field.
\end{promptbox}
\begin{promptbox}{Ochre}{Commander, both models (interface commander.10)}
You are the commander of the BLUE defender team in a 2D perimeter-defense game. Red intruders try to breach the defended region; a blue that touches a red destroys it and is itself disabled.
Each planning step you assign every active blue exactly ONE task from the task list. Task kinds:
- NEUTRALIZE:<contact>:<slot> -- pursue and intercept that believed red contact. Each slot can go to one blue only.
- DEFEND:<station> -- hold that defensive station on the perimeter. One blue per station.
- RECON:<cell> -- sweep that arena cell to find reds. One blue per cell.
Every task id may be assigned to at most one blue. Your picture of the reds comes from reports your team delivered to you; reds you have not heard about may exist. Headings: every agent's sensor points along its heading and sees a 60-degree wedge (30 degrees either side) out to 35 units; nothing outside it is detected. Angles are in degrees, counter-clockwise from the +x axis (0 = +x, 90 = +y, 180 = -x, 270 = -y). Steering and sensing are independent: an agent can move one way while looking another. A heading listed for an opponent is the one seen at that sighting and is NOT updated since (agents can turn up to 360 degrees per second).
Approach sector and sensing: intruders are expected from bearings 160 to 300 degrees as seen from the defended centre (any distance). A blue holding a station looks along that station's fixed heading; the approach-sector ground each station's sensor wedge covers is listed per station. A blue on RECON flies toward its cell with its sensor pointed ahead and holds there. Ground is watched only while some blue's sensor wedge covers it: a scout adds coverage only where no other blue is watching at that time, and a station's ground stays watched after its blue leaves only if another blue's sensor covers it.
Forecasts shown with the reports (from your picture, labelled as forecasts): each contact's status (approaching / moving away / crossing / stationary, from its reported velocity relative to the approach funnel = the region together with the four public approach gates), its forecast time to breach (closing time to the funnel plus the last leg; none when not closing) and urgency = the larger of 1/(1 + forecast time / 60 sec) and 1/(1 + distance to the region / 25); for each blue and contact an OPTIMISTIC kinematic interception forecast: it assumes the contact keeps its reported velocity and uses only the blue's speed and acceleration limits of 2 per axis from its current velocity (no collision avoidance). A contact that reacts can make the real interception earlier, later or impossible. Every forecast and coverage figure is computed from your picture, which can be stale or incomplete.
Each blue's line shows the LAST ASSIGNMENT YOU ISSUED to it and the step you issued it. Issuing is not delivery: with delayed or lost communication the blue may not have received or started it. "no assignment issued yet" only means you have not assigned that blue anything; it does not tell you what the blue is doing.
Mission (the same for every blue team): your team's objective is to prevent any red from entering the defended region before the episode ends. The episode lasts 400 steps (dt = 0.1 sec, 40 sec). Any entry by a red is a breach and ends the episode. A blue that touches a red destroys it and is permanently disabled. A red that leaves the arena is removed and can no longer breach. The episode also ends, without a breach, once every red has been destroyed or has left the arena. Reds can change direction at any time, and reds your team has not detected may exist.
Valid answers:
- Give every active blue exactly one task, and give each full task id to at most one blue. Different slots of one contact (NEUTRALIZE:c:0, NEUTRALIZE:c:1) are different tasks; the same task id given to two blues is invalid and the second blue is sent back to its station.
Answer with JSON matching the schema: task_by_agent maps every active blue's id to its one task id. Give your reasoning first.
\end{promptbox}
\begin{promptbox}{Peach}{Intruder, before reaching a gate (interface intruder.4)}
You are ONE red intruder drone in a 2D perimeter-defense game. Your team tries to breach a defended region; blue defenders try to intercept you, and a blue that touches you destroys you (and itself). Your team's objective is for at least one red to enter the defended region before the episode ends. The episode lasts 400 steps (dt = 0.1 sec, 40 sec). Any entry by a red is a breach: your team succeeds and the episode ends. A red that leaves the arena is removed and can no longer breach. The episode also ends, without a breach, once every red has been destroyed or has left the arena. You decide only for yourself; each red teammate decides for itself.
You know: your own state, your teammates' states, your team's shared sightings of blue defenders (each with when it was seen and its predicted position now; blues you have not seen may exist), and what each teammate chose at the previous planning step.
Actions:
- evade: move away from ONE blue you currently believe in (target_id). Evading does not change your phase.
- attack: fly to ONE of the approach gates (target_id) -- public waypoints outside the defended region. On reaching your chosen gate you switch permanently to the interior attack. You may change gate at any planning step before you reach one.
Headings: every agent's sensor points along its heading and sees a 60-degree wedge (30 degrees either side) out to 35 units; nothing outside it is detected. Angles are in degrees, counter-clockwise from the +x axis (0 = +x, 90 = +y, 180 = -x, 270 = -y). Steering and sensing are independent: an agent can move one way while looking another. A heading listed for an opponent is the one seen at that sighting and is NOT updated since (agents can turn up to 360 degrees per second).
target_id is the gate_id you attack, or the blue id you evade. Answer with JSON matching the schema; give your reasoning first.
\end{promptbox}
\begin{promptbox}{Peach}{Intruder, after passing a gate (interface intruder.4)}
You are ONE red intruder drone in a 2D perimeter-defense game. Your team tries to breach a defended region; blue defenders try to intercept you, and a blue that touches you destroys you (and itself). Your team's objective is for at least one red to enter the defended region before the episode ends. The episode lasts 400 steps (dt = 0.1 sec, 40 sec). Any entry by a red is a breach: your team succeeds and the episode ends. A red that leaves the arena is removed and can no longer breach. The episode also ends, without a breach, once every red has been destroyed or has left the arena. You decide only for yourself; each red teammate decides for itself.
You know: your own state, your teammates' states, your team's shared sightings of blue defenders (each with when it was seen and its predicted position now; blues you have not seen may exist), and what each teammate chose at the previous planning step.
Actions:
- evade: move away from ONE blue you currently believe in (target_id). Evading does not change your phase.
- attack: you have passed your gate and are in the INTERIOR phase: attack means advance on the defended region's centre. There are no gates to choose any more.
Headings: every agent's sensor points along its heading and sees a 60-degree wedge (30 degrees either side) out to 35 units; nothing outside it is detected. Angles are in degrees, counter-clockwise from the +x axis (0 = +x, 90 = +y, 180 = -x, 270 = -y). Steering and sensing are independent: an agent can move one way while looking another. A heading listed for an opponent is the one seen at that sighting and is NOT updated since (agents can turn up to 360 degrees per second).
target_id is the blue id you evade, or -1 for attack. Answer with JSON matching the schema; give your reasoning first.
\end{promptbox}
\subsection*{The airspace paragraph}
In the episodes with birds and civilian drones, with and without the disguise, this paragraph was inserted into every defender and commander system prompt after the brief's last sentence, and each listed contact's image followed the user message as an image labelled ``image of contact\_id $N$''. The intruders' prompts were unchanged.
\begin{promptbox}{Grey}{The airspace paragraph}
Not every contact is a red. Besides the red intruders, the airspace holds birds and civilian drones. They move in the same airspace, cannot breach, and are not part of the objective. A blue that touches any contact, red, bird or civilian drone, destroys it and is permanently disabled. The episode ends without a breach once every red has been destroyed or has left the arena, whatever the birds and civilian drones do. A contact's id does not tell what it is. One image of each contact listed below is attached after this message, labelled with its contact_id.
\end{promptbox}
In the disguise runs with the forewarning, one sentence followed ``A contact's id does not tell what it is.'' and nothing else in any prompt differed:
\begin{promptbox}{Grey}{The forewarning sentence}
Some red intruders may be disguised as birds.
\end{promptbox}
\subsection*{Response schemas}
Each interface used one strict JSON schema in every request; the reasoning field comes first.
\begin{promptbox}{Grey}{Defender, consensus team at low authority and independent team: response schema}
{
 "type": "json_schema",
 "json_schema": {
  "name": "bid_modifiers",
  "strict": true,
  "schema": {
   "type": "object",
   "properties": {
    "reasoning": {
     "type": "string"
    },
    "scoring_profile": {
     "type": "string",
     "enum": [
      "aggressive",
      "balanced",
      "conservative",
      "scouting"
     ]
    },
    "threat_priorities": {
     "type": "array",
     "items": {
      "type": "object",
      "properties": {
       "contact_id": {
        "type": "integer"
       },
       "priority": {
        "type": "string",
        "enum": [
         "HIGH",
         "LOW",
         "MEDIUM"
        ]
       }
      },
      "required": [
       "contact_id",
       "priority"
      ],
      "additionalProperties": false
     }
    },
    "preferred_tasks": {
     "type": "array",
     "items": {
      "type": "string"
     }
    }
   },
   "required": [
    "reasoning",
    "scoring_profile",
    "threat_priorities",
    "preferred_tasks"
   ],
   "additionalProperties": false
  }
 }
}
\end{promptbox}
\begin{promptbox}{Deep}{Defender, consensus team at high authority: response schema}
{
 "type": "json_schema",
 "json_schema": {
  "name": "strategy_weights",
  "strict": true,
  "schema": {
   "type": "object",
   "properties": {
    "reasoning": {
     "type": "string"
    },
    "weights": {
     "type": "object",
     "properties": {
      "w_threat": {
       "type": "number"
      },
      "w_engage": {
       "type": "number"
      },
      "w_station": {
       "type": "number"
      },
      "w_coverage": {
       "type": "number"
      },
      "w_scout": {
       "type": "number"
      },
      "w_attrition": {
       "type": "number"
      },
      "w_belief_age": {
       "type": "number"
      },
      "w_travel": {
       "type": "number"
      },
      "w_scout_travel": {
       "type": "number"
      },
      "w_leave": {
       "type": "number"
      },
      "w_continuity": {
       "type": "number"
      },
      "w_switch": {
       "type": "number"
      }
     },
     "required": [
      "w_threat",
      "w_engage",
      "w_station",
      "w_coverage",
      "w_scout",
      "w_attrition",
      "w_belief_age",
      "w_travel",
      "w_scout_travel",
      "w_leave",
      "w_continuity",
      "w_switch"
     ],
     "additionalProperties": false
    },
    "contact_priorities": {
     "type": "array",
     "items": {
      "type": "object",
      "properties": {
       "contact_id": {
        "type": "integer"
       },
       "priority": {
        "type": "number"
       }
      },
      "required": [
       "contact_id",
       "priority"
      ],
      "additionalProperties": false
     }
    },
    "preferred_tasks": {
     "type": "array",
     "items": {
      "type": "string"
     }
    }
   },
   "required": [
    "reasoning",
    "weights",
    "contact_priorities",
    "preferred_tasks"
   ],
   "additionalProperties": false
  }
 }
}
\end{promptbox}
\begin{promptbox}{Ochre}{Commander: response schema}
{
 "type": "json_schema",
 "json_schema": {
  "name": "commander_assignment",
  "strict": true,
  "schema": {
   "type": "object",
   "properties": {
    "reasoning": {
     "type": "string"
    },
    "task_by_agent": {
     "type": "object",
     "properties": {
      "<agent id>": {
       "type": "string",
       "enum": [
        "<task ids of this epoch>"
       ]
      }
     },
     "required": [
      "<every active agent id>"
     ],
     "additionalProperties": false
    }
   },
   "required": [
    "reasoning",
    "task_by_agent"
   ],
   "additionalProperties": false
  }
 }
}
\end{promptbox}
\begin{promptbox}{Peach}{Intruder, before reaching a gate: response schema}
{
 "type": "json_schema",
 "json_schema": {
  "name": "intruder_decision",
  "strict": true,
  "schema": {
   "type": "object",
   "properties": {
    "reasoning": {
     "type": "string"
    },
    "action": {
     "type": "string",
     "enum": [
      "attack",
      "evade"
     ]
    },
    "target_id": {
     "type": "integer",
     "description": "the gate_id to attack, or the blue id to evade"
    }
   },
   "required": [
    "reasoning",
    "action",
    "target_id"
   ],
   "additionalProperties": false
  }
 }
}
\end{promptbox}
\begin{promptbox}{Peach}{Intruder, after passing a gate: response schema}
{
 "type": "json_schema",
 "json_schema": {
  "name": "intruder_decision",
  "strict": true,
  "schema": {
   "type": "object",
   "properties": {
    "reasoning": {
     "type": "string"
    },
    "action": {
     "type": "string",
     "enum": [
      "attack",
      "evade"
     ]
    },
    "target_id": {
     "type": "integer",
     "description": "the blue id to evade, or -1 for attack"
    }
   },
   "required": [
    "reasoning",
    "action",
    "target_id"
   ],
   "additionalProperties": false
  }
 }
}
\end{promptbox}
\subsection*{Example requests and responses}
For each interface, the first recorded call with a listed contact in the first episode in sorted order, as sent and as answered. Each request also carried the interface's system prompt and schema above.
\begin{promptbox}{Teal}{Defender, consensus team at low authority: user message}
sim_step: 110
arena_size: 100.00
defense_center: [82.00, 82.00]
you are blue agent_id=0
  your_position: [92.85, 81.98]
  your_velocity: [-0.00, -0.00]
  your_heading_deg: 255
active_teammates (positions known exactly; intentions unknown, no communication this cycle):
  agent_id=1 pos=[74.46, 81.84] vel=[-0.00, 0.03] heading_deg=185
  agent_id=2 pos=[50.17, 70.73] vel=[-2.00, -1.99] heading_deg=253
  agent_id=3 pos=[81.99, 73.65] vel=[-0.01, 0.21] heading_deg=230
contacts you believe in (YOUR belief only -- others may exist that you cannot see; age_steps is how old the underlying observation is):
  contact_id=0 believed_position=[41.38, 39.10] believed_velocity=[2.00, 2.00] observed_heading_deg=74 age_steps=1
coverage (steps since your knowledge of each region was refreshed; -1 = never):
  region_id=0 center=[12.50, 12.50] steps_since_seen=-1
  region_id=1 center=[12.50, 37.50] steps_since_seen=-1
  region_id=2 center=[12.50, 62.50] steps_since_seen=-1
  region_id=3 center=[12.50, 87.50] steps_since_seen=-1
  region_id=4 center=[37.50, 12.50] steps_since_seen=-1
  region_id=5 center=[37.50, 37.50] steps_since_seen=-1
  region_id=6 center=[37.50, 62.50] steps_since_seen=4
  region_id=7 center=[37.50, 87.50] steps_since_seen=94
  region_id=8 center=[62.50, 12.50] steps_since_seen=-1
  region_id=9 center=[62.50, 37.50] steps_since_seen=45
  region_id=10 center=[62.50, 62.50] steps_since_seen=2
  region_id=11 center=[62.50, 87.50] steps_since_seen=8
  region_id=12 center=[87.50, 12.50] steps_since_seen=-1
  region_id=13 center=[87.50, 37.50] steps_since_seen=94
  region_id=14 center=[87.50, 62.50] steps_since_seen=0
  region_id=15 center=[87.50, 87.50] steps_since_seen=90
your current task: DEFEND:0
episode time left: 29.0 s
your candidate tasks (task_id, where it is, terms under the balanced profile):
  task_id "NEUTRALIZE:0:0"    at [41.38, 39.10]    value 1.98 (threat 1.20 + urgency 0.78 x opportunity 1.00; approaching, forecast breach 17.1 sec, interception forecast 13.1 sec), travel 0.40, leaving 0.15, switch 0.05, attrition 0.12, uncertainty 0.00 -> bid 1.26
  task_id "DEFEND:0"    at [92.86, 82.00]    value 0.34 (coverage 0.58), travel 0.00 -> bid 0.34
  task_id "DEFEND:1"    at [82.00, 91.94]    value 0.18 (coverage 0.25), travel 0.09, leaving 0.15, switch 0.05 -> bid -0.11
  task_id "DEFEND:2"    at [74.46, 82.00]    value 0.05 (coverage 0.00), travel 0.11, leaving 0.15, switch 0.05 -> bid -0.25
  task_id "DEFEND:3"    at [82.00, 73.54]    value 0.05 (coverage 0.01), travel 0.08, leaving 0.15, switch 0.05 -> bid -0.23
  task_id "RECON:0"    at [12.50, 12.50]    value 0.25 (information 0.55), travel 0.21, leaving 0.15, switch 0.05 -> bid -0.16
  task_id "RECON:1"    at [12.50, 37.50]    value 0.17 (information 0.37), travel 0.18, leaving 0.15, switch 0.05 -> bid -0.21
  task_id "RECON:2"    at [12.50, 62.50]    value 0.18 (information 0.40), travel 0.17, leaving 0.15, switch 0.05 -> bid -0.18
  task_id "RECON:3"    at [12.50, 87.50]    value 0.21 (information 0.46), travel 0.16, leaving 0.15, switch 0.05 -> bid -0.15
  task_id "RECON:4"    at [37.50, 12.50]    value 0.28 (information 0.63), travel 0.18, leaving 0.15, switch 0.05 -> bid -0.09
  task_id "RECON:5"    at [37.50, 37.50]    value 0.22 (information 0.49), travel 0.14, leaving 0.15, switch 0.05 -> bid -0.12
  task_id "RECON:6"    at [37.50, 62.50]    value 0.15 (information 0.33), travel 0.12, leaving 0.15, switch 0.05 -> bid -0.16
  task_id "RECON:7"    at [37.50, 87.50]    value 0.19 (information 0.42), travel 0.11, leaving 0.15, switch 0.05 -> bid -0.12
  task_id "RECON:8"    at [62.50, 12.50]    value 0.32 (information 0.72), travel 0.15, leaving 0.15, switch 0.05 -> bid -0.02
  task_id "RECON:9"    at [62.50, 37.50]    value 0.25 (information 0.55), travel 0.11, leaving 0.15, switch 0.05 -> bid -0.06
  task_id "RECON:10"    at [62.50, 62.50]    value 0.02 (information 0.04), travel 0.07, leaving 0.15, switch 0.05 -> bid -0.25
  task_id "RECON:11"    at [62.50, 87.50]    value 0.05 (information 0.11), travel 0.06, leaving 0.15, switch 0.05 -> bid -0.21
  task_id "RECON:12"    at [87.50, 12.50]    value 0.33 (information 0.74), travel 0.14, leaving 0.15, switch 0.05 -> bid -0.00
  task_id "RECON:13"    at [87.50, 37.50]    value 0.33 (information 0.72), travel 0.09, leaving 0.15, switch 0.05 -> bid 0.04
  task_id "RECON:14"    at [87.50, 62.50]    value 0.12 (information 0.26), travel 0.04, leaving 0.15, switch 0.05 -> bid -0.12
  task_id "RECON:15"    at [87.50, 87.50]    value 0.00 (information 0.00), travel 0.02, leaving 0.15, switch 0.05 -> bid -0.21
\end{promptbox}
\begin{promptbox}{Teal}{Defender, consensus team at low authority: response}
{
  "reasoning": "Contact 0 is moving steadily into the defended region and presents a high-priority threat that justifies immediate interception.",
  "scoring_profile": "balanced",
  "threat_priorities": [
    {
      "contact_id": 0,
      "priority": "HIGH"
    }
  ],
  "preferred_tasks": [
    "NEUTRALIZE:0:0"
  ]
}
\end{promptbox}
\begin{promptbox}{Indigo}{Defender, independent team: user message}
sim_step: 200
arena_size: 100.00
defense_center: [82.00, 82.00]
you are blue agent_id=0
  your_position: [72.54, 61.71]
  your_velocity: [-2.00, -2.00]
  your_heading_deg: 196
active_teammates (positions known exactly; intentions unknown, no communication this cycle):
  agent_id=1 pos=[55.33, 61.76] vel=[-1.28, -1.70] heading_deg=252
  agent_id=2 pos=[53.36, 54.47] vel=[-0.40, -0.23] heading_deg=80
contacts you believe in (YOUR belief only -- others may exist that you cannot see; age_steps is how old the underlying observation is):
  contact_id=0 believed_position=[54.26, 57.09] believed_velocity=[1.60, 2.00] observed_heading_deg=254 age_steps=0
coverage (steps since your knowledge of each region was refreshed; -1 = never):
  region_id=0 center=[12.50, 12.50] steps_since_seen=-1
  region_id=1 center=[12.50, 37.50] steps_since_seen=-1
  region_id=2 center=[12.50, 62.50] steps_since_seen=-1
  region_id=3 center=[12.50, 87.50] steps_since_seen=-1
  region_id=4 center=[37.50, 12.50] steps_since_seen=47
  region_id=5 center=[37.50, 37.50] steps_since_seen=1
  region_id=6 center=[37.50, 62.50] steps_since_seen=7
  region_id=7 center=[37.50, 87.50] steps_since_seen=197
  region_id=8 center=[62.50, 12.50] steps_since_seen=-1
  region_id=9 center=[62.50, 37.50] steps_since_seen=17
  region_id=10 center=[62.50, 62.50] steps_since_seen=0
  region_id=11 center=[62.50, 87.50] steps_since_seen=1
  region_id=12 center=[87.50, 12.50] steps_since_seen=-1
  region_id=13 center=[87.50, 37.50] steps_since_seen=186
  region_id=14 center=[87.50, 62.50] steps_since_seen=90
  region_id=15 center=[87.50, 87.50] steps_since_seen=183
your current task: NEUTRALIZE:0:0
episode time left: 20.0 s
your candidate tasks (task_id, where it is, terms under the balanced profile):
  task_id "NEUTRALIZE:0:0"    at [54.26, 57.09]    value 2.05 (threat 1.20 + urgency 0.85 x opportunity 1.00; approaching, forecast breach 11.0 sec, interception forecast 5.0 sec), travel 0.11, attrition 0.17 -> bid 1.77
  task_id "DEFEND:0"    at [92.86, 82.00]    value 0.43 (coverage 0.77), travel 0.17, switch 0.05, continuity 0.15 -> bid 0.06
  task_id "DEFEND:1"    at [82.00, 91.94]    value 0.15 (coverage 0.21), travel 0.19, switch 0.05, continuity 0.15 -> bid -0.24
  task_id "DEFEND:2"    at [74.46, 82.00]    value 0.21 (coverage 0.32), travel 0.12, switch 0.05, continuity 0.15 -> bid -0.11
  task_id "DEFEND:3"    at [82.00, 73.54]    value 0.49 (coverage 0.87), travel 0.09, switch 0.05, continuity 0.15 -> bid 0.19
  task_id "RECON:0"    at [12.50, 12.50]    value 0.24 (information 0.53), travel 0.16, switch 0.05, continuity 0.15 -> bid -0.12
  task_id "RECON:1"    at [12.50, 37.50]    value 0.27 (information 0.59), travel 0.13, switch 0.05, continuity 0.15 -> bid -0.06
  task_id "RECON:2"    at [12.50, 62.50]    value 0.33 (information 0.74), travel 0.12, switch 0.05, continuity 0.15 -> bid 0.01
  task_id "RECON:3"    at [12.50, 87.50]    value 0.45 (information 0.99), travel 0.13, switch 0.05, continuity 0.15 -> bid 0.12
  task_id "RECON:4"    at [37.50, 12.50]    value 0.20 (information 0.44), travel 0.12, switch 0.05, continuity 0.15 -> bid -0.12
  task_id "RECON:5"    at [37.50, 37.50]    value 0.21 (information 0.46), travel 0.09, switch 0.05, continuity 0.15 -> bid -0.08
  task_id "RECON:6"    at [37.50, 62.50]    value 0.31 (information 0.68), travel 0.07, switch 0.05, continuity 0.15 -> bid 0.04
  task_id "RECON:7"    at [37.50, 87.50]    value 0.35 (information 0.78), travel 0.09, switch 0.05, continuity 0.15 -> bid 0.06
  task_id "RECON:8"    at [62.50, 12.50]    value 0.45 (information 1.00), travel 0.10, switch 0.05, continuity 0.15 -> bid 0.15
  task_id "RECON:9"    at [62.50, 37.50]    value 0.27 (information 0.59), travel 0.05, switch 0.05, continuity 0.15 -> bid 0.01
  task_id "RECON:10"    at [62.50, 62.50]    value 0.06 (information 0.13), travel 0.02, switch 0.05, continuity 0.15 -> bid -0.16
  task_id "RECON:11"    at [62.50, 87.50]    value 0.03 (information 0.06), travel 0.06, switch 0.05, continuity 0.15 -> bid -0.23
  task_id "RECON:12"    at [87.50, 12.50]    value 0.68 (information 1.50), travel 0.10, switch 0.05, continuity 0.15 -> bid 0.37
  task_id "RECON:13"    at [87.50, 37.50]    value 0.59 (information 1.32), travel 0.06, switch 0.05, continuity 0.15 -> bid 0.34
  task_id "RECON:14"    at [87.50, 62.50]    value 0.08 (information 0.18), travel 0.03, switch 0.05, continuity 0.15 -> bid -0.15
  task_id "RECON:15"    at [87.50, 87.50]    value 0.03 (information 0.06), travel 0.06, switch 0.05, continuity 0.15 -> bid -0.23
\end{promptbox}
\begin{promptbox}{Indigo}{Defender, independent team: response}
{
  "reasoning": "Contact 0 is moving towards the defended region, and neutralizing it is the highest priority to prevent a breach before the episode ends.",
  "scoring_profile": "balanced",
  "threat_priorities": [
    {
      "contact_id": 0,
      "priority": "HIGH"
    }
  ],
  "preferred_tasks": [
    "NEUTRALIZE:0:0"
  ]
}
\end{promptbox}
\begin{promptbox}{Ochre}{Commander, low-cost LLM: user message}
time_step: 210 (dt=0.1 sec), episode time left 19.0 s
blues (active):
  blue 0 pos=[92.8, 82.0] heading_deg=255 last issued assignment DEFEND:0 (issued at step 200)
  blue 1 pos=[82.0, 91.9] heading_deg=220 last issued assignment DEFEND:1 (issued at step 200)
  blue 2 pos=[70.8, 77.1] heading_deg=268 last issued assignment NEUTRALIZE:1:0 (issued at step 200)
  blue 3 pos=[77.1, 68.7] heading_deg=212 last issued assignment NEUTRALIZE:0:0 (issued at step 200)
red_contacts (as reported to you):
  contact 0 predicted pos=[61.3, 59.1] vel=[1.99, 2.00] observed_heading_deg=32 seen 1 steps ago
    forecast: approaching, breach 7.1 sec, urgency 0.89; optimistic interception: blue 0 8.1 sec, blue 1 8.4 sec, blue 2 4.5 sec, blue 3 3.9 s
  contact 1 predicted pos=[70.1, 50.4] vel=[2.00, 2.00] observed_heading_deg=67 seen 2 steps ago
    forecast: approaching, breach 9.1 sec, urgency 0.87; optimistic interception: blue 0 8.1 sec, blue 1 10.6 sec, blue 2 6.6 sec, blue 3 4.5 s
tasks:
  NEUTRALIZE:0:0 at [61.3, 59.1]
  NEUTRALIZE:1:0 at [70.1, 50.4]
  DEFEND:0 at [92.9, 82.0] sensor heading_deg=255 approach coverage 0.77 (0.58 not covered by any other station)
  DEFEND:1 at [82.0, 91.9] sensor heading_deg=220 approach coverage 0.83 (0.26 not covered by any other station)
  DEFEND:2 at [74.5, 82.0] sensor heading_deg=185 approach coverage 1.00 (0.55 not covered by any other station)
  DEFEND:3 at [82.0, 73.5] sensor heading_deg=230 approach coverage 1.00 (0.69 not covered by any other station)
  RECON:0 at [12.5, 12.5]
  RECON:1 at [12.5, 37.5]
  RECON:2 at [12.5, 62.5]
  RECON:3 at [12.5, 87.5]
  RECON:4 at [37.5, 12.5]
  RECON:5 at [37.5, 37.5]
  RECON:6 at [37.5, 62.5]
  RECON:7 at [37.5, 87.5]
  RECON:8 at [62.5, 12.5]
  RECON:9 at [62.5, 37.5]
  RECON:10 at [62.5, 62.5]
  RECON:11 at [62.5, 87.5]
  RECON:12 at [87.5, 12.5]
  RECON:13 at [87.5, 37.5]
  RECON:14 at [87.5, 62.5]
  RECON:15 at [87.5, 87.5]
coverage (YOUR record: steps since each region was last seen, -1 = never; share = fraction of the region in the approach sector):
  region 0 center=[12.5, 12.5] steps_since_seen=-1 share=1.00
  region 1 center=[12.5, 37.5] steps_since_seen=-1 share=1.00
  region 2 center=[12.5, 62.5] steps_since_seen=-1 share=1.00
  region 3 center=[12.5, 87.5] steps_since_seen=-1 share=1.00
  region 4 center=[37.5, 12.5] steps_since_seen=-1 share=1.00
  region 5 center=[37.5, 37.5] steps_since_seen=-1 share=1.00
  region 6 center=[37.5, 62.5] steps_since_seen=-1 share=1.00
  region 7 center=[37.5, 87.5] steps_since_seen=177 share=0.91
  region 8 center=[62.5, 12.5] steps_since_seen=-1 share=1.00
  region 9 center=[62.5, 37.5] steps_since_seen=198 share=1.00
  region 10 center=[62.5, 62.5] steps_since_seen=1 share=1.00
  region 11 center=[62.5, 87.5] steps_since_seen=1 share=0.55
  region 12 center=[87.5, 12.5] steps_since_seen=-1 share=1.00
  region 13 center=[87.5, 37.5] steps_since_seen=-1 share=1.00
  region 14 center=[87.5, 62.5] steps_since_seen=0 share=0.70
  region 15 center=[87.5, 87.5] steps_since_seen=-1 share=0.00
\end{promptbox}
\begin{promptbox}{Ochre}{Commander, low-cost LLM: response}
{"reasoning": "Blue 3 is optimally positioned to continue intercepting Red 0 (predicted breach in 7.1s, Blue 3 intercept time 3.9s). Blue 2 is optimally positioned to continue intercepting Red 1 (predicted breach in 9.1s, Blue 2 intercept time 6.6s). Blue 0 and 1 remain at their assigned defensive stations to cover the perimeter and watch the approach sector.", "task_by_agent": {"0": "DEFEND:0", "1": "DEFEND:1", "2": "NEUTRALIZE:1:0", "3": "NEUTRALIZE:0:0"}}
\end{promptbox}
\begin{promptbox}{Ochre}{Commander, near-frontier LLM: user message}
time_step: 180 (dt=0.1 sec), episode time left 22.0 s
blues (active):
  blue 0 pos=[92.8, 82.0] heading_deg=255 last issued assignment DEFEND:0 (issued at step 170)
  blue 1 pos=[55.2, 64.9] heading_deg=270 last issued assignment NEUTRALIZE:0:0 (issued at step 170)
  blue 2 pos=[74.5, 82.0] heading_deg=185 last issued assignment DEFEND:2 (issued at step 170)
  blue 3 pos=[77.1, 68.6] heading_deg=241 last issued assignment NEUTRALIZE:1:0 (issued at step 170)
red_contacts (as reported to you):
  contact 0 predicted pos=[55.4, 51.1] vel=[2.00, 2.00] observed_heading_deg=91 seen 1 steps ago
    forecast: approaching, breach 10.5 sec, urgency 0.85; optimistic interception: blue 0 9.5 sec, blue 1 3.4 sec, blue 2 7.9 sec, blue 3 5.4 s
  contact 1 predicted pos=[64.1, 44.4] vel=[2.00, 2.00] observed_heading_deg=61 seen 1 steps ago
    forecast: approaching, breach 11.4 sec, urgency 0.84; optimistic interception: blue 0 9.6 sec, blue 1 none, blue 2 9.6 sec, blue 3 6.0 s
tasks:
  NEUTRALIZE:0:0 at [55.4, 51.1]
  NEUTRALIZE:1:0 at [64.1, 44.4]
  DEFEND:0 at [92.9, 82.0] sensor heading_deg=255 approach coverage 0.77 (0.58 not covered by any other station)
  DEFEND:1 at [82.0, 91.9] sensor heading_deg=220 approach coverage 0.83 (0.26 not covered by any other station)
  DEFEND:2 at [74.5, 82.0] sensor heading_deg=185 approach coverage 1.00 (0.55 not covered by any other station)
  DEFEND:3 at [82.0, 73.5] sensor heading_deg=230 approach coverage 1.00 (0.69 not covered by any other station)
  RECON:0 at [12.5, 12.5]
  RECON:1 at [12.5, 37.5]
  RECON:2 at [12.5, 62.5]
  RECON:3 at [12.5, 87.5]
  RECON:4 at [37.5, 12.5]
  RECON:5 at [37.5, 37.5]
  RECON:6 at [37.5, 62.5]
  RECON:7 at [37.5, 87.5]
  RECON:8 at [62.5, 12.5]
  RECON:9 at [62.5, 37.5]
  RECON:10 at [62.5, 62.5]
  RECON:11 at [62.5, 87.5]
  RECON:12 at [87.5, 12.5]
  RECON:13 at [87.5, 37.5]
  RECON:14 at [87.5, 62.5]
  RECON:15 at [87.5, 87.5]
coverage (YOUR record: steps since each region was last seen, -1 = never; share = fraction of the region in the approach sector):
  region 0 center=[12.5, 12.5] steps_since_seen=-1 share=1.00
  region 1 center=[12.5, 37.5] steps_since_seen=-1 share=1.00
  region 2 center=[12.5, 62.5] steps_since_seen=-1 share=1.00
  region 3 center=[12.5, 87.5] steps_since_seen=-1 share=1.00
  region 4 center=[37.5, 12.5] steps_since_seen=-1 share=1.00
  region 5 center=[37.5, 37.5] steps_since_seen=6 share=1.00
  region 6 center=[37.5, 62.5] steps_since_seen=50 share=1.00
  region 7 center=[37.5, 87.5] steps_since_seen=147 share=0.91
  region 8 center=[62.5, 12.5] steps_since_seen=-1 share=1.00
  region 9 center=[62.5, 37.5] steps_since_seen=1 share=1.00
  region 10 center=[62.5, 62.5] steps_since_seen=15 share=1.00
  region 11 center=[62.5, 87.5] steps_since_seen=1 share=0.55
  region 12 center=[87.5, 12.5] steps_since_seen=-1 share=1.00
  region 13 center=[87.5, 37.5] steps_since_seen=-1 share=1.00
  region 14 center=[87.5, 62.5] steps_since_seen=0 share=0.70
  region 15 center=[87.5, 87.5] steps_since_seen=-1 share=0.00
\end{promptbox}
\begin{promptbox}{Ochre}{Commander, near-frontier LLM: response}
{
  "reasoning": "Keep blue 1 intercepting contact 0 and blue 3 intercepting contact 1: their optimistic interception times of 3.4 sec and 6.0 sec precede the respective breach forecasts of 10.5 sec and 11.4 sec. These fresh reports support continuing pursuit, while recognizing that red maneuvers could change the forecasts. Keep blues 0 and 2 at their current stations to preserve complementary perimeter coverage against undetected intruders and avoid unnecessary repositioning.",
  "task_by_agent": {
    "0": "DEFEND:0",
    "1": "NEUTRALIZE:0:0",
    "2": "DEFEND:2",
    "3": "NEUTRALIZE:1:0"
  }
}
\end{promptbox}
\begin{promptbox}{Peach}{Intruder, before reaching a gate: user message}
time_step: 50 (dt=0.1 sec)
you: red 0 pos=[29.4, 27.1] vel=[1.99, 2.00] heading_deg=45
phase: approach
defended_region_centre: [82.0, 82.0]
your_current_action: attack gate 2
gates:
  gate 0 at [69.4, 82.0] distance 67.9
  gate 1 at [70.2, 75.2] distance 63.0
  gate 2 at [74.0, 68.1] distance 60.6
  gate 3 at [82.0, 68.6] distance 67.0
teammates:
  red 1 pos=[38.1, 18.4] vel=[2.00, 2.00] heading_deg=45; chose attack gate 2 at step 40
blue_sightings (team-shared; others may exist unseen): none
\end{promptbox}
\begin{promptbox}{Peach}{Intruder, before reaching a gate: response}
{
  "reasoning": "There are no blue defenders detected, so I have no immediate need to evade. My current heading of 45 degrees is well-aligned for reaching Gate 2 at [74.0, 68.1]. I will continue my approach to gate 2 as it is a direct path to the defended region.",
  "action": "attack",
  "target_id": 2
}
\end{promptbox}
\begin{promptbox}{Peach}{Intruder, after passing a gate: user message}
time_step: 280 (dt=0.1 sec)
you: red 1 pos=[82.0, 69.5] vel=[-0.06, 1.99] heading_deg=196
phase: interior
defended_region_centre: [82.0, 82.0]
your_current_action: attack
teammates:
  red 0 pos=[53.6, 59.2] vel=[2.00, 2.00] heading_deg=244; chose evade blue 2 at step 270
blue_sightings (team-shared; others may exist unseen):
  blue 0 predicted pos=[80.1, 69.3] vel=[-1.60, -1.60] observed_heading_deg=227 seen 73 steps ago
  blue 1 predicted pos=[74.5, 82.0] vel=[-0.00, -0.00] observed_heading_deg=185 seen 51 steps ago
  blue 2 predicted pos=[49.5, 51.1] vel=[-1.14, -0.60] observed_heading_deg=63 seen 0 steps ago
  blue 3 predicted pos=[64.9, 64.4] vel=[-2.00, -0.71] observed_heading_deg=205 seen 0 steps ago
\end{promptbox}
\begin{promptbox}{Peach}{Intruder, after passing a gate: response}
{
  "reasoning": "I am in the interior phase and very close to the defended region center ([82.0, 82.0]). My current position is [82.0, 69.5], meaning I am directly south of the target. While blue 1 is nearby at y=82.0, it is offset by 7.5 units on the x-axis. Attacking directly is the optimal path to achieve the objective before the time limit, as I am already in the interior phase and the target is within reach.",
  "action": "attack",
  "target_id": -1
}
\end{promptbox}
\begin{promptbox}{Deep}{Defender, consensus team at high authority, low-cost LLM: user message}
sim_step: 180
arena_size: 100.00
defense_center: [82.00, 82.00]
you are blue agent_id=0
  your_position: [92.84, 81.99]
  your_velocity: [-0.00, -0.00]
  your_heading_deg: 255
active_teammates (positions known exactly; intentions unknown, no communication this cycle):
  agent_id=1 pos=[81.96, 91.91] vel=[0.31, 0.29] heading_deg=220
  agent_id=2 pos=[74.45, 81.99] vel=[0.00, -0.00] heading_deg=185
  agent_id=3 pos=[81.99, 73.53] vel=[-0.00, -0.00] heading_deg=218
contacts you believe in (YOUR belief only -- others may exist that you cannot see; age_steps is how old the underlying observation is):
  contact_id=0 believed_position=[55.36, 53.09] believed_velocity=[2.00, 2.00] observed_heading_deg=38 age_steps=1
  contact_id=1 believed_position=[64.08, 44.40] believed_velocity=[2.00, 2.00] observed_heading_deg=58 age_steps=1
coverage (steps since your knowledge of each region was refreshed; -1 = never):
  region_id=0 center=[12.50, 12.50] steps_since_seen=-1
  region_id=1 center=[12.50, 37.50] steps_since_seen=-1
  region_id=2 center=[12.50, 62.50] steps_since_seen=-1
  region_id=3 center=[12.50, 87.50] steps_since_seen=-1
  region_id=4 center=[37.50, 12.50] steps_since_seen=-1
  region_id=5 center=[37.50, 37.50] steps_since_seen=-1
  region_id=6 center=[37.50, 62.50] steps_since_seen=125
  region_id=7 center=[37.50, 87.50] steps_since_seen=98
  region_id=8 center=[62.50, 12.50] steps_since_seen=-1
  region_id=9 center=[62.50, 37.50] steps_since_seen=152
  region_id=10 center=[62.50, 62.50] steps_since_seen=1
  region_id=11 center=[62.50, 87.50] steps_since_seen=1
  region_id=12 center=[87.50, 12.50] steps_since_seen=-1
  region_id=13 center=[87.50, 37.50] steps_since_seen=163
  region_id=14 center=[87.50, 62.50] steps_since_seen=0
  region_id=15 center=[87.50, 87.50] steps_since_seen=160
your current task: DEFEND:0
episode time left: 22.0 s
your budgets: w_threat + w_engage + w_station + w_coverage + w_scout sum to at most 3.2; the priorities of your 2 believed contacts sum to at most 2
your candidate tasks (task_id, where it is, features, bid under the default weights):
  task_id "NEUTRALIZE:0:0"    at [55.36, 53.09]    urgency 0.86 x opportunity 1.00 (approaching, forecast breach 10.1 sec, interception forecast 9.6 sec), 1/(1+teammates) 0.25, age term 0.01, distance/25 1.89, leaving 0.66, switching 1, continuity 0.00 -> default bid 1.43
  task_id "NEUTRALIZE:1:0"    at [64.08, 44.40]    urgency 0.84 x opportunity 1.00 (approaching, forecast breach 11.4 sec, interception forecast 9.6 sec), 1/(1+teammates) 0.25, age term 0.01, distance/25 1.89, leaving 0.66, switching 1, continuity 0.00 -> default bid 1.42
  task_id "DEFEND:0"    at [92.86, 82.00]    coverage 0.66, distance/25 0.00, leaving 0.00, switching 0, continuity 0.00 -> default bid 0.38
  task_id "DEFEND:1"    at [82.00, 91.94]    coverage 0.00, distance/25 0.59, leaving 0.66, switching 1, continuity 0.00 -> default bid -0.25
  task_id "DEFEND:2"    at [74.46, 82.00]    coverage 0.00, distance/25 0.74, leaving 0.66, switching 1, continuity 0.00 -> default bid -0.28
  task_id "DEFEND:3"    at [82.00, 73.54]    coverage 0.20, distance/25 0.55, leaving 0.66, switching 1, continuity 0.00 -> default bid -0.15
  task_id "RECON:0"    at [12.50, 12.50]    information 0.82, distance/25 4.25, leaving 0.66, switching 1, continuity 0.00 -> default bid -0.06
  task_id "RECON:1"    at [12.50, 37.50]    information 0.60, distance/25 3.67, leaving 0.66, switching 1, continuity 0.00 -> default bid -0.13
  task_id "RECON:2"    at [12.50, 62.50]    information 0.43, distance/25 3.31, leaving 0.66, switching 1, continuity 0.00 -> default bid -0.19
  task_id "RECON:3"    at [12.50, 87.50]    information 0.34, distance/25 3.22, leaving 0.66, switching 1, continuity 0.00 -> default bid -0.23
  task_id "RECON:4"    at [37.50, 12.50]    information 0.91, distance/25 3.55, leaving 0.66, switching 1, continuity 0.00 -> default bid 0.02
  task_id "RECON:5"    at [37.50, 37.50]    information 0.77, distance/25 2.84, leaving 0.66, switching 1, continuity 0.00 -> default bid -0.01
  task_id "RECON:6"    at [37.50, 62.50]    information 0.48, distance/25 2.35, leaving 0.66, switching 1, continuity 0.00 -> default bid -0.12
  task_id "RECON:7"    at [37.50, 87.50]    information 0.31, distance/25 2.22, leaving 0.66, switching 1, continuity 0.00 -> default bid -0.19
  task_id "RECON:8"    at [62.50, 12.50]    information 0.98, distance/25 3.03, leaving 0.66, switching 1, continuity 0.00 -> default bid 0.07
  task_id "RECON:9"    at [62.50, 37.50]    information 0.93, distance/25 2.15, leaving 0.66, switching 1, continuity 0.00 -> default bid 0.09
  task_id "RECON:10"    at [62.50, 62.50]    information 0.42, distance/25 1.44, leaving 0.66, switching 1, continuity 0.00 -> default bid -0.10
  task_id "RECON:11"    at [62.50, 87.50]    information 0.11, distance/25 1.23, leaving 0.66, switching 1, continuity 0.00 -> default bid -0.23
  task_id "RECON:12"    at [87.50, 12.50]    information 0.93, distance/25 2.79, leaving 0.66, switching 1, continuity 0.00 -> default bid 0.06
  task_id "RECON:13"    at [87.50, 37.50]    information 1.00, distance/25 1.79, leaving 0.66, switching 1, continuity 0.00 -> default bid 0.14
  task_id "RECON:14"    at [87.50, 62.50]    information 0.54, distance/25 0.81, leaving 0.66, switching 1, continuity 0.00 -> default bid -0.01
  task_id "RECON:15"    at [87.50, 87.50]    information 0.00, distance/25 0.31, leaving 0.66, switching 1, continuity 0.00 -> default bid -0.23
\end{promptbox}
\begin{promptbox}{Deep}{Defender, consensus team at high authority, low-cost LLM: response}
{
  "reasoning": "I am shifting focus from defense to interception because there are two active intruders closing in on the defended sector. I have increased w_threat and w_engage to prioritize neutralizing these threats over maintaining current station coverage.",
  "weights": {
    "w_threat": 1.5,
    "w_engage": 1.2,
    "w_station": 0.1,
    "w_coverage": 0.2,
    "w_scout": 0.2,
    "w_attrition": 0.5,
    "w_belief_age": 0.1,
    "w_travel": 0.2,
    "w_scout_travel": 0.1,
    "w_leave": 0.3,
    "w_continuity": 0.2,
    "w_switch": 0.1
  },
  "contact_priorities": [
    {
      "contact_id": 0,
      "priority": 1.2
    },
    {
      "contact_id": 1,
      "priority": 0.8
    }
  ],
  "preferred_tasks": [
    "NEUTRALIZE:0:0",
    "NEUTRALIZE:1:0"
  ]
}
\end{promptbox}
 }
 \clearpage
\sinote{note:models}{Choice of the two LLMs}

\textbf{Why a low-cost model.} Every agent on both teams consults its own LLM once per second, so the main study alone made 160,788 calls in 900 episodes (Supplementary Table~\ref{tab:profiles}). The price and speed of the model therefore set how many seeds and conditions a study can afford. A deployed team in which every robot consults its own LLM once per second would face the same arithmetic.

\textbf{Selection of the low-cost model.} We chose the low-cost model in a selection study run before any study episode, on development seeds that the study never used, under a selection rule fixed before any answer was scored. Four low-cost models were compared (Supplementary Table~\ref{tab:models}). Each answered the same 60 requests for each of four interfaces (defender, commander, and intruder before and after passing a gate), three times each. The requests were built from recorded game states: 30 natural states, 24 constructed states that each test one consideration, such as an imminent intruder or a stale report, and 6 copies with their lines reordered. A rubric fixed in advance scored the decision that each answer produced from 0 (unsupported or unusable) to 2 (well supported), using only what the prompt showed, and the scores were scaled to 0 to 100. We also recorded the share of answers that the game could use and how many usable answers each model's provider returned per second with four requests at a time. The rule took the model with the highest score and every model within 5 points of it, and among these preferred the one that would finish the study sooner. DeepSeek V4.1 Flash scored highest. Gemini 3.1 Flash-Lite scored 2.9 points lower, within the 5 points, returned 3.4 times as many usable answers per second and had every answer usable on every interface, so it was selected. In a live check on one development seed, 36 episodes with every agent on Gemini 3.1 Flash-Lite had no failed call, whereas 36 episodes with DeepSeek V4.1 Flash had 32 failed calls under its provider's rate limits.

\textbf{Commander check before the main study.} The selection study ran every model at its lowest reasoning effort, and at that effort the commander failed checks on scouting and on assigning intruders while its prompt was being developed. Before the main study, we therefore ran a check to tell whether more reasoning or a more capable model was needed. Three configurations answered the same 60 commander requests three times each: the low-cost model at minimal and at medium reasoning effort, and GPT-6.1 Sol at medium effort. The requests had been recorded on development seeds, and a fixed random rule drew them from four kinds of state: 10 before contact in which no defender was scouting, 10 before contact in which one was, 20 in which the listed intruders were moving away and none approached, and 20 with intruders both approaching and moving away. Every configuration received the same earlier version of the commander prompt. Before any answer was scored, we fixed the pass rule: at least 179 of the 180 answers usable, no truncated or failed call, and at least 54 of 60 answers passing each of three checks, namely a scout assigned before contact and every listed intruder assigned in each of the two kinds of state with intruders. The low-cost model failed all three checks at minimal effort and passed them at medium effort, and all 180 answers of GPT-6.1 Sol passed every check (Supplementary Table~\ref{tab:commandercheck}). These requests had also been used while the prompt was developed, so the check is not an independent validation. It scores answers against the checks and does not measure breaches.

\textbf{Reasoning effort and tier.} Every LLM agent therefore runs at medium reasoning effort. The low-cost model is called at its provider's Flex tier, at half the standard price, after a development run on one seed passed a rule fixed in advance (30 of 30 valid episodes, no failed call).

\textbf{Why GPT-6.1 Sol.} To ask whether a finding depends on model capability, we needed a model clearly more capable than the low-cost one, whose price still allowed us to run it in the settings where that question arose. GPT-6.1 Sol's developer describes it as close in performance to its most capable model at a lower cost (Methods), and it comes from a different developer than the low-cost model. Unlike the low-cost model, it was not chosen by comparing candidates. It was the only more capable model we evaluated, in the commander check above, where all 180 of its answers passed. At the tiers used, its list price is 16 times the low-cost model's per prompt token and 13 times per output token, and a commander call took a median of 5.7\,sec against 2.0\,sec (Supplementary Table~\ref{tab:profiles}). We therefore ran it only where the low-cost model's results raised the question of capability: the consensus team at high authority (ideal network, four intruders), the commander team (ideal network and falsified reports, both intruder counts) and the airspace challenges.

\begin{table}[ht]
\setstretch{1}
\caption{\textbf{Low-cost models compared in the selection study.}
Development seeds only; no study episode was run and no breach was read for the selection. Rubric score: mean score of the decisions produced by each model's answers, scaled to 0 to 100, over 720 answers per model (60 requests per interface, four interfaces, three repetitions); unusable answers score 0. Usable answers: the lowest share, over the four interfaces, of answers that the game could apply. Throughput: usable answers per second with four requests at a time. Models (OpenRouter ids): \texttt{deepseek/deepseek-v4.1-flash}, \texttt{google/gemini-3.1-flash-lite}, \texttt{openai/gpt-6-luna}, \texttt{xiaomi/mimo-v2.6-flash}, each at its lowest reasoning effort.}
\label{tab:models}
\centering
\tablebodyfont
\begin{tabular}{@{}lcccc@{}}
\toprule
Model & Rubric score & Usable answers (\%) & Throughput (sec$^{-1}$) & Selected \\
\midrule
DeepSeek V4.1 Flash & 84.3 & 97.8 & 1.47 & \\
Gemini 3.1 Flash-Lite & 81.4 & 100 & 5.00 & yes \\
GPT-6 Luna & 78.3 & 86.1 & 2.03 & \\
MiMo V2.6 Flash & 71.0 & 82.8 & 0.45 & \\
\bottomrule
\end{tabular}
\end{table}
\begin{table}[ht]
\setstretch{1}
\caption{\textbf{Commander check before the main study.}
Development requests only; no study episode was run. Each configuration answered the same 60 commander requests three times, with the same earlier version of the commander prompt, and the pass thresholds were fixed before any answer was scored. Before contact: answers that assign at least one scout, over the 20 requests with no intruder listed. Moving away: answers that assign every listed intruder, over the 20 requests in which the listed intruders were moving away and none approached. Mixed: the same, over the 20 requests with intruders both approaching and moving away.}
\label{tab:commandercheck}
\centering
\tablebodyfont
\begin{tabular}{@{}lcccc@{}}
\toprule
Check & Pass threshold & \makecell{Low-cost,\\minimal effort} & \makecell{Low-cost,\\medium effort} & \makecell{GPT-6.1 Sol,\\medium effort} \\
\midrule
Usable answers & 179 of 180 & 180 & 180 & 180 \\
Truncated or failed calls & 0 & 0 & 0 & 0 \\
Before contact: scout assigned & 54 of 60 & 48 & 59 & 60 \\
Moving away: every intruder assigned & 54 of 60 & 53 & 59 & 60 \\
Mixed: every intruder assigned & 54 of 60 & 50 & 58 & 60 \\
Requests with any failing answer & -- & 21 of 60 & 3 of 60 & 0 of 60 \\
All checks passed & -- & no & yes & yes \\
\bottomrule
\end{tabular}
\end{table}

\end{document}